\documentclass{article}

\PassOptionsToPackage{dvipsnames}{xcolor}
 \usepackage[preprint]{neurips_2026}

\usepackage[utf8]{inputenc} %
\usepackage[T1]{fontenc}    %
\usepackage{hyperref}       %
\usepackage{url}            %
\usepackage{booktabs}       %
\usepackage{amsfonts}       %
\usepackage{nicefrac}       %
\usepackage{microtype}      %
\usepackage{xcolor}         %

\usepackage{adjustbox}

\usepackage{amssymb}%
\usepackage{amsmath}
\usepackage{mathtools}
\usepackage{amsthm}
\usepackage{xspace}

\usepackage{makecell}

\usepackage{multirow}

\usepackage{algorithm}
\usepackage{algpseudocode}

\usepackage{subcaption}

\definecolor{darkblue}{rgb}{0, 0, 0.5}
\hypersetup{colorlinks=true, citecolor=darkblue, linkcolor=darkblue, urlcolor=darkblue}

\usepackage{siunitx}
\usepackage{wrapfig}

\usepackage{numprint}

\usepackage{pifont}%
\usepackage{minted}

\definecolor{codebg}{rgb}{0.97,0.97,0.97}

\definecolor{blind_blue}{HTML}{547FEF}
\definecolor{blind_magenta}{HTML}{DC267F}
\definecolor{blind_yellow}{HTML}{FFB000}
\definecolor{blind_orange}{HTML}{FE6100}
\newcommand{\colourbase}{blind_orange}
\newcommand{\coloursubword}{blind_magenta}
\newcommand{\colourcharacter}{blind_blue}
\newcommand{\colourvariable}{purple}
\newcommand{\colourcombinatorial}{ForestGreen}

\newcommand{\colourtok}{black}

\usepackage[capitalize]{cleveref}

\theoremstyle{plain}

\newtheorem{theorem}{Theorem}[section]

\newtheorem{definition}[theorem]{Definition}
\newtheorem{observation}[theorem]{Observation}

\newcommand{\mytok}[2]{\newcommand{#1}{{\color{\colourtok}#2}}}

\definecolor{blind_blue}{HTML}{547FEF}
\definecolor{blind_magenta}{HTML}{DC267F}
\definecolor{blind_orange}{HTML}{FE6100}

\newcommand{\colourtoken}{blind_magenta}

\newcommand{\mymacro}[2]{\newcommand{#1}{{\color{\colourbase}#2}}}

\newcommand{\myvar}[2]{\newcommand{#1}{{\color{\colourvariable}#2}}}
\newcommand{\mycombinatorial}[2]{\newcommand{#1}{{\color{\colourcombinatorial}#2}}}
\newcommand{\mytoken}[2]{\newcommand{#1}{{\color{\colourtoken}#2}}}
\newcommand{\mycharacter}[2]{\newcommand{#1}{{\color{\colourcharacter}#2}}}
\newcommand{\inner}[1]{\left\langle #1 \right\rangle}
\newcommand{\mysubword}[2]{\newcommand{#1}{{\color{\coloursubword}#2}}}

\newcommand{\defn}[1]{\textbf{#1}}
\mycharacter{\character}{c}
\mycharacter{\byte}{b}
\mycharacter{\characters}{\mathbf{\character}}
\mycharacter{\bytes}{\mathbf{\byte}}
\newcommand{\bytesn}{\bytes^{(n)}}
\mycharacter{\alphabet}{\Sigma}
\mycharacter{\specialtokens}{\Phi}
\mytok{\dataset}{\mathcal{D}}
\mytoken{\token}{t}
\mytoken{\tokens}{\mathbf{\token}}
\mytoken{\vocab}{\mathcal{T}}
\mymacro{\tokenise}{\mathtt{tok}}
\mymacro{\concat}{\mathtt{concat}}
\mysubword{\subword}{s}
\mysubword{\subtoken}{t}

\mysubword{\vocabopt}{\vocab_{\mathtt{opt}}}
\mysubword{\subwords}{\mathbf{\subword}}
\mysubword{\subtokens}{\mathbf{\subtoken}}
\mymacro{\stringequiv}{\stackrel{{\circ}}{=}}
\newcommand{\defeq}{\stackrel{\texttt{\tiny def}}{=}}
\mymacro{\detokenise}{\mathtt{detok}}
\mymacro{\objectivefunc}{f}

\mycombinatorial{\vertices}{\mathcal{V}}
\mycombinatorial{\vertex}{v}
\mycombinatorial{\edges}{\mathcal{E}}
\newcommand{\edgesall}{\edges_{\mathtt{all}}}
\mycombinatorial{\edge}{e}
\newcommand{\edgesbyte}{\edges_{\mathtt{byte}}}

\newcommand{\edgestok}{\edges_{\mathtt{tok}}}
\newcommand{\vocabmax}{\vocab_{\mathtt{all}}}
\mycombinatorial{\colourset}{\mathcal{C}}
\mycombinatorial{\edgescolour}{c}
\newcommand{\vocabsize}{K}

\mymacro{\lengthfun}{\mathtt{length}}

\mymacro{\directtoken}{\tokenise_{\Rightarrow}}

\mymacro{\bottomuptoken}{\tokenise_{\uparrow}}

\newcommand{\charstring}[1]{{\color{\colourcharacter}#1}}
\newcommand{\bytestring}[1]{{\color{\colourcharacter}#1}}

\newcommand{\tokenstringnobrackets}[1]{{\color{\colourtoken}#1}}
\newcommand{\tokenstring}[1]{{\color{\colourtoken}\langle#1\rangle}}

\mymacro{\bijectionvocabsat}{\mathrm{Conv}_{\vocab\to\satvals}}
\mymacro{\bijectionmergesat}{\mathrm{Conv}_{\merges\to\satvals}}

\newcommand{\bpe}{\texttt{BPE}\xspace}

\DeclareMathOperator*{\argmin}{\mathrm{argmin}}

\newcommand{\tokname}{\texttt{ConvexTok}\xspace}
\newcommand{\bpetok}{\texttt{BPE}\xspace}

\newcommand{\biastok}{\texttt{Bias}\xspace}
\newcommand{\dettok}{\texttt{Det}\xspace}
\newcommand{\inttok}{\texttt{Int}\xspace}

\newcommand{\bpb}{\texttt{BpB}\xspace}
\newcommand{\core}{\texttt{CORE}\xspace}

\newcommand{\np}{\ensuremath{\mathsf{NP}}\xspace}

\newcommand{\pequalnp}{$\mathsf{P}=\mathsf{NP}$\xspace}

\makeatletter
\DeclareRobustCommand*{\escapeus}[1]{%
    \begingroup\@activeus\scantokens{#1\endinput}\endgroup}
\begingroup\lccode`\~=`\_\relax
\lowercase{\endgroup\def\@activeus{\catcode`\_=\active \let~\_}}
\makeatother
\usepackage{tabularray}
\usepackage{cellspace}
\newcommand\cincludegraphics[2][]{\raisebox{-0.3\height}{\includegraphics[#1]{#2}}}
\usepackage{setspace}
\newcommand{\makesf}[1]{\textsf{{\escapeus{#1}}}}

\usepackage{ifxetex}

\ifxetex
  \usepackage{fontspec}
\else
  \usepackage[T1]{fontenc}
  \usepackage[utf8]{inputenc}
\fi

\usepackage{xparse}

\ExplSyntaxOn
\NewDocumentCommand{\capitalize}{>{\SplitList{~}}m}
 {
  \seq_clear:N \l_capitalize_words_seq
  \ProcessList{#1}{\CapitalizeFirst}
  \seq_use:Nn \l_capitalize_words_seq { ~ }
 }
\NewDocumentCommand{\CapitalizeFirst}{m}
 {
  \capitalize_word:n { #1 }
 }

\sys_if_engine_pdftex:TF
 {
  \cs_set_eq:Nc \capitalize_tl_set:Nn { protected@edef }
 }
 {
  \cs_set_eq:NN \capitalize_tl_set:Nn \tl_set:Nn
 }

\cs_new_protected:Nn \capitalize_word:n
 {
  \capitalize_tl_set:Nn \l_capitalize_word_tl { #1 }
  \seq_if_in:NfTF \g_capitalize_exceptions_seq { \tl_to_str:n { #1 } }
   { \seq_put_right:Nn \l_capitalize_words_seq { #1 } } %
   { \seq_put_right:Nx \l_capitalize_words_seq { \tl_mixed_case:V \l_capitalize_word_tl } }
 }
\cs_generate_variant:Nn \tl_mixed_case:n { V }
\NewDocumentCommand{\AppendToList}{m}
 {
  \clist_map_inline:nn { #1 }
   {
    \seq_gput_right:Nx \g_capitalize_exceptions_seq { \tl_to_str:n { ##1 } }
   }
 }
\cs_generate_variant:Nn \seq_if_in:NnTF { Nf }
\seq_new:N \l_capitalize_words_seq
\seq_new:N \g_capitalize_exceptions_seq
\ExplSyntaxOff

\AppendToList{a,is,of,óf,and,by,the}

\title{Tokenisation via Convex Relaxations}

\newcommand{\ethzidx}{1}
\newcommand{\kenshoidx}{2}

\author{%
  Jan Tempus,
  Philip Whittington,$^\ethzidx$ 
  Craig W. Schmidt,$^\kenshoidx$ 
  Dennis Komm,$^\ethzidx$ 
  Tiago Pimentel$^\ethzidx$ \\
  $^\ethzidx$ETH Zurich, 
  $^\kenshoidx$Kensho Technologies \\
  \texttt{jan.tempus@gmail.com}, \texttt{craig.schmidt@kensho.com} \\
  \texttt{\{philip.whittington,dennis.komm,tiago.pimentel\}@inf.ethz.ch,} \\
}

\begin{document}

\maketitle

\begin{abstract}
Tokenisation is an integral part of the current NLP pipeline.
Current tokenisation algorithms such as \bpe and Unigram are greedy algorithms---they make locally optimal decisions without considering the resulting vocabulary as a whole.
We instead formulate tokeniser construction as a linear program and solve it using convex optimisation tools, yielding a new algorithm we call \tokname.
We find \tokname consistently improves intrinsic tokenisation metrics and the bits-per-byte (\bpb) achieved by language models; it also improves downstream task performance, but less consistently. Furthermore, \tokname\ allows the user to certify how far their tokeniser is from optimal (at a certain objective) via a lower bound, and we empirically found it to be within 1\% of optimal at common vocabulary sizes.\looseness=-1

\begin{tblr}{colspec = {Q[c,m]|X[l,m] Q[c,m]|X[l,m]}, stretch = 0}
    \cincludegraphics[width=1.2em, keepaspectratio]{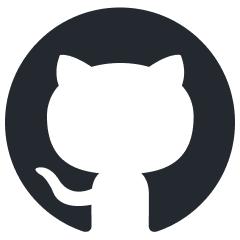}
     &
     \href{https://github.com/JanTempus/tokenisation_lp}{{\makesf{code in this link}}}
     &
     \cincludegraphics[width=1.2em, keepaspectratio]{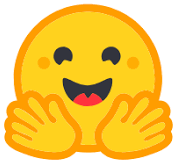}
     &
    \setstretch{.5}\href{https://huggingface.co/JanTempus/cross-over-climbmix400b-s7-tokenizers}{tokenisers in this link}
\end{tblr}
\vspace{-15pt}
\end{abstract}

\section{Introduction}

Tokenisation is an integral part of every language modelling pipeline, taking sequences of \defn{bytes} as input, and converting them into sequences of \defn{tokens}, which serve as input to a language model (LM).
Notably, while it is generally an open research question what makes a good tokeniser \citep{gowda-may-2020-finding,cognetta-etal-2024-two,ali-etal-2024-tokenizer,schmidt-etal-2024-tokenization}, a number of recent works show that a tokeniser's compression correlates (at least to some extent) with its downstream performance in LLMs \citep{galle-2019-investigating,zouhar-etal-2023-tokenization}.
Given these results, we might expect researchers to test how LMs would perform given a compression-optimal tokeniser.
Unfortunately, finding such optimal tokenisers is \np-hard \citep{kozma2024theoreticalanalysisbytepairencoding, whittington-etal-2025-tokenisation, lim2025a, kastreva2026tokenisation}. 
Hence, in practice approximate optimisation methods are employed.\looseness=-1

Currently, the de-facto standard tokenisation algorithm is byte-pair encoding \citep[\bpe;][]{gage1994new, sennrich-etal-2016-neural}, which originated as a compression algorithm. 
It is a simple greedy algorithm that iteratively merges the most frequent pair of tokens in a dataset into a new token until a desired vocabulary size is reached.
\bpe's use of greedily chosen pair-wise token-merging means that unnecessary intermediate tokens are created and compressive ability is lost.
Many approaches have been proposed as possible fixes to this problem \citep{cognetta24BPEtrimming, Chizhov2024bpegetspicky, Lia+25scaffoldBPE,liu2025superbpe,schmidt2025boundlessbpe}, but they all rely on relatively minor modifications or extensions to \bpe, and do not consider optimising tokenisation beyond greedy solutions.

The core technical contribution of this paper is showing how to solve tokenisation using \emph{polyhedral techniques}.\footnote{Polyhedral techniques are a branch of mathematics which translates combinatorial problems into studying the geometry of objects called polyhedra, or polytopes when the set is bounded. Polytopes are simply higher-dimensional version of polygons.}
We first identify an integer program (IP) which is equivalent to the optimisation problem solved by tokenisation algorithms.
Second, we relax this problem into a linear program (LP),\footnote{Integer programming optimises problems over discrete sets, while linear programming optimises problems over continuous sets. If an integer program requires a variable $x \in \{0,1\}$, we can relax it into a linear program to allow $x \in [0,1]$. } for which we can efficiently compute exact (or near-exact) solutions using a common solver.
Such an LP solution, however, includes `partial' tokens, which need to be discretised before we can convert it into a functional tokeniser. 
To this end, we propose three simple rounding schemes.
After rounding, constructing a tokeniser from this solution is trivial.
Additionally, solving the LP provides a lower bound on the compression achieved by any tokeniser on the used dataset.
Our method thus allows us to compute tight bounds on how close-to-optimal any tokeniser is.

Empirically, we evaluate our proposed tokeniser's performance in five parts, using \bpe as a baseline.
First, we natively analyse the behaviour of our constructed LP and the effect of different rounding techniques on its solutions' quality.
We see that even though the problem is \np-hard, the LP's solution is not far from integral (especially at larger vocabulary sizes). 
Second, we use our LP's solution to certify how close the various tokenisers are to being optimal compressors.
We see that already at common vocabulary sizes the various tokenisers we consider are within $1\% $ of optimal. 
Third, we study how stable tokenisers are to a specific choice of training dataset, finding that \bpe is consistently more stable than \tokname.
Fourth, we evaluate our tokenisers using common intrinsic metrics, including compression rate, vocabulary utilisation, and Rényi entropy.
For these metrics, one of our rounding schemes (\biastok) consistently outperforms all other tokenisers.
Finally, we evaluate the effect of our tokenisers on downstream language modelling performance.
In these experiments, a deterministic rounding scheme (\dettok) consistently performed best on bits-per-byte (\bpb), and often (although not always) outperformed \bpe on downstream (\core) tasks \citep{COREmetrics}. 

\vspace{-5pt}
\section{Tokenisation}

Before we formally define the term tokeniser, we start with some preliminary definitions. 
First, let $\alphabet$ be some alphabet, and define $\bytes \in \alphabet^*$ to be a \defn{byte-string},\footnote{Formally, $\alphabet^*$ denotes the Kleene star of $\alphabet$ (i.e., $\cup_{i=0}^{\infty} \alphabet^i$), and $\alphabet^+$ denotes its Kleene plus (i.e., $\cup_{i=1}^{\infty} \alphabet^i$).}
which we can expand as $\bytes=\byte_1\byte_2\cdots\byte_{|c|}$. 
Second, let a dataset be a multi-set of byte-strings, denoted by $\dataset = \{\bytesn\}^N_{n=1}$. 
A tokeniser's job is to \emph{segment} these byte-strings into substrings, which will correspond to the tokens used as input for our model.
As LMs require a finite set of tokens to compose their vocabularies, we introduce this constraint directly into the tokenisation step, defining the tokeniser's \defn{vocabulary} as a finite set of byte-substrings, i.e., $\vocab \subset \alphabet^+$ with $|\vocab| <\infty$. 
Further, to guarantee that any byte-string has at least one possible valid segmentation, we also require all elements of the alphabet to be contained in the vocabulary, or formally, $\alphabet \subseteq \vocab$.
In practice, we then often fix the size of the vocabulary given a \defn{budget} $\vocabsize$ to be $|\vocab| = |\alphabet| + \vocabsize$.
Finally, we term $\tokens \in \vocab^*$ as a \defn{token-string}.

\newcommand{\citeposs}[1]{\citeauthor{#1}'s (\citeyear{#1})}
\mytoken{\tokeniser}{\mathbb{T}}

Formally, we can now define a tokeniser as the 3-tuple $\tokeniser \defeq \langle \vocab, \tokenise, \detokenise \rangle$, where  $\tokenise\colon\alphabet^* \to \vocab^* $ is an \defn{encoding function}, which segments byte-strings into token-strings, and $\detokenise\colon\vocab^* \to \alphabet^*$ is a \defn{decoding function}, which maps token-strings back to byte-strings.
As we say an encoding function \emph{segments} a byte-string, we require that, whenever $\tokens = \tokenise(\bytes)$, we have that $\bytes = \token_1 \circ \token_2 \circ \cdots \circ \token_{|\tokens|}$.
Further, the decoding function simply undoes the encoding mapping, being thus defined as $\detokenise(\tokens) \smash{\defeq} \token_1 \circ \token_2 \circ \cdots \circ \token_{|\tokens|}$.
Notably, for a fixed vocabulary, many different encoding functions may successfully segment a byte-string;
e.g., given the vocabulary $\vocab = \{ \tokenstringnobrackets{d},\tokenstringnobrackets{o},\tokenstringnobrackets{g},\tokenstringnobrackets{do },\tokenstringnobrackets{og } \}$, we could segment byte-string $\bytestring{dog}$ as either $ \tokenise (\bytestring{dog}) =\tokenstring{do,g}$, $\tokenise (\bytestring{dog}) =\tokenstring{d,og}$, or $\tokenise (\bytestring{dog}) =\tokenstring{d,o,g}$.
A tokeniser $\tokeniser$ thus represents both a choice of vocabulary ($\vocab$) and of segmentation strategy ($\tokenise$).\looseness=-1

\subsection{What Do We Want from Our Tokeniser?}

Given the description above, we are left with a question: how should we select a tokeniser? 
If $\objectivefunc$ is an objective function, which, given a tokeniser, returns a score associated with it, we would ideally choose a tokeniser by computing a solution to the optimisation problem: $\argmin_{\tokeniser} \objectivefunc(\tokeniser) $.
Unfortunately, there are two issues with this approach: (i) it is not obvious which objective function to use, and (ii) given an objective, we don't know how to solve this optimisation problem efficiently.

Several objective functions have been proposed in response to (i), including  compression \citep{galle-2019-investigating}, unigram log-likelihood \citep{kudo-2018-subword}, or Rényi efficiency \citep{zouhar-etal-2023-tokenization}.
While we believe that more research should go into what optimisation function to use, we will focus on compression here, following a battery of prior work either proposing new tokenisers \citep{cognetta24BPEtrimming, Chizhov2024bpegetspicky, Lia+25scaffoldBPE} or theoretically analysing them \citep{zouhar-etal-2023-formal}.
Given a dataset ($\dataset$), we write our \defn{compression objective function} as
\begin{align}
    \objectivefunc(\tokeniser) = \sum_{\bytes \in \dataset} \lengthfun\big(\tokenise(\bytes)\big)\;,
\end{align}
where we note that the encoding function ($\tokenise$) implicitly depends on the tokeniser's vocabulary ($\vocab$).

\mymacro{\identityfunc}{\mathtt{id}}
\mymacro{\mergefunc}{\mathtt{merge}}

Even with an objective in hand, challenge  (ii) remains: finding a tokeniser that optimises it, and doing so efficiently. 
Unfortunately, as mentioned above, this optimisation problem is \np-hard, meaning that we cannot solve it efficiently, unless \pequalnp.
We thus focus on approximation algorithms instead.
The most popular such method is \defn{byte-pair encoding} (BPE), a greedy algorithm used in virtually all modern LLMs \citep{openai2024gpt4,touvron2023llama,biderman-etal-2023-pythia,olmo2026olmo3}.
Notably, \bpe is a greedy algorithm which iteratively chooses locally-optimal tokens to be merged one at a time; in general, however, these tokens may not be optimal globally, which may lead to the choice of a suboptimal vocabulary.
(See \cref{sec:bpe_description} for a more formal description of \bpe.)
\
As a small example, consider the dataset $\dataset = \{\charstring{abc}, \charstring{abd}, \charstring{abe}, \charstring{bc}, \charstring{bd}, \charstring{be}\}$ and $\vocabsize = 3$.
\bpe would first choose to merge $\tokenstring{a}$ and $\tokenstring{b}$ into a new token $\tokenstring{ab}$ for a saving of 3 symbols, and any further merge can save only 1 symbol, so with two more arbitrary merges, \bpe manages to save 5.
It would however be optimal to choose the new tokens $\tokenstring{bc}, \tokenstring{bd}, \tokenstring{be}$ to save 2 symbols three times, yielding a total saving of 6.\looseness=-1

\vspace{-5pt}
\section{Tokenisation via Convex Relaxations}
\vspace{-4pt}

\begin{wrapfigure}{r}{0.33\textwidth}
    \centering
    \vspace{-20pt}
    \adjustbox{trim=0 0 0 {0.6\height}, clip}{%
        \includegraphics[width=\linewidth]{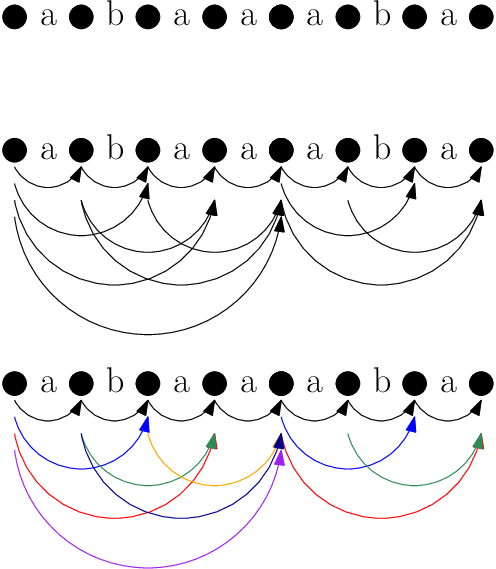}%
    }
    \caption{Tokenisation graph constructed from $\dataset = \{\charstring{abaa}, \charstring{aba}\}$. 
    Black edges represent $\edgesbyte$ and others represent $\edgestok$.}
    \label{fig:string_to_graph}
\end{wrapfigure}

Our paper's main contribution is an LP-based tokenisation algorithm which directly \emph{approximates} a globally-optimal tokeniser; thus, avoiding locally-optimal solutions.
As our tokeniser relies on a convex relation, we term it \tokname.
Before we get to our LP, however, we first express the problem as a shortest path problem in a directed acyclic graph, which will make the LP more intuitive. 
More specifically, given a dataset ($\dataset$), we construct a graph problem whose solutions can be easily translated into tokenisers, and then use this graph problem to build the LP.

We first define the graph.
For each byte-string in our dataset, we introduce a vertex for every position in between two bytes in the string; we also introduce a vertex for the position before the first and after the last byte in the string. 
A string of length $n$ thus contributes $n+1$ vertices, with the first and last vertices designated as its start and end vertices, respectively. 
We then merge the last vertex of each byte-string with the the first vertex of the next, essentially making them a single vertex.
Next, we connect each adjacent vertex within a byte-string with what we call a \defn{byte-edge}.
Finally, we connect each non-adjacent vertex within byte-strings with a \defn{token-edge};
this is denoted formally as:
\begin{subequations}\label{eq:tokgraph_vert_edges}
\begin{align}
    \vertices &\defeq \{\vertex^{n}_{i} \mid \bytes^{(n)} \in \dataset, 0 \leq i \leq \lengthfun(\bytes^{(n)})\},\qquad \forall_{\bytes^{(n)} \in \dataset}: \vertex^{n}_{\lengthfun(\bytes^{(n)})} = \vertex^{n+1}_{0} \\
    \edgesbyte &\defeq \{\langle \vertex^{n}_{i}, \vertex^{n}_{i+1} \rangle \mid \bytes^{(n)} \in \dataset, 0 \leq i < \lengthfun(\bytes^{(n)})\} \\
    \edgestok &\defeq \{\langle \vertex^{n}_{i}, \vertex^{n}_{j} \rangle \mid \bytes^{(n)} \in \dataset, 0 \leq i < \lengthfun(\bytes^{(n)}), i+2 \leq j \leq \lengthfun(\bytes^{(n)})\}
\end{align}
\end{subequations}
The set of edges in our graph is then defined as the union of byte- and token-edges: $\edgesall = \edgesbyte \cup \edgestok$.
Finally, we colour the token-edges based on the bytes they represent.
Formally, let the set of \defn{potential tokens} in dataset $\dataset$ be denoted by $\vocabmax$.
We can then define a \defn{colour-partition}  $\colourset$ as a \emph{partition} of the token-edges $\edgestok$ based on the bytes they represent:\looseness=-1
\begin{subequations}
\begin{align}
    \vocabmax &\defeq \{\bytes^{(n)}_{ij} \mid \bytes^{(n)} \in \dataset, 0 \leq i < \lengthfun(\bytes^{(n)}), i+2 \leq j \leq \lengthfun(\bytes^{(n)})\} \\
    \colourset &\defeq \bigg\{\underbrace{\{\langle \vertex^{n}_{i}, \vertex^{n}_{j} \rangle \mid \langle \vertex^{n}_{i}, \vertex^{n}_{j} \rangle \in \edgestok, \bytes^{(n)}_{ij} = \bytes'\}}_{\text{edges representing token }\bytes'} \Bigm| \bytes' \in \vocabmax\bigg\}
\end{align}
\end{subequations}
This is exemplified in \cref{fig:string_to_graph}.
Note that each set $\edgescolour \in \colourset$ represents a potential token, containing all edges with a certain colour.
Further, note that $\colourset$ partitions $\edgestok$; we thus have that $\cup_{\edgescolour \in \colourset}\, \edgescolour = \edgestok$ and that the sets of edges $\edgescolour \in \colourset$ are disjoint.
We can now recast tokenisation as a graph problem.

\newcommand{\vertexstart}{\vertex^0_0}
\newcommand{\vertexend}{\vertex^N_{L}}

\begin{definition}\label{def:graph-segmentation-tokenisation-graph}
We define a \defn{tokenisation graph} to be the 3-tuple $\langle \vertices, \edgesall, \colourset \rangle$.
Further, we say that $\vertexstart$ and $\vertexend$ are, respectively, its start and end vertices, where $L = \lengthfun(\bytes^{N})$.
Finally, given a budget $\vocabsize$, we define a \defn{graph-vocabulary} as a choice of $\vocabsize$ colours from $\colourset$, and a \defn{graph-segmentation} as a path from $\vertexstart$ to $\vertexend$ using only byte-edges, or token-edges from these $\vocabsize$ colours.\looseness=-1
\end{definition}

\mycombinatorial{\Gcomb}{G} %
\mycombinatorial{\Vcomb}{\mathcal{V}} %
\mycombinatorial{\Ecomb}{\mathcal{E}} %
\mycombinatorial{\Tcomb}{\mathcal{T}} %
\mycombinatorial{\scomb}{s} %
\mycombinatorial{\tcomb}{t} %
\mycombinatorial{\kcomb}{\vocabsize} %
\mycombinatorial{\dcomb}{d} %
\mycombinatorial{\Fcomb}{\mathcal{F}} %
\mycombinatorial{\ecomb}{e}
\mycombinatorial{\vcomb}{v}
\mycombinatorial{\ucomb}{u}
\mycombinatorial{\edgesfree}{\Fcomb}

\mycombinatorial{\basevocab}{S}

\myvar{\incidendcepriced}{\mathbf{P}}
\myvar{\incidencefree}{\mathbf{F}}
\myvar{\colourmatrix}{\mathbf{C}}
\myvar{\incidencefreeitem}{\mathbf{F}}
\myvar{\incidendcepriceditem}{\mathbf{P}}
\myvar{\colourmatrixitem}{\mathbf{C}}
\myvar{\vertexdiff}{\mathbf{d}}
\myvar{\usededgesfree}{\mathbf{f}}
\myvar{\usededgespriced}{\mathbf{p}}
\myvar{\usedcolor}{\mathbf{c}}

\myvar{\fvar}{f}
\myvar{\tvar}{t}
\myvar{\gvar}{g}
\myvar{\polytope}{\mathcal{Q}}
\myvar{\IP}{\mathcal{Q}^{\texttt{IP}}}

Note that, by construction, there is a one-to-one correspondence between a graph-vocabulary and a (traditional) vocabulary $\vocab$ with budget $ \vocabsize$, as the tokens in a traditional vocabulary have a direct mapping from the colours in a graph-vocabulary.
There is also a one-to-one correspondence between a graph-segmentation and the segmentation of a dataset $\dataset$ by a tokeniser $\tokeniser$; we can thus build an encoding function which segments each byte-string in $\dataset$ similarly to the graph.
Given a graph-vocabulary and graph-segmentation, we can thus construct an equivalent tokeniser $\tokeniser$.\footnote{For completeness, we also define the encoding function to encode byte-strings which are not in $\dataset$ using PathPiece \citep{schmidt-etal-2024-tokenization}, which is compression-optimal for a fixed vocabulary.}
Notably, the compression achieved by this tokeniser will be identical to the length of the graph-segmentation.
Given this relationship between a tokeniser's compression and a graph-segmentation's path, we now define a graph problem equivalent to the problem of finding compression-optimal tokenisers.

\begin{definition}\label{defn:tokenisation_shortest_path}
  Consider a tokenisation graph $\langle \vertices, \edgesall, \colourset \rangle$ and a budget $\vocabsize$.
  The \defn{shortest tokenisation problem} is to find the graph-segmentation in $(\vertices, \edgesall)$ with shortest path from $\vertexstart$ to $\vertexend$ using a graph-vocabulary of at most $\vocabsize$ colours.\footnote{We note that related variants of \Cref{def:graph-segmentation-tokenisation-graph,defn:tokenisation_shortest_path} have been studied before, as connectivity problems of labelled graphs, or coloured graphs \citep{PathsAndCyclesInColoredGraphs,Hassin2006ApproximationAA,ZhangMoreApproxAlgos,Hedge-graphs_mohsen_david}.}
\end{definition}

\mycombinatorial{\edgespriced}{\mathcal{P}}
\mycombinatorial{\edgespruned}{\mathcal{E}}

\vspace{-7pt}
\subsection{Generalising the Tokenisation Problem}
\vspace{-4pt}

The construction above creates a tokenisation graph that is equivalent to a typical tokenisation problem.
We now mildly generalise this construction to allow for more flexibility in the tokeniser's optimisation.
Starting from the set of vertices $\vertices$ and edges $\edgesall$ as constructed above (in \cref{eq:tokgraph_vert_edges}), we select any subset of $\edgesall$ to form a set of \defn{free edges} $\edgesfree \subseteq \edgesall$. 
These edges represent tokens which must always be included in a tokeniser's vocabulary, and which are thus available for ``free''; in the standard definition above, these would consist of the byte-edges $\edgesbyte$.\footnote{When we say ``free'' here, this is in terms of not occupying a part of the allowed budget $\vocabsize$. These edges are still costly in terms of increasing a path's length in a graph-segmentation.}
We then define a set of \defn{priced edges} as $\edgespriced \subseteq \edgesall \setminus \Fcomb$; these edges represent the tokens which can be added to the vocabulary---and are thus ``priced''---typically containing the entire set of token-edges.
Whenever $\edgespriced = \edgestok$, we have a graph with edges between all possible byte-sequences in the dataset as before.
However, it could be reasonable to consider in $\edgespriced$ only edges that do not cross unicode,
grapheme or morpheme boundaries (analogously to \texttt{Boundless}\bpe; \citealp{schmidt2025boundlessbpe}), or that represent tokens with non-negligible frequencies.\footnote{Our method can in fact be used to solve tokenisation with any set of candidate tokens, as defined by \citet{lim2025a}.}
We then denote this (sub)set of ``interesting'' edges by $\edgespruned \defeq \edgesfree \cup \edgespriced$.

\begin{definition}
\label{defn:comb_token_def}
We define a \defn{generalised tokenisation graph} as a 4-tuple $\langle \vertices,  \edgesfree, \edgespriced, \colourset \rangle$, where
    $(\vertices, \edgesfree \cup \edgespriced)$ represents a directed acyclic graph with free and priced edges, and
    $\colourset$ is a colour-partition of the priced edges.
    As before, a \defn{graph-vocabulary} is any subset of the colours $\colourset$ in this graph, and a \defn{graph-segmentation} is a path from $\vertexstart$ to $\vertexend$ using only edges from this graph-vocabulary or free edges.\looseness=-1
\end{definition}

Note that the relationship discussed in the previous section (between compression and graph-segmentation length) is preserved when working with generalized tokenisation graphs.

\vspace{-7pt}
\subsection{Tokenisation as an Integer Program}
\vspace{-3pt}

The sections above build a graph problem which is, in a sense, equivalent to the problem of finding compression-optimal tokenisers.
This graph problem, however, is not necessarily easier to solve than the original problem over strings.
We now leverage it to build yet another equivalent problem using integer programming. 
While the resulting IP is still \np-hard, we will relax it into an LP which we can solve efficiently. 
Later we will rely on rounding schemes to recover an approximately globally-optimal tokeniser.
Notably, relaxing the integral constraints to continuous ones is a design choice we take here.
Please see \Cref{app:branch-bound} for an alternative approach.
Furthermore, we note that linear programming has a long history of use in approximating algorithms for combinatorial problems \citep{LPtextbook,approx-algorithms-book-vazanri,approx-generalized-assignment,combinatorial-optimization}.\looseness=-1

\mymacro{\halfspace}{H}
\newcommand{\mathcomment}[1]{\textcolor{gray}{\text{#1}}}

Consider a generalised tokenisation graph $\langle \vertices, \edgesfree, \edgespriced, \colourset \rangle$.
To convert it into an IP, we first define a \defn{free incidence matrix}\footnote{We follow the convention that when a matrix is defined as $\{0,1\}^{\vertices \times \edgesfree}$ we may index it using elements of $ \vertices$ and $\edgesfree$.} $\incidencefree \in \{-1,0,1\}^{\vertices \times \edgesfree}$ and a \defn{priced incidence matrix} $\incidendcepriced \in \{-1,0,1\}^{\vertices \times \edgespriced}$ to have elements $\incidencefreeitem_{\vertex, \edge}$ and $\incidendcepriceditem_{\vertex, \edge}$ which encode whether an edge ($\edge \in \edgesfree$ or $\edge \in \edgespriced$, respectively) starts or ends at some vertex ($\vertex \in \vertices$).
Second, we define an \defn{edge-colour matrix} $\colourmatrix \in \{0,1\}^{\edgespriced \times \colourset}$ to have elements $\colourmatrixitem_{\edge, \edgescolour}$ encoding whether a priced edge $\edge \in \edgespriced$ has a specific colour $\edgescolour \in \colourset$.
Formally:
\begin{align}
    \incidencefreeitem_{\vertex, \edge} \!\defeq\! \left\{\!\!\!\!\begin{array}{rr}
         -1 \!& \texttt{if } \edge = \langle \vertex, \vertex' \rangle \\
         1 \!& \texttt{elif } \edge = \langle \vertex', \vertex \rangle  \\
         0 \!& \texttt{else} 
    \end{array} \right.\!\!\!\!,\,\,
    \incidendcepriceditem_{\vertex, \edge} \!\defeq\! \left\{\!\!\!\!\begin{array}{rr}
         -1 \!& \texttt{if } \edge = \langle \vertex, \vertex' \rangle  \\
         1 \!& \texttt{elif } \edge = \langle \vertex', \vertex \rangle \\
         0 \!& \texttt{else}
    \end{array} \right.\!\!\!\!,\,\,
    \colourmatrixitem_{\edge, \edgescolour} \!\defeq\! \left\{\!\!\begin{array}{rr}
         1 \!& \texttt{if } \edge \in \edgescolour \\
         0 \!& \texttt{else}
    \end{array} \right.\!\!
\end{align}

Recall that each colour $\edgescolour \in \colourset$ represents a potential token; $\colourmatrix_{\edge, \edgescolour} = 1$ then means that an edge $\edge$ represents an instance of $\edgescolour$'s token.
Notably, these matrices represent the structural constraints of a tokenisation problem.
We now define three vectors which we will use to define a specific tokeniser:
(i) a \defn{free token-instance vector} $\usededgesfree \in \{0, 1\}^{\edgesfree}$;
(ii) a \defn{priced token-instance vector} $\usededgespriced \in \{0, 1\}^{\edgespriced}$; and
(iii) a \defn{priced-colour vector}  $\usedcolor \in \{0, 1\}^{\colourset}$.
These vectors represent, respectively, which free and priced edges are being used to segment a tokenisation-graph (forming, together, a graph-segmentation) and which tokens are selected as colours (forming a graph-vocabulary).
As we can choose at most $\vocabsize$ colours, we introduce the constraint $\inner{1,\usedcolor } \leq \vocabsize$.
Further, a priced token-instance can only be used if its priced-colour is a part of the vocabulary, motivating the constraint $\usededgespriced - \colourmatrix \usedcolor \leq 0$.
Finally, we introduce flow-constraints \citep[from][]{MaxFlowPolytope,DantzigFlowPolytope} to guarantee that the solution $\usededgespriced,\usededgesfree$ forms a valid path (or segmentation) over the graph. 
To this end, we define a \defn{flow-difference} vector $\vertexdiff \in \{-1,0,1\}^{\vertices}$ which is defined as $\vertexdiff_{\vertexstart} = -1$, $\vertexdiff_{\vertexend} = 1$  for the starting and ending vertices, and as $\vertexdiff_{\vertex} = 0$ for other vertices.
This vector can then be used to guarantee that the input-output flow difference in all vertices is zero, except for the starting vertex (with a $-1$ flow difference) and the ending vertex (with a $+1$ flow difference):
$\incidendcepriced \usededgespriced +\incidencefree \usededgesfree  =\vertexdiff$.
Together, these constraints enforce that any choice of $\usededgesfree,\usededgespriced,\usedcolor$ corresponds to a valid graph-segmentation and graph-vocabulary.
The length of this graph-segmentation can then be measured as $\inner{1,\usededgespriced}+\inner{1,\usededgesfree}$.
We can now define the IP as:\looseness=-1
\begin{align}\label{eq:ip_full}
\!\!\begin{array}{cc}
     \min \!\!\!\!&\inner{1,\usededgespriced}+\inner{1,\usededgesfree} \\[0.2em]
     \text{s.t.}  & \usededgesfree,\usededgespriced,\usedcolor \in \polytope^{\texttt{IP}}
\end{array}\!\!\!\!,
 \texttt{ where }
  \polytope^{\texttt{IP}} \defeq\left\{
\begin{aligned}
\usededgesfree \in &\{0,1\}^{\edgesfree} \\
\usededgespriced \in &\{0,1\}^{\edgespriced} \\
\usedcolor \in &\{0,1\}^{\colourset}
\end{aligned}\!
 :\,\, 
 \begin{aligned}
 \incidendcepriced \usededgespriced +\incidencefree \usededgesfree  &=\vertexdiff  
 & \!\mathcomment{\# valid segmentation} \\
 \usededgespriced-\colourmatrix \usedcolor &\leq 0
 & \!\!\!\mathcomment{\# token in vocabulary} \\
\inner{1,\usedcolor }&\leq \vocabsize 
& \mathcomment{\# vocabulary budget} 
\end{aligned} \right\}  \!\!
\end{align}

\mymacro{\convertfunc}{\mathtt{ip2tok}}

Note that, as mentioned above, any choice of $\usededgesfree,\usededgespriced,\usedcolor \in \polytope^{\texttt{IP}}$ corresponds to a valid graph-vocabulary and graph-segmentation, which, in turn, correspond to a valid tokeniser.
We make two observations regarding this IP.

\begin{observation}
Let $\usededgesfree,\usededgespriced,\usedcolor$ be any element of $\polytope^{\texttt{IP}}$ (as defined in \cref{eq:ip_full}). 
We can convert this instance into a tokeniser $\tokeniser$ with compression $\objectivefunc(\tokeniser) = \inner{1,\usededgespriced}+\inner{1,\usededgesfree}$.
\end{observation}
\begin{observation}
Let $\tokeniser$ be any valid tokeniser with budget $\vocabsize$.
If $\edgesfree = \edgesbyte$ and $\edgespriced = \edgestok$, we can convert this tokeniser into an instance of $\usededgesfree,\usededgespriced,\usedcolor \in \polytope^{\texttt{IP}}$  with path length $\inner{1,\usededgespriced}+\inner{1,\usededgesfree} = \objectivefunc(\tokeniser) $.
\end{observation}

These observations guarantee that all elements in the space of solutions $\polytope^{\texttt{IP}}$ correspond to valid tokenisers with equivalent compressions, and that all valid tokenisers correspond to elements in the non-relaxed polytope $\polytope^{\texttt{IP}}$ with equivalent path-lengths. Please see \Cref{app:IP_tok_mapping} for a mapping between the integer program and token sequences.

\subsection{Approximating the Integer Program via Convex Relaxations}
\label{subsec:solving_lp}

As mentioned above, the IP in \cref{eq:ip_full} is still an \np-hard optimisation problem.
We can, however, use a convex relaxation to transform it into an LP, which can be solved efficiently.
To this end, we simply relax all variables $\usededgesfree,\usededgespriced,\usedcolor$ from being optimised over the discrete space $\{0, 1\}$ to be optimised over the continuous $[0,1]$ interval instead.
The priced-colour vector, for instance, becomes $\usedcolor \in [0,1]^{\colourset}$, allowing for partial tokens to be selected for a tokeniser.
Similarly, the priced token-instance vector allows for edges to be partially included in the solution.
Formally, we write:
\begin{align}\label{eq:lp_full}
\begin{array}{cc}
     \min \!\!\!\!&\inner{1,\usededgespriced}+\inner{1,\usededgesfree} \\
     \text{s.t.}  & \usededgesfree,\usededgespriced,\usedcolor \in \polytope
\end{array}\!\!\!\!,
 \texttt{ where }
  \polytope \defeq \left\{
\begin{aligned}
\usededgesfree \in &[0,1]^{\edgesfree} \\
\usededgespriced \in &[0,1]^{\edgespriced} \\
\usedcolor \in &[0,1]^{\colourset}
\end{aligned}
 :\,\, 
 \begin{aligned}
 \incidendcepriced \usededgespriced +\incidencefree \usededgesfree  &=\vertexdiff  
 & \mathcomment{\# valid segmentation} \\
 \usededgespriced-\colourmatrix \usedcolor &\leq 0
 & \!\!\mathcomment{\# token in vocabulary} \\
\inner{1,\usedcolor }&\leq \vocabsize 
& \mathcomment{\# vocabulary budget} 
\end{aligned} \right\}  
\end{align}
We term this relaxation of $\polytope$ the \defn{tokenisation polytope}.\footnote{A polytope is defined as the finite closed intersection of halfspaces. }
Notably, this problem can be solved efficiently (or near-exactly, up to numeric approximation).
Tokenisers' vocabularies, however, are discrete sets; thus, we cannot straightforwardly convert instances $\usededgesfree,\usededgespriced,\usedcolor \in \polytope$ of this LP that involve fractional variables into a tokeniser.
Hence, we will need to discretise the output of this LP, a process  typically known as \defn{rounding}.
We consider three rounding schemes here, noting that this is far from an exhaustive list: deterministic, biased, and integral-only rounding.%

\defn{Deterministic rounding (\dettok)} rounds the $\vocabsize$ colours in $\usedcolor$ with the largest value to 1, and sets others to zero.
\defn{Biased rounding (\biastok)}, instead ranks each colour in $\usedcolor$ by its value divided by the length of the token it represents ($\usedcolor/\lengthfun(\edgescolour)$, with a slight abuse of notation to let $\lengthfun(\edgescolour)$ denote the length of the corresponding token), and then rounds these values (to 1 or 0) as before. 
This biases the selection towards shorter tokens when LP scores are comparable, and is motivated by the fact that shorter tokens are more likely to occur outside the training strings in $\dataset$.
Finally, \defn{integral-only rounding (\inttok)} keeps only the colours $\usedcolor$ whose value is already essentially one, setting others to zero. 
This can be interpreted as selecting the tokens that the LP relaxation treats as forced, but it will typically return a vocabulary with fewer tokens than the allowed budget $\vocabsize$.
These rounding schemes are depicted in \cref{fig:rounding_schemes}. Given a (rounded) choice of $\usedcolor$, it is then trivial to compute optimal discrete values of $\usededgespriced$ and $\usededgesfree$ using a shortest path algorithm with the allowed colours.

\setminted[python]{
    escapeinside=@@,
    fontsize=\scriptsize,
    breaklines,
    frame=none,
    framesep=0pt,
    xleftmargin=0pt,
    xrightmargin=0pt,
    numbersep=0pt,
    autogobble
}
\begin{figure}[t]
    \centering
    \begin{subfigure}[t]{0.32\textwidth}
\begin{minted}[escapeinside=@@,fontsize=\footnotesize,breaklines]{python}
def det_rounding(@$\vocabsize$@, @$\usedcolor$@):
    col= sort(@$ \usedcolor$@, by=@$\usedcolor$@, ascending=False)
    @$ \usedcolor^{\prime} =0 $@
    for @$\edgescolour$@ in col@$ [0:\vocabsize]: $@
        @$ \usedcolor_{\edgescolour}^{\prime} = 1 $@ 
    return @$ \usedcolor^{\prime} $@
\end{minted}
    \end{subfigure}%
    \hfill%
    \begin{subfigure}[t]{0.34\textwidth}
\begin{minted}[escapeinside=@@,fontsize=\footnotesize,breaklines]{python}
def biased_rounding(@$ \vocabsize $@, @$ \usedcolor $@):
    col = sort(@$ \usedcolor $@,
               by=@$ \frac{\usedcolor_{\edgescolour}}{ \lengthfun(\edgescolour)} $@,
               ascending=False)
    @$ \usedcolor^{\prime}=0 $@
    for @$ \edgescolour $@ in col@$ [0: \vocabsize ]: $@
        @$\usedcolor_{\edgescolour}^{\prime} =1 $@ 
    return @$\usedcolor^{\prime}$@
\end{minted}
    \end{subfigure}%
    \hfill%
    \begin{subfigure}[t]{0.34\textwidth}
\begin{minted}[escapeinside=@@,fontsize=\footnotesize,breaklines]{python}
def integral_rounding(@$\usedcolor_{\edgescolour}$@):
    @$\usedcolor^{\prime}=0$@ 
    for @$\edgescolour \in\colourset $@:
        if @$\usedcolor_\edgescolour \geq 0.999:$@
            @$\usedcolor_\edgescolour^{\prime} =1 $@ 
    return @$\usedcolor^{\prime} $@
\end{minted}
    \end{subfigure}
    \vspace{-5pt}
    \caption{The \dettok (left), \biastok (center), and \inttok (right) rounding schemes.}
    \label{fig:rounding_schemes}
    \vspace{-7pt}
\end{figure}

\vspace{-5pt}
\section{Experimental Setup}

\vspace{-5pt}
\paragraph{Data.} We used ClimbMix400B \citep{nanochat}\footnote{Nanochat has an MIT license and is available at this \href{https://github.com/karpathy/nanochat}{GitHub repository}.} for training both tokenisers and LMs. ClimbMix400B is a large-scale pretraining corpus derived from NVIDIA's Nemotron-ClimbMix dataset \citep{diao2025climb}. It has been optimised automatically using multiple techniques to get a highly curated dataset. As is common when tokenising, the text is first pre-tokenised using a regular expression that splits the input into linguistically and typographically meaningful chunks. This pre-tokenisation step prevents tokenisers from crossing these coarse boundaries, so the learned tokens are built within local text categories such as words, numbers, punctuation, or whitespace blocks. We use the same regular expression for pretokenisation as nanochat \citep{nanochat}.

\vspace{-5pt}
\paragraph{Tokenisers.}
Throughout our experiments, we use \bpe as a baseline. We compare it to  \tokname with the three rounding schemes presented above: \biastok, \dettok, and \inttok. Further, 
we train all tokenisers with $593{,}920$ documents and with vocabulary sizes from $8k$ to $256k$, in power-of-two increments.
We note that we abbreviate powers of 2 in the vocabulary sizes below, e.g., 8$k$ corresponds to 8192. %
Furthermore, in practice, we may want to include a number of special tokens $\specialtokens $ (e.g., end-of-sequence, padding, masking, or output-start) in our tokeniser. We thus enforce $\alphabet \cup \specialtokens \subset \vocab$ and define $|\vocab| = |\alphabet| + |\specialtokens| + \vocabsize,$ where $|\vocab|$ is our vocabulary size.
Finally, at inference, our tokenisers use a standard shortest path algorithm (similar to UnigramLM's; \citealp{kudo-2018-subword}) to produce token-strings; \tokname thus has a similar runtime as UnigramLM and \bpe.

\vspace{-5pt}
\paragraph{Solving the LP.}
We solve the LP using the PDLP method implemented in the NVIDIA CuOPT library~\citep{nvidia_cuopt}.\footnote{NVIDIA CuOPT has an Apache-2.0 license and can be found in \href{https://docs.nvidia.com/cuopt/user-guide/latest/introduction.html}{this link}.}
We used the default stopping criteria of a primal dual gap of \num{10000}. We also merged identical subgraphs and adjusted the objective coefficients to preserve optimality. The final LP we solved has \num{99168445} constraints in total, \num{20093064} are equality constraints while \num{79075381} are inequality constraints. Furthermore, it has \num{105997943} variables corresponding to \num{79075380} priced edges and \num{18096951} free edges, with \num{8825612} different colour variables.

\vspace{-5pt}
\paragraph{Language Models.}
Following the nanochat standard, we trained GPT-style decoder-only transformer models.
Although the overall structure follows the standard GPT design, nanochat upgrades it with modern architectural techniques.
We also follow nanochat's training configuration, whose hyperparameter choices are motivated by empirical scaling laws \citep{Kaplan2020trn,Hoffman2022Scaling}. 
In our experiments, we fix these hyperparameters across tokenisers and vocabulary sizes. 
They do, however, vary with depth.
They are chosen using the \bpe baseline at vocabulary size $32k$, and are then reused unchanged for \biastok~and \dettok, as well as for different vocabulary sizes.
Model training time varied from roughly 17 minutes on 4 GH200s for the smallest models (with $135M$ parameters) up to 199 minutes on 16 GH200s for the largest models (with $3.5B$ parameters).
Training \tokname takes about 4 hours on 1 GH200. We estimate that in total we needed about 200 GPU hours across all of our training runs.
Please see \Cref{app:plots_training} for loss curves across training.

\begin{table}
\centering
\caption{Characteristics of the solutions for the LP.
\% of 1s (and of $\neg$ 0s) measures the ratio of $\usedcolor$ which are 1 (or not 0) divided by the vocabulary budget.}
\label{tab:lp_running_metrics}
\resizebox{\textwidth}{!}{%
\begin{tabular}{lcccccccccc}
\toprule
\multirow{2}{*}{\makecell{Vocabulary \\ Size}}
&&&& \multicolumn{2}{c}{$\usedcolor$}
& \multicolumn{2}{c}{$\usededgesfree$}
& \multicolumn{2}{c}{$\usededgespriced$} \\
\cmidrule(lr){5-6}
\cmidrule(lr){7-8}
\cmidrule(lr){9-10}
 &\makecell{\# steps} &  \makecell{Time (sec)} &\makecell{ LP Value} & \makecell{\% of 1s }& \makecell{\% of $\neg$ 0s } & \makecell{\# of 1s }& \makecell{\# of $ \neg$ 0s }& \makecell{\# of 1s }& \makecell{\# of $\neg$ 0s } \\
\midrule               
$8k$ & \num{65000} & 889.927 & \num{427366252}& $66.41\% $ & $151.01\%$& \num{1466122} & \num{5889259} & \num{577768} & \num{16244391}  \\      
$16k$ & \num{5600} & 182.661  & \num{393224648} & $73.22 \% $& $ 142.31 \% $& \num{1021750} & \num{4724165}  & \num{750072}  & \num{15712089} \\
$32k$ & \num{6800} & 196.123& \num{371886133} & $ 81.5\% $ & $129.95\% $ & \num{678332} & \num{3263220} & \num{1247839} & \num{13796121} \\
$64k$ & \num{6400} & 181.455& \num{359626839} & $ 84.39\% $ & $126.26\% $ & \num{349172} & \num{2402815} & \num{1380456} & \num{12738868}  \\
$128k$ & \num{9700} & 226.768 & \num{352723064}& $ 90.91\% $ & $ 113.97\% $ &\num{204014} &\num{1556595} &\num{1717709} &\num{10338653}\\
$256k$ & \num{9500} & 230.920 & \num{349028128} & $ 90.53\%$ & $114.59\% $ & \num{121694} & \num{1207058} & \num{1560819} & \num{9564059} \\
\bottomrule
\end{tabular}}
\vspace{-5pt}
\end{table}

\section{Results}

\paragraph{Behaviour of Solving the Linear Programs.}
Table~\ref{tab:lp_running_metrics} reports several optimisation statistics obtained when solving the LP relaxation for different vocabulary sizes.
Unsurprisingly, the primal objective (i.e., the LP value) decreases monotonically as the vocabulary size increases. 
More interestingly, the solution becomes more integral as the vocabulary size increases.\footnote{In practice, we used 0.999 as an approximation for 1 and 0.001 as an approximation for 0, due to numerical imprecision.}
This suggests that, at larger vocabulary sizes, the optimisation problem becomes less ambiguous, and therefore more tokens are already determined by the LP solution, leaving less work for the subsequent rounding procedure. Another thing to note is the relative stability of the running time, with the obvious exception of the much slower run with a vocabulary size of $8k$. This could be due to the small vocabulary budget making the feasible region substantially more constrained, causing the solver to spend much longer resolving the trade-offs between competing token choices. 
Finally, we also note how small the number of non-zero entries are in comparison to how many variables exist.

\paragraph{Performance of Rounding Schemes on Compression, and Certifying Optimality.}
\Cref{tab:integrality_gap} presents the LP's solutions,\footnote{
When reading \Cref{tab:integrality_gap}, one should keep in mind that the LP values are obtained numerically and are therefore subject to solver tolerances and numerical imprecision. The gap for \dettok at vocabulary size $256k$ is within that tolerance.
} as well as the compression obtained by our tokenisers.
Interestingly, it shows that the various tokenisers are close to the LP's corresponding solutions, especially \dettok. 
As the LP value provides a provable lower bound on the optimal compression, it follows that our tokenisers are close to optimal.
Considering how close the various rounded tokenisers are to optimal, we may assume that the rounding step is not necessarily that critical. 
This is especially so at vocabulary sizes of $128k$ and $256k$, where even \inttok (with fewer tokens) is close to optimal. 
This trend suggests a saturation effect; after the most compression-important tokens have been included, additional tokens have diminishing marginal value.
Finally, it is worth noting how close \bpetok also is to being compression-optimal, despite being a greedy method.

\begin{table}
\caption{Comparison between the LP relaxation value and the value obtained by each tokeniser.}
\label{tab:integrality_gap}
\resizebox{\textwidth}{!}{%
\centering
\begin{tabular}{llcccllccc}
\toprule
\makecell{Vocabulary \\ Size} & Tokeniser & \makecell{LP \\Value } & \makecell{Tokenised \\ Value } & \makecell{ Integrality \\ Gap Ratio } & \makecell{Vocabulary \\ Size} & Tokeniser & \makecell{LP \\Value } & \makecell{Tokenised \\ Value } & \makecell{ Integrality \\ Gap Ratio } \\
\cmidrule(lr){1-5} \cmidrule(lr){6-10}
\multirow{4}{*}{$8k$}&\bpetok &  \multirow{4}{*}{ \num{427366252}} & \num{441669198 } & $ 103.347\% $ & \multirow{4}{*}{$64k$}&\bpetok & \multirow{4}{*}{\num{359626839}} & \num{362088805  } & $100.685 \% $ \\
&\dettok  &   & \num{431045026} & $ 100.860 \% $ & &\dettok  &  & \num{359690431 } &  $ 100.018\% $\\
&\biastok &  & \num{448151069 } &  $104.863 \% $ & &\biastok &  & \num{361962390 } &  $ 100.649 \% $\\
&\inttok  &   & \num{524085422 } &  $ 122.631\% $ & &\inttok & & \num{366041138 } &  $ 101.784\% $\\
\cmidrule(lr){1-5} \cmidrule(lr){6-10}
\multirow{4}{*}{$16k$}&\bpetok   & \multirow{4}{*}{\num{393224648}} & \num{401802733 } & $  102.181\% $ & \multirow{4}{*}{$128k$}&\bpetok & \multirow{4}{*}{\num{352723064}} & \num{354079660 } & $100.385 \% $  \\
&\dettok  & & \num{ 394344618 } & $100.285 \% $ & &\dettok  &  & \num{353538264} &  $ 100.231\% $\\
&\biastok  & & \num{403757674} &  $ 102.679\% $ & &\biastok  & & \num{352753850 } &  $ 100.009\% $\\
&\inttok   & & \num{439773017 } &  $ 111.838\% $ & &\inttok & & \num{354160993} &  $ 100.408\% $\\
\cmidrule(lr){1-5} \cmidrule(lr){6-10}
\multirow{4}{*}{$32k$} &\bpetok & \multirow{4}{*}{\num{371886133}} & \num{ 376680738 } & $ 101.289\% $ & \multirow{4}{*}{$256k$} &\bpetok & \multirow{4}{*}{\num{349028128}}  & \num{349745778 } & $ 100.206\% $ \\
&\dettok &  & \num{ 372156050 } & $ 100.073 \% $ & &\dettok &  & \num{349020177 }   &  $99.997\% $\\
&\biastok & & \num{ 375996735 } &  $101.105 \% $ & &\biastok &  & \num{349264651 } &  $100.068\% $\\
&\inttok &  & \num{386575802 } &  $103.950 \% $ & &\inttok &   & \num{ 350090533} &  $100.304\% $\\
\bottomrule
\end{tabular}}
\vspace{-5pt}
\end{table}

\begin{figure}
    \centering
    \includegraphics[trim={0 0 0 2cm},clip,width=\linewidth]{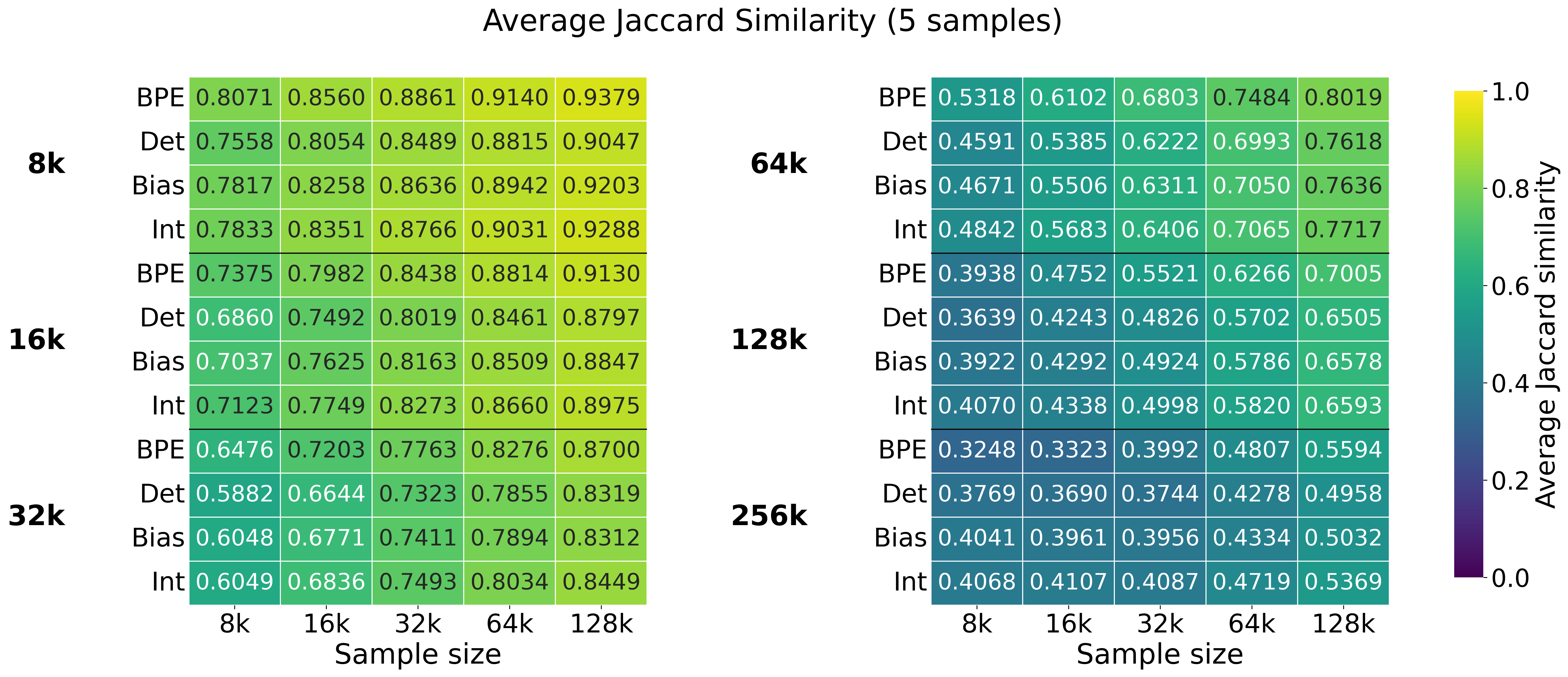}
    \vspace{-5pt}
    \caption{Average Jaccard similarity between vocabularies when retraining a tokeniser on independently sampled subsets of the same data (for different numbers of documents).}
    \label{tab:robustness_metrics}
    \vspace{-7pt}
\end{figure}

\begin{table}
\centering
\caption{Tokenisers' performances on intrinsic metrics.}
\label{tab:comprehensive_metrics_nlp_out_dist}
\resizebox{\textwidth}{!}{%
\begin{tabular}{llcccccccc}
\toprule
\makecell{Vocabulary \\ Size} & Tokeniser
& \makecell{Vocabulary \\ Utilisation (\textuparrow)}
& \makecell{Type-Token \\ Ratio (\textuparrow)}
& \makecell{Rényi Entropy \\ ($\alpha$=1) (\textuparrow)}
& \makecell{Rényi Entropy \\ ($\alpha$=2.5) (\textuparrow)}
& \makecell{Avg Token \\ Rank (\textdownarrow)}
& \makecell{Token \\ Length}
& \makecell{Tokens per \\ Line (\textdownarrow)}
& \makecell{Compression \\ Rate (\textuparrow)} \\
\midrule
\multirow{4}{*}{$8k$}
& \bpetok  & {98.6}\% & {0.0055} & {10.61} & {7.09} & {1121.5} & {3.99} & {63.8} & \textbf{{0.0014}} \\
& \dettok  & \underline{\textbf{{99.0}\%}} & \textbf{{0.0056}} & {10.61} & {7.02} & {1141.3} & {4.09} & \textbf{{62.3}} & \textbf{{0.0014}} \\
& \biastok & \underline{\textbf{{99.0}\%}} & {0.0054} & \textbf{{10.69}} &\underline{\textbf{{7.10}}} & {1171.0} & {3.93} & {64.6} & {0.0013} \\
& \inttok  & {98.6}\% & {0.0031} & {9.31} & {6.90} & \underline{\textbf{{492.3}}} & {3.39} & {75.6} & {0.0011} \\
\midrule
\multirow{4}{*}{$16k$}
& \bpetok  & {98.1}\% & {0.0120} & {10.90} & {6.88} & {1780.4} & {4.38} & {58.1} & \textbf{{0.0015}} \\
& \dettok  & \textbf{{98.8}\%} & \textbf{{0.0123}} & {10.90} & {6.83} & {1833.8} & {4.46} & \textbf{{57.0}} & \textbf{{0.0015}} \\
& \biastok & \textbf{{98.8}\%} & {0.0120} & \textbf{{10.95}} & {6.88} & {1860.9} & {4.36} & {58.3} & \textbf{{0.0015}} \\
& \inttok  & \textbf{{98.8}\%} & {0.0082} & {10.18} & \textbf{{6.99}} & \textbf{{1033.6}} & {4.03} & {63.6} & {0.0014} \\
\midrule
\multirow{4}{*}{$32k$}
& \bpetok  & {92.1}\% & {0.0240} & {11.04} & {6.73} & {2516.8} & {4.67} & {54.5} & \textbf{{0.0016}} \\
& \dettok  & {92.4}\% & \textbf{{0.0244}} & {11.03} & {6.70} & {2586.1} & {4.72} & \textbf{{53.9}} & \textbf{{0.0016}} \\
& \biastok & {92.6}\% & {0.0242} & \textbf{{11.06}} & {6.72} & {2609.1} & {4.68} & {54.4} & \textbf{{0.0016}} \\
& \inttok  & \textbf{{93.4}\%} & {0.0195} & {10.82} & \textbf{{6.79}} & \textbf{{1975.1}} & {4.55} & {56.0} & {0.0015} \\
\midrule
\multirow{4}{*}{$64k$}
& \bpetok  & {70.7}\% & {0.0384} & \underline{\textbf{{11.07}}} & {6.64} & {3138.0} & {4.85} & {52.4} & \textbf{{0.0017}} \\
& \dettok  & {71.2}\% & \textbf{{0.0389}} & {11.05} & {6.62} & {3210.4} & {4.88} & \textbf{{52.1}} & \textbf{{0.0017}} \\
& \biastok & {71.5}\% & {0.0388} & \underline{\textbf{{11.07}}} & {6.63} & {3226.2} & {4.85} & {52.4} & \textbf{{0.0017}} \\
& \inttok  & \textbf{{73.8}\%} & {0.0336} & {10.98} & \textbf{{6.66}} & \textbf{{2760.2}} & {4.80} & {53.0} & {0.0016} \\
\midrule
\multirow{4}{*}{$128k$}
& \bpetok  & {43.3}\% & {0.0481} & \textbf{{11.05}} & \textbf{{6.59}} & {3504.5} & {4.96} & {51.3} & \textbf{{0.0017}} \\
& \dettok  & {43.8}\% & {0.0489} & {11.03} & {6.58} & {3550.8} & {4.98} & \textbf{{51.1}} & \textbf{{0.0017}} \\
& \biastok & {44.0}\% & \textbf{{0.0490}} & {11.04} & {6.58} & {3566.7} & {4.97} & {51.2} & \textbf{{0.0017}} \\
& \inttok  & \textbf{{45.8}\%} & {0.0463} & {11.03} & {6.58} & \textbf{{3410.6}} & {4.96} & {51.3} & \textbf{{0.0017}} \\
\midrule
\multirow{4}{*}{$256k$}
& \bpetok  & {23.3}\% & {0.0525} & \textbf{{11.02}} & \textbf{{6.56}} & {3635.4} & {5.02} & {50.7} & \textbf{{0.0017}} \\
& \dettok  & {23.6}\% & {0.0532} & {11.01} & {6.55} & {3660.9} & {5.03} & \underline{\textbf{{50.6}}} & \textbf{{0.0017}} \\
& \biastok & {23.7}\% & \underline{\textbf{{0.0534}}} & {11.01} & {6.55} & {3671.1} & {5.02} & \underline{\textbf{{50.6}}} & \textbf{{0.0017}} \\
& \inttok  & \textbf{{25.2}\%} & {0.0514} & {11.00} & \textbf{{6.56}} & \textbf{{3563.0}} & {5.01} & {50.7} & \textbf{{0.0017}} \\
\bottomrule
\end{tabular}}
\vspace{-6pt}
\end{table}

\paragraph{Stability of Tokenisers to Randomness on the Training Sample.}
In \Cref{tab:robustness_metrics}, we study our tokenisers' stability relating to a specific choice of training set.
To this end, we train tokenisers using independently sampled subsets of our training data, and compare their vocabularies using the Jaccard similarity, defined, for two vocabularies $\vocab_1$ and $\vocab_2$, as $\frac{|\vocab_1 \cap \vocab_2|}{|\vocab_1 \cup \vocab_2|}$.
Further, we repeat this procedure for different training dataset sizes.
We say a tokeniser is more \defn{stable}, if its Jaccard similarity is higher.
\Cref{tab:robustness_metrics} shows that stability decreases as the vocabulary size increases.
As this trend is consistent for both \bpetok and \tokname, it is likely due to larger vocabularies containing more low-frequency tokens, which are more sensitive to the sampled dataset. 
Another thing to note is that \bpetok is consistently more stable than \tokname tokenisers. 
This aligns with \bpe being a local and frequency driven method; due to the power-law nature of natural language distributions, high-frequency merges are unlikely to change due to resampling.
The higher stability of \inttok compared to \biastok or \dettok is also to be expected, as \inttok includes only tokens which the LP deems ``critical''.\looseness=-1

\vspace{-2pt}
\paragraph{Performance of Tokenisers on Various Intrinsic Metrics.}
\cref{tab:comprehensive_metrics_nlp_out_dist}
shows how our tokenisers perform on a suite of classical intrinsic tokenisation metrics, computed using the library by \citet{meister_tokenizer_analysis_2025} on a held-out set of ClimbMix that was not used to train the tokenisers.
(See \Cref{tab:comprehensive_metrics_nlp_in_dist} and \Cref{tab:comprehensive_metrics_nlp} in the Appendix for similar metrics computed on, respectively, a subset of the tokeniser's training data and a multilingual dataset.)
This table shows that \tokname variants tend to outperform \bpe in vocabulary utilisation, type-token ratio, and tokens per line.
In terms of R\'enyi Entropy, results are more mixed, with \bpetok often outperforming both \biastok and \dettok.
Finally, the difference in the performance of \inttok vs.\ other tokenisers is stark. 
\inttok is consistently outperformed on compression by tokenisers with a smaller vocabulary size, e.g., \inttok with a budget of 16$k$ has worse compression than \dettok at 8k, even though (as can be verified using the \% of 1s information in \cref{tab:lp_running_metrics}) this \inttok tokeniser still has about 11.7$k$ tokens.

\paragraph{Performance of Models on \bpb.}
\Cref{tab:bpb_core_metrics} shows the performance (both in terms of \bpb and \core metrics) of LMs trained with \tokname or \bpetok.
(See \Cref{app:plots_downstream} for a more comprehensive breakdown of these results.)
This table shows that, for 12-layer models, the LP-based tokeniser outperforms \bpetok for all sizes, with the \dettok rounding scheme being the best at all vocabulary sizes but one.
For larger depths, \dettok wins for $32k$ and $128k$ vocabulary sizes, and is only slightly behind \bpetok for $8k$; \biastok trails behind \dettok.
Overall, in terms of \bpb, \dettok thus seems to be the best performing tokeniser.\looseness=-1

\paragraph{Performance of Models on \core.}
\Cref{tab:bpb_core_metrics} also presents our model's \core performance in downstream tasks.
(\core is a benchmark of downstream tasks covering reasoning, multiple choice and common sense questions; \citealp{COREmetrics}.)
For the \core metric, there is no clear trend in results.
At depth $12$, \tokname is better on average, but the results are close and \bpetok outperforms \tokname for some vocabulary sizes. 
At depth $24$, \bpetok is slightly better at $8k$, while \biastok is better at $32k$ with \dettok being the best at $128k$. %
Overall, the \core results suggest \tokname is at least competitive with \bpetok, but its advantage is less clear than for the \bpb metric.
A common trend across \core and \bpb, though, is that \tokname seems to be better at larger vocabulary sizes.
Notably, at the largest vocabulary sizes, \tokname still always matches or outperforms \bpetok for both metrics and all depths.

\begin{table}
\centering
\caption{Language model performances (as \bpb and \core metrics).
For depth 12, we trained three models per tokeniser (with different random seeds) and report average scores (standard errors in parentheses). Other configurations have a single training run due to computational restrictions.}
\label{tab:bpb_core_metrics}
\resizebox{\textwidth}{!}{%
\begin{tabular}{llcccccc}
\toprule
&& \multicolumn{3}{c}{Validation \bpb(\textdownarrow)}
& \multicolumn{3}{c}{\core Metrics (\textuparrow)} \\
\cmidrule(lr){3-5} \cmidrule(lr){6-8}
\makecell{Vocabulary Size} & Tokeniser
& \makecell{ Depth  12 } &\makecell{ Depth  18} & \makecell{Depth  24}
  & \makecell{Depth  12} & \makecell{Depth   18 }& \makecell{Depth   24} \\
\midrule
\multirow{3}{*}{$8k$}
& \bpetok   & $0.8785 (\pm 0.0012)$ &   \textbf{0.7767} & \textbf{{0.7145}} & \textbf{0.1345}$ (\pm 0.008)$ & \textbf{{0.195}} & \textbf{{0.270}} \\
& \biastok  & $0.8812 (\pm 0.0001)$ &   0.7793 & {0.7165} & $0.1318 {(\pm 0.002)}$  & \textbf{{0.195}} & {0.265} \\
& \dettok   & \textbf{0.8782}$(\pm 0.0008)$ &   0.7770      &0.7154          & $0.1294 {(\pm 0.005)}$ &   0.186          & 0.251          \\
\midrule
\multirow{3}{*}{$16k$}
& \bpetok    & ${0.8648} (\pm 0.0006)$ & --           & --          & $0.139 (\pm 0.020)$ & --           & --           \\
& \biastok   & $0.8655 (\pm 0.0001)$ & --           & --           &\textbf{0.144}$ (\pm 0.002)$ & --           & --           \\
& \dettok    & \textbf{0.8645}$(\pm 0.0003)$  & --           & --           & $0.137 (\pm 0.008)$ & --           & --           \\
\midrule

\multirow{3}{*}{$32k$}
& \bpetok   & $0.8536 (\pm 0.0002)$  & {0.7616} & {0.7042} &  0.150${(\pm 0.0066)}$&  \textbf{{0.232}} & {0.279} \\
& \biastok  & $0.8531 (\pm 0.0003)$  & {0.7608} & {0.7040} &   $0.148{(\pm 0.0087)}$ &  {0.220} & \textbf{{0.287}} \\
& \dettok   & \textbf{{0.8525}}$( \pm 0.0006)$  & \textbf{0.7560}        & \textbf{0.7033}          &  \textbf{0.151}$ {(\pm 0.0059)}$ &  0.220           & 0.278          \\
\midrule
\multirow{3}{*}{$64k$}
& \bpetok   & $0.8465 (\pm 0.0009)$  & --           & --           & $0.158 (\pm 0.002)$ & --           & --           \\
& \biastok  &  $0.8452 (\pm 0.0001)$  & --           & --           &$0.164 (\pm 0.007)$   & --           & --           \\
& \dettok   & $\textbf{0.8438}(\pm 0.0004) $ & --           & --           & \textbf{0.165}$ (\pm 0.006)$  & --           & --           \\
\midrule
\multirow{3}{*}{$128k$}
& \bpetok   & $0.8410 (\pm 0.0004)$  &{0.7515} & {0.6968} & $0.161 {(\pm 0.005)}$&  {0.233} & 0.296\\
& \biastok  & \textbf{0.8393}$(\pm 0.0007)$  & {0.7510} & 0.6959 & $0.156 {(\pm 0.009)}$ &  \textbf{{0.238}} &{0.296} \\
& \dettok   & $0.8403 (\pm 0.0006)$ &  \textbf{0.7504}          &  \textbf{0.6951}         & \textbf{0.170}$ {(\pm 0.008)}$ & 0.229         & \textbf{0.302}          \\
\midrule
\multirow{3}{*}{$256k$}
& \bpetok   &  $0.8411 (\pm 0.0002)$  &  --           & --           &  $0.159 (\pm 0.004)$ & --           & --           \\
& \biastok  & $0.8402 (\pm 0.0006)$  & --           & --           & \textbf{0.163}$(\pm 0.008)$  & --           & --           \\
& \dettok  &  \textbf{0.8398}$(\pm 0.0004)$ & --           & --          & $0.158 (\pm 0.007)$ & --           & --           \\

\bottomrule
\end{tabular}%
}
\vspace{-5pt}
\end{table}

\section{Conclusion}

In this paper, we reinterpreted tokenisation as a graph problem and formulated the resulting compression objective as an IP.
We then studied its corresponding relaxed formulation (i.e., its LP) experimentally, including both its solvability and three rounding procedures for converting fractional LP solutions into discrete tokenisers.
Our experiments show that the \tokname tokenisers are strong both in terms of intrinsic metrics and \bpb; results on \core metrics are more mixed, but suggest \tokname still at least matches \bpetok's performance.
Overall, our results suggest that globally optimised tokenisation is a promising alternative to locally greedy merge-based methods such as \bpe.

\paragraph{Limitations and Future Work.}
The focus of this paper is on compression; however, \Cref{tab:integrality_gap} shows that getting tokenisers with near-optimal compression on a dataset is not too hard, as at $128k$ and $256k$ our approach achieves results within $1\,\%$ of the LP lower bound. 
Significantly improving training-data compression is thus senseless under similar constraints.
Recent approaches like SuperBPE \citep{liu2025superbpe}, however, propose methods which partially ignore pretokenisation; it would be interesting to extend our analyses to those settings.
Further, extending our LP-based approach to other objective functions $\objectivefunc$ could also lead to novel and interesting tokenisers.

\section*{Acknowledgements}

We would like to thank Clara Meister, Gregor Bachmann, Dimitri von R\"utte, Pietro Lesci, Louis Barinka, Marius Mosbach, Thomas Hofmann and Saibo Geng for feedback and comments they have brought at various stages of the project.

\bibliography{custom}
\bibliographystyle{acl_natbib}

\newpage

\appendix
\section{Formal description of \bpe}\label{sec:bpe_description}

Let $\mergefunc_{\token', \token''}$ be a function which, given a token-string, replaces all occurrences of the bigram $\token', \token''$ in it with the merged token $\token^{\texttt{new}} \defeq \token' \circ \token''$.
The \bpe algorithm then works as follows: 
it initialises a character-level tokeniser, $\tokeniser_{0} = \langle \alphabet, \identityfunc, \detokenise \rangle$, where $\identityfunc$ is the identity function; 
for a pre-specified number of steps $\vocabsize$, 
it then takes the current tokeniser $\tokeniser_{k-1} = \langle \vocab_{k-1}, \tokenise_{k-1}, \detokenise \rangle$ and selects the pair of existing tokens $\token', \token'' \in \vocab_{k-1}$ which, if merged, will lead to maximal compression, and adds it to the vocabulary, as $\vocab_{k} = \vocab_{k-1} \cup \{\token^{\texttt{new}}\}$, and the encoding function as $\tokenise_{k}(\bytes) = \mergefunc_{\token', \token''}(\tokenise_{k-1}(\bytes))$.

\section{Alternatives to Linear Programming for solving IPs}\label{app:branch-bound}

An alterative choice to linear programming is to use methods such as branch-and-bound methods to solve the IP directly.
While branch-and-bound algorithms perform well in practice on highly structured IPs, such the travelling salesman problems \citep{Branch-And-Bound-TSP}, they exhibit poor scaling behaviours.
Hence, we decided to use linear programming with rounding, as this method scales better \citep{BranchAndBround}.
We hope future work explores other approaches to investigate this IP.\looseness=-1

\section{Mapping Between Vectors of the IP and the Token Sequence}\label{app:IP_tok_mapping}

In this section, we describe how to map an element of $\polytope$ to a tokeniser $\tokeniser$, and back. 
We give an informal sketch here, as the process is relatively simple and doing it formally would drown the reader with technical details which would obscure the idea.

\subsection{From an IP Solution to a Tokeniser}

Consider some free token-instance vector, priced token-instance vector, and priced-colour vector $ \usededgesfree,\usededgespriced,  \usedcolor \in \polytope^{\texttt{IP}}$.
First, we can recover the vocabulary by taking the union of all elements of $\usedcolor$ which are equal to 1 and the alphabet.
Formally, the vocabulary is $\{ \token_\edgescolour \in \colourset \mid \usedcolor_{\edgescolour}=1 \} \cup \alphabet $, where we denote as $\token_\edgescolour$ the token corresponding to colour $\edgescolour$.
Second, with the free and priced vectors $ \usededgesfree,\usededgespriced$ one can recover a valid segmentation of the dataset $\dataset$.
Recall that each potential use of a token to encode a byte-string $\bytes \in \dataset$ corresponds to exactly one edge in either $\usededgesfree$ or $ \usededgespriced$.
As such, to recover the segmentation of $\bytes  \in \dataset$ by selecting a token if and only if its corresponding edge from $\usededgesfree \cup \usededgespriced$ is equal to 1.
For completeness, given the extracted vocabulary, we use PathPiece \citep{schmidt-etal-2024-tokenization} to define the encoding function's behaviour on other byte-strings not in the dataset, computing:
\begin{align}
    \forall_{\bytes \notin \dataset}: \tokenise(\bytes) = \argmin_{\tokens \in \vocab^*} |\tokens|, \texttt{ s.t. } \bytes \stringequiv \tokens
\end{align}

\subsection{From a Tokeniser to an IP Solution}

Now we sketch how to take a tokeniser $\tokeniser$ and convert it to vectors contained within the IP.
This is simply the process described above, but in reverse. 
Given the tokeniser's vocabulary, we set each element $\usedcolor_\edgescolour$ in vector $ \usedcolor$ to 1 if and only if its corresponding token $\token_\edgescolour$ is included in the vocabulary.
For the dataset $\dataset$ then, we consider each of its byte-strings $\bytes \in \dataset $ and run it under the tokeniser's encoding function to get a token-string. 
For each token in this token-string, we then find its correspondence in either $\usededgesfree$ or $\usededgespriced$ and set it
to $1$; the rest of these vectors are all zero.

\newpage

\section{Intrinsic Tokenisation Results on our Tokeniser's Training Data}

\Cref{tab:comprehensive_metrics_nlp_in_dist} presents intrinsic tokenisation metrics on a subset of the data used to train our tokenisers.
The similarity between the compression results here and in \cref{tab:comprehensive_metrics_nlp_out_dist} suggests that our tokenisers generalise to new samples of the same distribution.

\begin{table}[H]
\centering
\caption{Tokenisers' performances on intrinsic metrics computed using a subset of our tokeniser's training data.}
\label{tab:comprehensive_metrics_nlp_in_dist}
\resizebox{\textwidth}{!}{%
\begin{tabular}{llcccccccc}
\toprule
\makecell{Vocabulary \\ Size} & Tokeniser
& \makecell{Vocabulary \\ Utilisation (\textuparrow)}
& \makecell{Type-Token \\ Ratio (\textuparrow)}
& \makecell{Rényi Entropy \\ ($\alpha$=1) (\textuparrow)}
& \makecell{Rényi Entropy \\ ($\alpha$=2.5) (\textuparrow)}
& \makecell{Avg Token \\ Rank (\textdownarrow)}
& \makecell{Token \\ Length}
& \makecell{Tokens per \\ Line (\textdownarrow)}
& \makecell{Compression \\ Rate (\textuparrow)} \\
\midrule
\multirow{4}{*}{$8k$}
& \bpetok  & {98.4}\% & {0.0054} & {10.59} & {7.08} & {1111.4} & {4.02} & {71.8} & {0.0013} \\
& \dettok  & \textbf{{98.9}\%} & \textbf{{0.0056}} & {10.59} & {7.00} & {1135.9} & {4.12} & \textbf{{70.0}} & \textbf{{0.0014}} \\
& \biastok & {98.8}\% & {0.0053} & \textbf{{10.66}} & \textbf{{7.09}} & {1154.0} & {3.96} & {72.7} & {0.0013} \\
& \inttok  & {98.3}\% & {0.0031} & {9.34} & {6.93} & \textbf{{499.1}} & {3.41} & {84.6} & {0.0011} \\
\midrule
\multirow{4}{*}{$16k$}
& \bpetok  & {97.8}\% & {0.0118} & {10.86} & {6.87} & {1747.9} & {4.41} & {65.5} & \textbf{{0.0015}} \\
& \dettok  & \textbf{{98.7}\%} & \textbf{{0.0121}} & {10.86} & {6.82} & {1798.9} & {4.50} & \textbf{{64.2}} & \textbf{{0.0015}} \\
& \biastok & \textbf{{98.7}\%} & {0.0118} & \textbf{{10.90}} & {6.87} & {1814.4} & {4.39} & {65.8} & \textbf{{0.0015}} \\
& \inttok  & \textbf{{98.7}\%} & {0.0081} & {10.19} & \textbf{{6.98}} & \textbf{{1039.3}} & {4.06} & {71.3} & {0.0014} \\
\midrule
\multirow{4}{*}{$32k$}
& \bpetok  & {91.9}\% & {0.0236} & {10.98} & {6.72} & {2438.0} & {4.71} & {61.5} & \textbf{{0.0016}} \\
& \dettok  & {92.2}\% & \textbf{{0.0240}} & {10.97} & {6.69} & {2505.5} & {4.76} & \textbf{{60.8}} & \textbf{{0.0016}} \\
& \biastok & {92.4}\% & {0.0238} & \textbf{{10.99}} & {6.72} & {2520.8} & {4.71} & {61.4} & \textbf{{0.0016}} \\
& \inttok  & \textbf{{93.3}\%} & {0.0192} & {10.78} & \textbf{{6.77}} & \textbf{{1939.9}} & {4.59} & {63.1} & {0.0015} \\
\midrule
\multirow{4}{*}{$64k$}
& \bpetok  & {70.2}\% & {0.0375} & \textbf{{11.00}} & {6.64} & {3009.4} & {4.89} & {59.3} & \textbf{{0.0016}} \\
& \dettok  & {70.6}\% & \textbf{{0.0379}} & {10.98} & {6.62} & {3067.2} & {4.92} & \textbf{{58.9}} & \textbf{{0.0016}} \\
& \biastok & {70.7}\% & {0.0377} & {10.99} & {6.63} & {3073.3} & {4.89} & {59.3} & \textbf{{0.0016}} \\
& \inttok  & \textbf{{73.4}\%} & {0.0329} & {10.92} & \textbf{{6.66}} & \textbf{{2665.7}} & {4.84} & {59.9} & \textbf{{0.0016}} \\
\midrule
\multirow{4}{*}{$128k$}
& \bpetok  & {42.6}\% & {0.0464} & \textbf{{10.97}} & \textbf{{6.59}} & {3317.1} & {4.99} & {58.1} & \textbf{{0.0017}} \\
& \dettok  & {43.2}\% & {0.0472} & {10.96} & {6.58} & {3358.7} & {5.01} & \textbf{{57.9}} & \textbf{{0.0017}} \\
& \biastok & {43.3}\% & \textbf{{0.0473}} & {10.96} & {6.58} & {3368.9} & {5.00} & {58.0} & \textbf{{0.0017}} \\
& \inttok  & \textbf{{45.3}\%} & {0.0449} & {10.96} & \textbf{{6.59}} & \textbf{{3243.3}} & {4.99} & {58.1} & \textbf{{0.0017}} \\
\midrule
\multirow{4}{*}{$256k$}
& \bpetok  & {22.8}\% & {0.0504} & \textbf{{10.94}} & \textbf{{6.56}} & {3415.6} & {5.05} & {57.4} & \textbf{{0.0017}} \\
& \dettok  & {23.1}\% & {0.0510} & {10.93} & \textbf{{6.56}} & {3430.7} & {5.06} & \textbf{{57.3}} & \textbf{{0.0017}} \\
& \biastok & {23.2}\% & \textbf{{0.0511}} & {10.93} & \textbf{{6.56}} & {3437.6} & {5.06} & \textbf{{57.3}} & \textbf{{0.0017}} \\
& \inttok  & \textbf{{24.7}\%} & {0.0494} & {10.93} & \textbf{{6.56}} & \textbf{{3356.0}} & {5.05} & {57.5} & \textbf{{0.0017}} \\
\bottomrule
\end{tabular}}
\vspace{-6pt}
\end{table}

\section{Intrinsic Tokenisation Results on a Multilingual Dataset}

\Cref{tab:comprehensive_metrics_nlp} presents intrinsic tokenisation metrics on the Flores+ dataset \citep{FLORES+}.
The difference between the compression results here and in \cref{tab:comprehensive_metrics_nlp_out_dist} suggests that \biastok generalises better to new out-of-distribution settings than \dettok.

\begin{table}[H]
\centering
\caption{Tokenisers' performances on intrinsic metrics computed using the FLORES+ dataset.}
\label{tab:comprehensive_metrics_nlp}
\resizebox{\textwidth}{!}{%
\begin{tabular}{llcccccccc}
\toprule
\makecell{Vocabulary \\ Size}& Tokeniser & \makecell{Vocabulary \\ Utilisation (\textuparrow )} & \makecell{Type-Token \\ Ratio (\textuparrow )} & \makecell{Rényi Entropy \\ ($\alpha$=1) ( \textuparrow ) } & \makecell{Rényi Entropy \\ ($\alpha$=2.5) (\textuparrow )} & \makecell{Avg Token \\ Rank (\textdownarrow )} & \makecell{Token \\ Length } &\makecell{ Tokens per \\ Line (\textdownarrow )} & \makecell{Compression \\ Rate (\textuparrow )} \\
\midrule
\multirow{4}{*}{$8k$}
&\bpetok & {67.7}\% & {0.0090} & {6.75} & {4.34} & {131.1} & {1.91} & {123.5} & {0.0081} \\
&\dettok  & {63.8}\% & {0.0085} & {6.73} & {4.31} & {127.8} & {1.93} & {123.4} & {0.0081} \\
&\biastok & \underline{\textbf{{69.2}\%}} & \textbf{{0.0093}} & \textbf{{6.79}} & {4.29} & {156.7} & {1.96} & \textbf{{121.9}} & \textbf{{0.0082}} \\
&\inttok  & {62.0}\% & {0.0051} & {6.41} & \textbf{{4.54}} & \underline{\textbf{{57.6}}} & {1.52} & {136.7} & {0.0073} \\
\midrule
\multirow{4}{*}{$16k$}
&\bpetok   & {46.5}\% & {0.0142} & {7.25} & \textbf{{4.90}} & {192.2} & {2.11} & {107.9} & {0.0093} \\
&\dettok  & {44.3}\% & {0.0138} & {7.29} & {4.86} & {196.3} & {2.16} & {105.8} & {0.0095} \\
&\biastok  & \textbf{{48.1}\%} & \textbf{{0.0154}} & \textbf{{7.37}} & {4.84} & {233.9} & {2.19} & \textbf{{102.9}} & \textbf{{0.0097}} \\
&\inttok   & {42.7}\% & {0.0084} & {6.76} & {4.80} & \textbf{{89.1}} & {1.77} & {124.4} & {0.0080} \\
\midrule
\multirow{4}{*}{$32k$}
&\bpetok & {28.7}\% & {0.0201} & {7.71} & \textbf{{5.27}} & {272.5} & {2.30} & {93.5} & {0.0107} \\
&\dettok  & {27.7}\% & {0.0200} & {7.76} & {5.25} & {285.4} & {2.36} & {91.0} & {0.0110} \\
&\biastok & \textbf{{29.4}\%} & \textbf{{0.0216}} & \textbf{{7.85}} & \textbf{{5.27}} & {322.5} & {2.39} & \textbf{{89.4}} & \textbf{{0.0112}} \\
&\inttok & {26.6}\% & {0.0132} & {7.22} & {5.13} & \textbf{{153.1}} & {2.06} & {108.8} & {0.0092} \\
\midrule
\multirow{4}{*}{$64k$}
&\bpetok & {16.7}\% & {0.0301} & {8.79} & \textbf{{6.68}} & {432.4} & {2.57} & {72.7} & {0.0138} \\
&\dettok & {16.6}\% & {0.0313} & {8.84} & {6.33} & {475.6} & {2.63} & {69.9} & {0.0143} \\
&\biastok & \textbf{{17.5}\%} & \textbf{{0.0334}} & \textbf{{8.91}} & {6.37} & {519.6} & {2.65} & \textbf{{69.1}} & \textbf{{0.0145}} \\
&\inttok & {15.8}\% & {0.0195} & {7.82} & {5.55} & \textbf{{245.6}} & {2.30} & {90.1} & {0.0111} \\
\midrule
\multirow{4}{*}{$128k$}
&\bpetok & {9.4}\% & {0.0405} & {9.52} & \textbf{{7.25}} & {632.3} & {2.83} & {61.0} & {0.0164} \\
&\dettok  & {9.7}\% & {0.0442} & {9.70} & {6.83} & {742.1} & {2.91} & {57.9} & {0.0173} \\
&\biastok  & \textbf{{10.0}\%} & \textbf{{0.0459}} & \textbf{{9.74}} & {6.85} & {779.8} & {2.92} & \textbf{{57.6}} & \textbf{{0.0174}} \\
&\inttok & {9.3}\% & {0.0321} & {8.72} & {6.37} & \textbf{{469.7}} & {2.66} & {69.6} & {0.0144} \\
\midrule
\multirow{4}{*}{$256k$}
&\bpetok & {5.4}\% & {0.0548} & {10.28} & \underline{\textbf{{7.71}}} & {951.0} & {3.10} & {51.9} & {0.0193} \\
&\dettok & {5.8}\% & {0.0608} & {10.41} & {7.56} & \num{1107.0} & {3.19} & \textbf{{49.8}} & \underline{\textbf{{0.0201}}} \\
&\biastok & \textbf{{5.9}\%} & \underline{\textbf{{0.0624}}} & \underline{\textbf{{10.46}}} & {7.60} & \num{1140.2} & {3.19} & {49.9} & {0.0200} \\
&\inttok & {5.2}\% & {0.0391} & {9.09} & {6.74} & \textbf{{590.3}} & {2.80} & {63.9} & {0.0157} \\
\bottomrule
\end{tabular}}
\vspace{-6pt}
\end{table}

\newpage

\section{Plots for Intrinsic Results on the FLORES+ Dataset}
\label{app:nlp_plots_flores}

\begin{figure}[H]
    \begin{subfigure}[b]{0.4\textwidth}
        \centering
        \includegraphics[width=\linewidth]{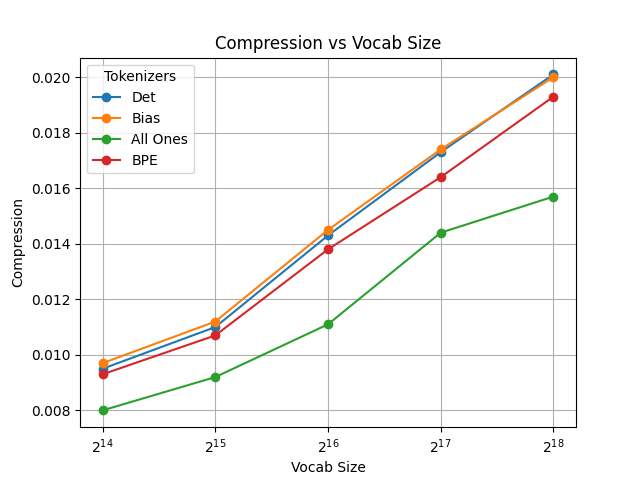}%
    \end{subfigure}\hfill
        \begin{subfigure}[b]{0.4\textwidth}
        \centering
        \includegraphics[width=\linewidth]{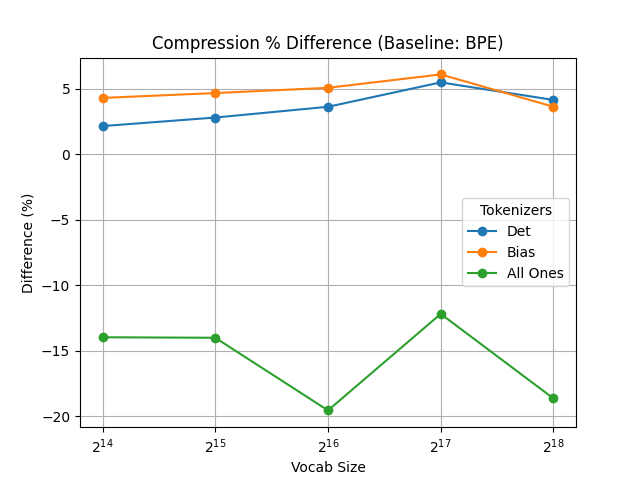}%
    \end{subfigure}\hfill
      \caption{
      Compression by the different tokenisers. (Left) absolute and (Right) relative to \bpe.}
\end{figure}

\begin{figure}[H]
\centering
\begin{subfigure}[b]{0.4\textwidth}
\centering
\includegraphics[width=\linewidth]{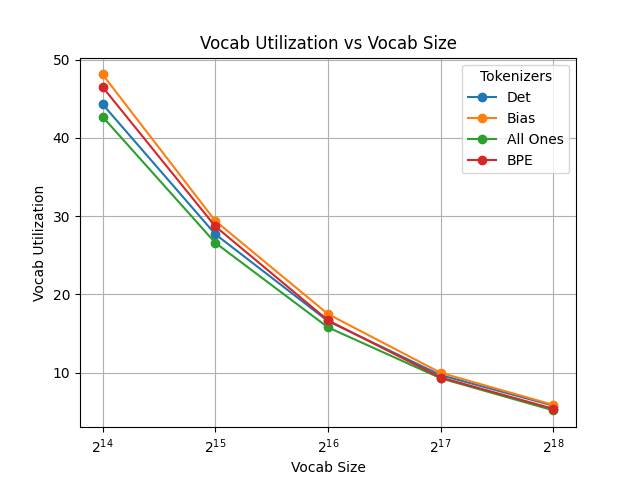}%
\end{subfigure}\hfill
    \begin{subfigure}[b]{0.4\textwidth}
        \centering
        \includegraphics[width=\linewidth]{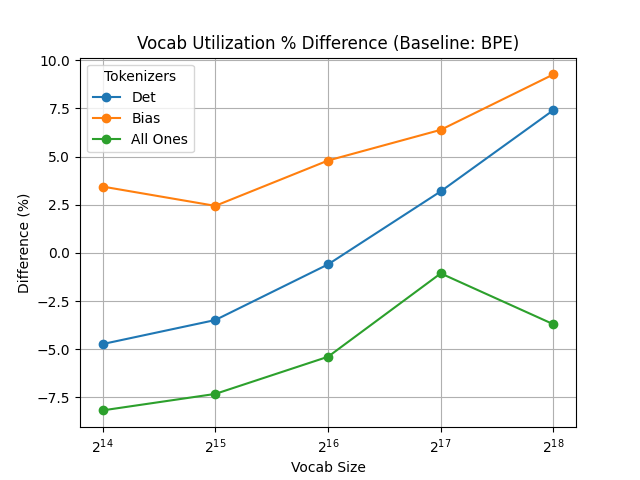}%
    \end{subfigure}
    \caption{Vocabulary utilisation by different tokenisers. (Left) absolute and (Right) relative to \bpe.}
\end{figure}

\begin{figure}[H]
    \begin{subfigure}[b]{0.4\textwidth}
        \centering
        \includegraphics[width=\linewidth]{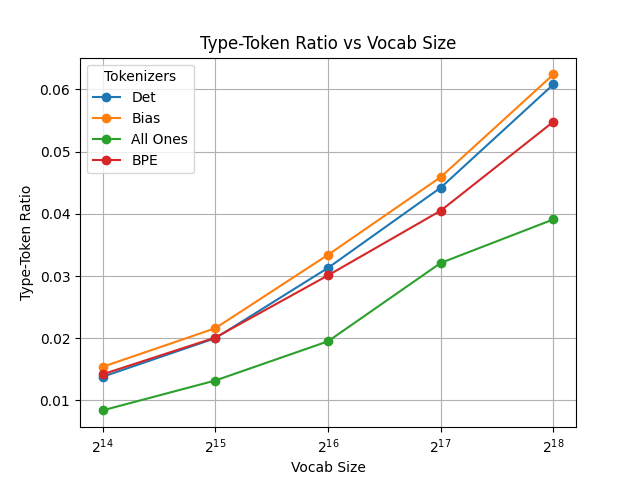}%
    \end{subfigure}\hfill
        \begin{subfigure}[b]{0.4\textwidth}
        \centering
        \includegraphics[width=\linewidth]{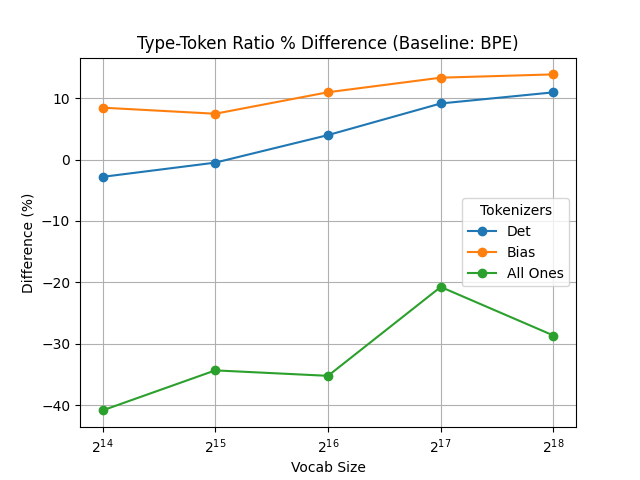}%
    \end{subfigure}
    \caption{Type-token ratio by different tokenisers. (Left) absolute and (Right) relative to \bpe.}
\end{figure}

\begin{figure}[H]
    \centering
    \begin{subfigure}[b]{0.4\textwidth}
        \centering
        \includegraphics[width=\linewidth]{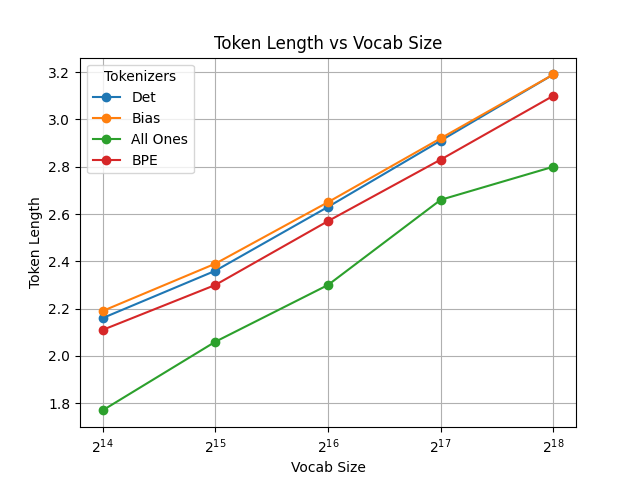}%
    \end{subfigure}\hfill
    \begin{subfigure}[b]{0.4\textwidth}
        \centering
        \includegraphics[width=\linewidth]{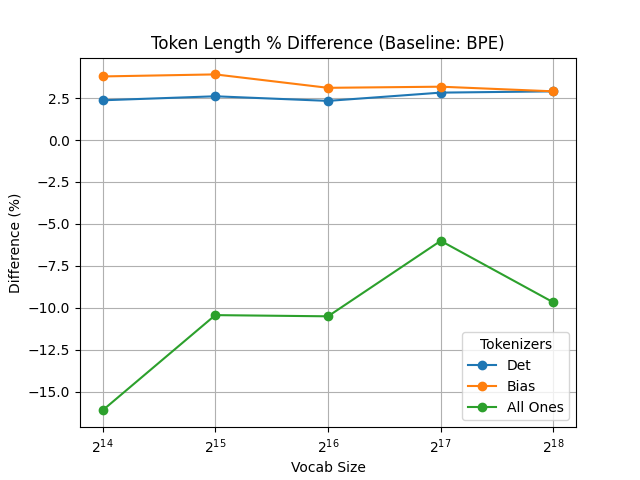}%
    \end{subfigure}
    \caption{Token length of different tokenisers. (Left) absolute and (Right) relative to \bpe.}
\end{figure}

\begin{figure}[H]
    \centering
    \begin{subfigure}[b]{0.4\textwidth}
        \centering
        \includegraphics[width=\linewidth]{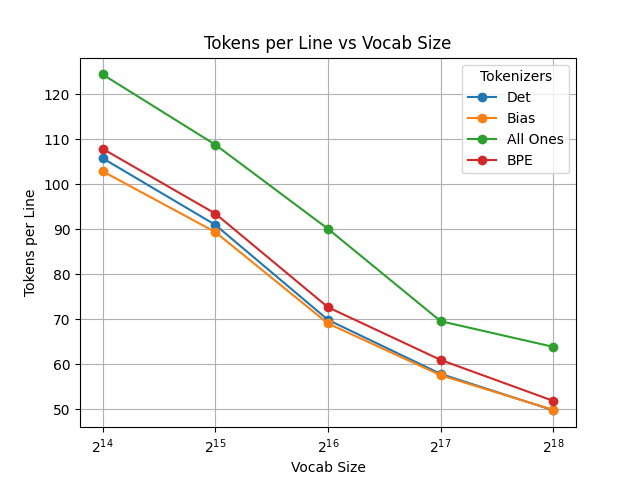}%
    \end{subfigure}\hfill
    \begin{subfigure}[b]{0.4\textwidth}
        \centering
        \includegraphics[width=\linewidth]{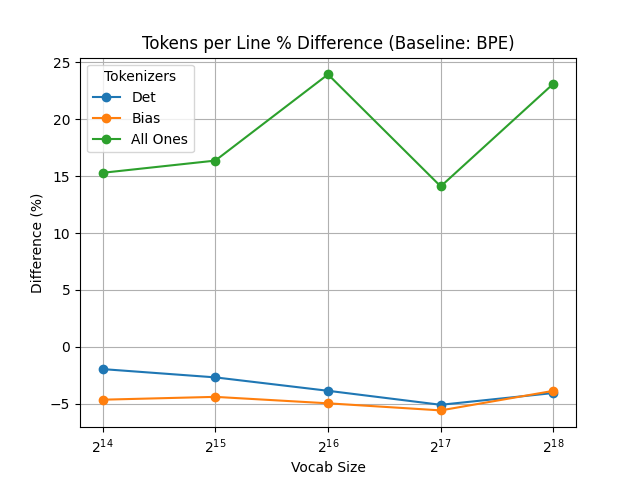}%
    \end{subfigure}
    \caption{Tokens per line of different tokenisers. (Left) absolute and (Right) relative to \bpe.}
\end{figure}

\begin{figure}[H]
    \begin{subfigure}[b]{0.4\textwidth}
        \centering
        \includegraphics[width=\linewidth]{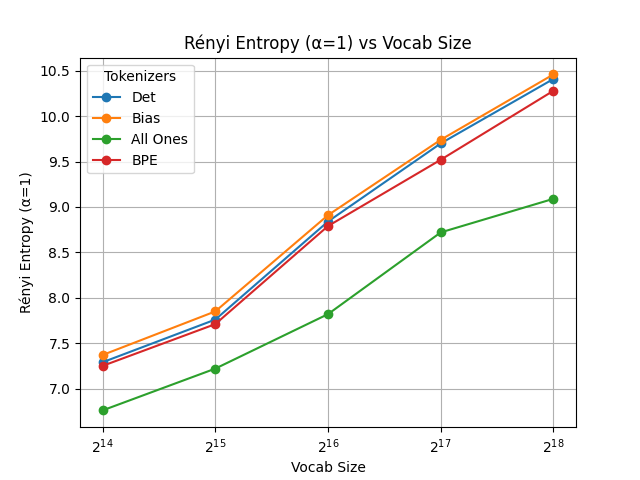}%
    \end{subfigure}\hfill
        \begin{subfigure}[b]{0.4\textwidth}
        \centering
        \includegraphics[width=\linewidth]{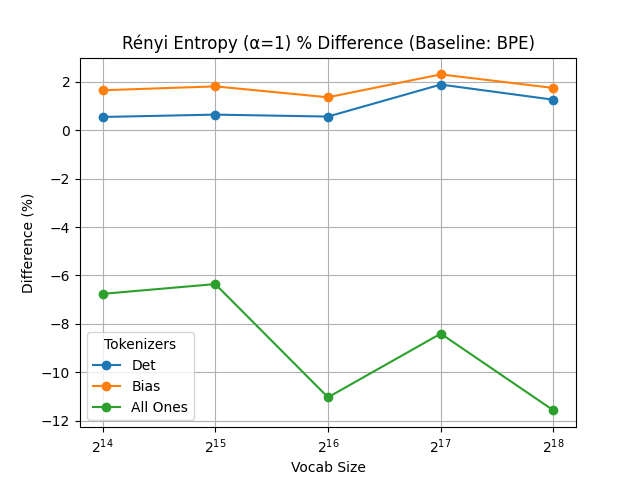}%
    \end{subfigure}
    \caption{Shannon entropy of different tokenisers. (Left) absolute and (Right) relative to \bpe.}
\end{figure}

\begin{figure}[H]
    \begin{subfigure}[b]{0.4\textwidth}
        \centering
        \includegraphics[width=\linewidth]{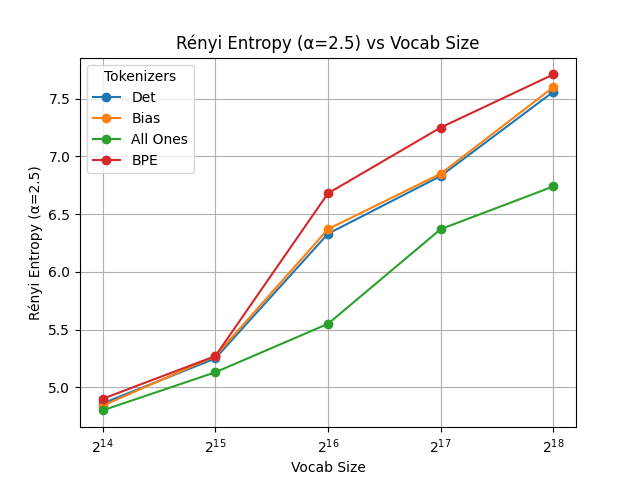}%
    \end{subfigure}\hfill
        \begin{subfigure}[b]{0.4\textwidth}
        \centering
        \includegraphics[width=\linewidth]{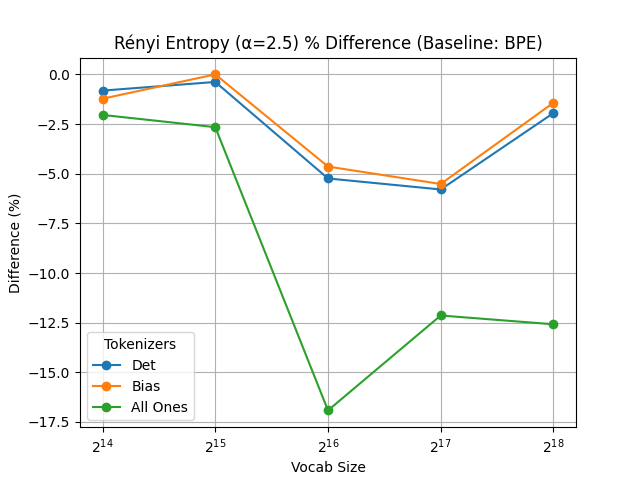}%
    \end{subfigure}
    \caption{R\'enyi entropy of different tokenisers. (Left) absolute and (Right) relative to \bpe.}
\end{figure}

\begin{figure}[H]
    \centering
    \begin{subfigure}[b]{0.4\textwidth}
        \centering
        \includegraphics[width=\linewidth]{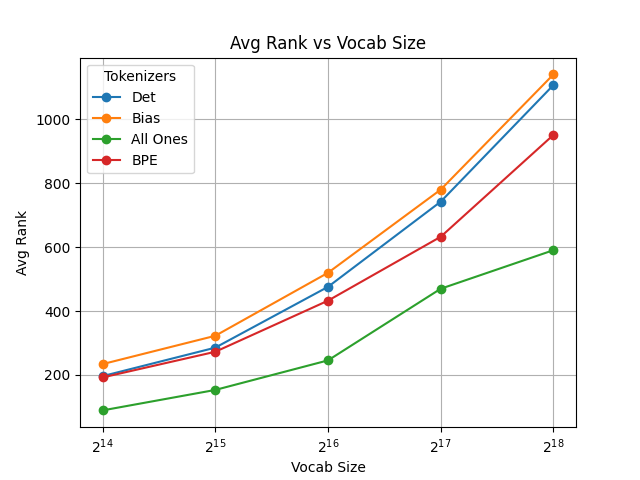}%
    \end{subfigure}\hfill
    \begin{subfigure}[b]{0.4\textwidth}
        \centering
        \includegraphics[width=\linewidth]{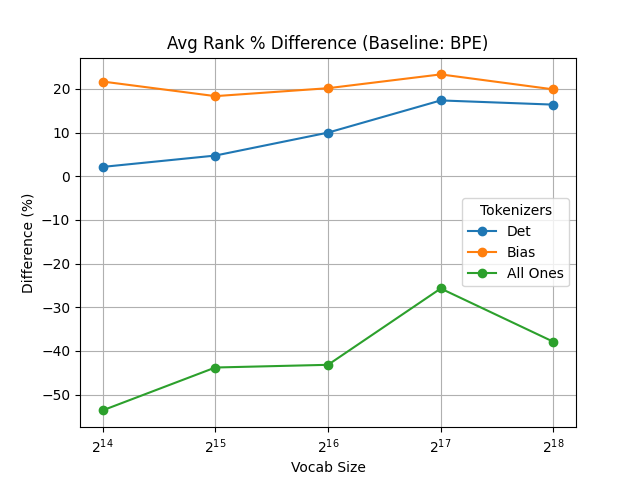}%
    \end{subfigure}
    \caption{Average rank of different tokenisers. (Left) absolute and (Right) relative to \bpe.}
\end{figure}

\section{Plots for Intrinsic Results on a Subset of the Tokenisers' Training Data}
\label{app:nlp_plots_in_dist}

\begin{figure}[H]
    \begin{subfigure}[b]{0.4\textwidth}
        \centering
        \includegraphics[width=\linewidth]{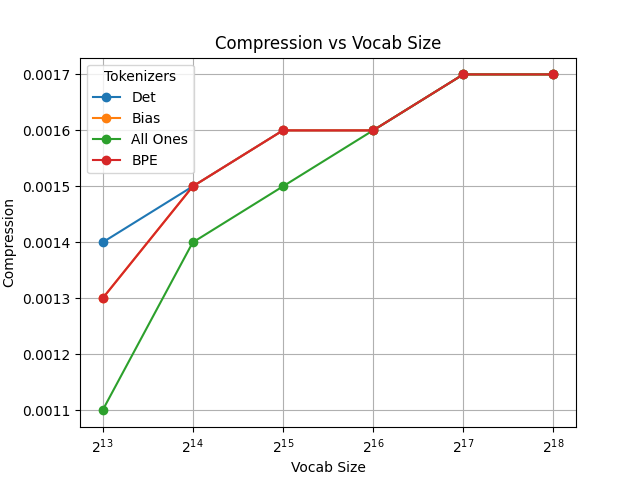}%
    \end{subfigure}\hfill
        \begin{subfigure}[b]{0.4\textwidth}
        \centering
        \includegraphics[width=\linewidth]{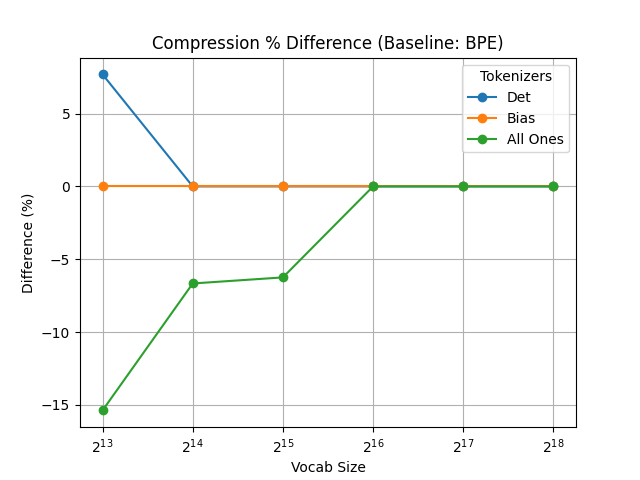}%
    \end{subfigure}\hfill
      \caption{
      Compression by the different tokenisers. (Left) absolute and (Right) relative to \bpe.}
\end{figure}

\begin{figure}[H]
\centering
\begin{subfigure}[b]{0.4\textwidth}
\centering
\includegraphics[width=\linewidth]{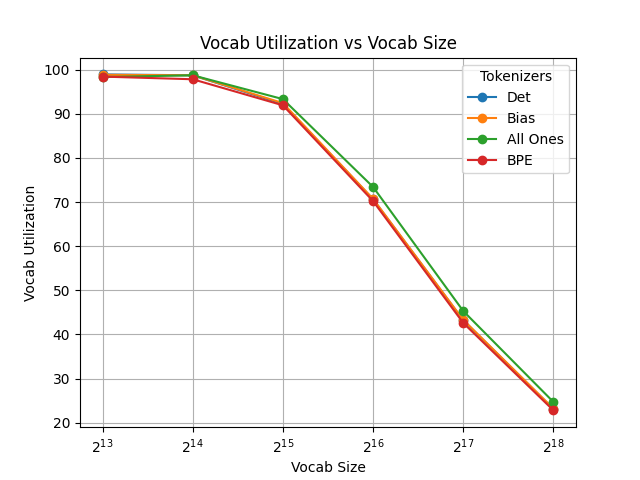}%
\end{subfigure}\hfill
    \begin{subfigure}[b]{0.4\textwidth}
        \centering
        \includegraphics[width=\linewidth]{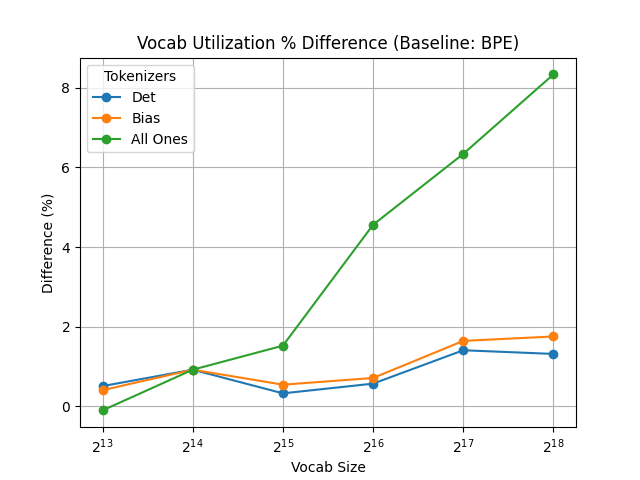}%
    \end{subfigure}
    \caption{Vocabulary utilisation by different tokenisers. (Left) absolute and (Right) relative to \bpe.}
\end{figure}

\begin{figure}[H]
    \begin{subfigure}[b]{0.4\textwidth}
        \centering
        \includegraphics[width=\linewidth]{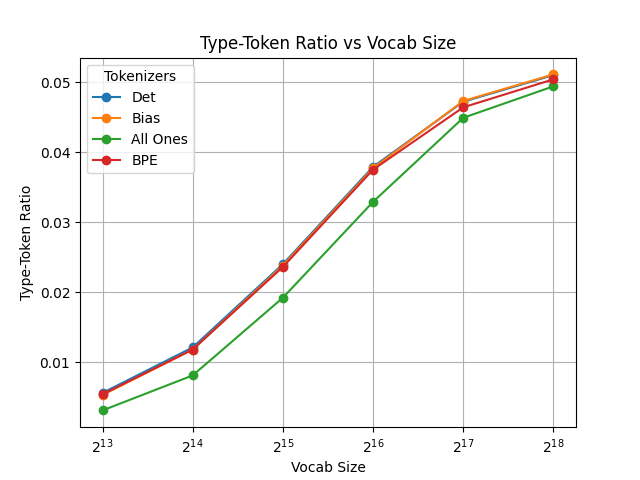}%
    \end{subfigure}\hfill
        \begin{subfigure}[b]{0.4\textwidth}
        \centering
        \includegraphics[width=\linewidth]{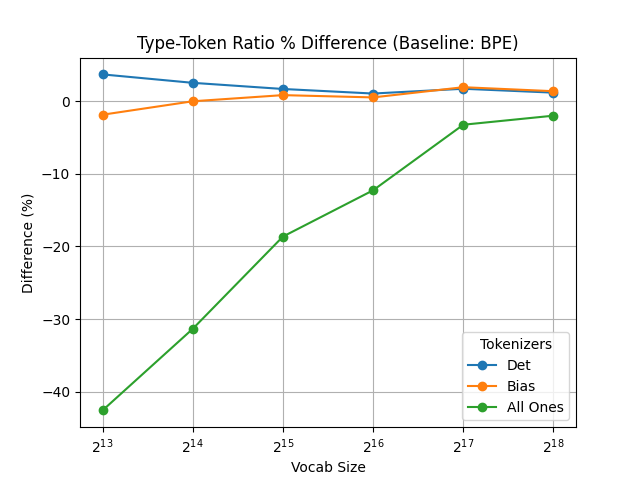}%
    \end{subfigure}
    \caption{Type-token ratio by different tokenisers. (Left) absolute and (Right) relative to \bpe.}
\end{figure}

\begin{figure}[H]
    \centering
    \begin{subfigure}[b]{0.4\textwidth}
        \centering
        \includegraphics[width=\linewidth]{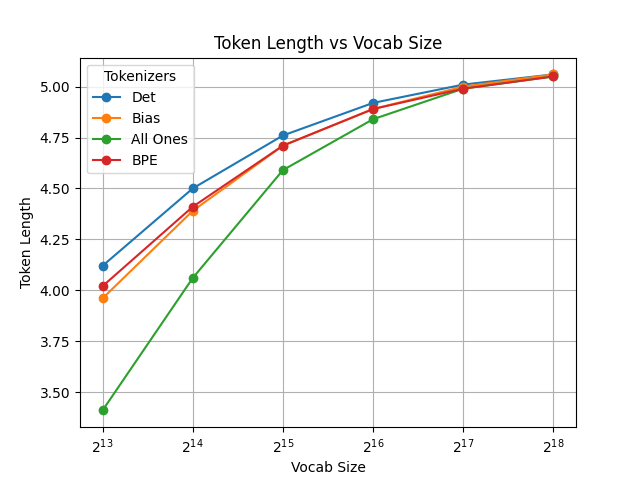}%
    \end{subfigure}\hfill
    \begin{subfigure}[b]{0.4\textwidth}
        \centering
        \includegraphics[width=\linewidth]{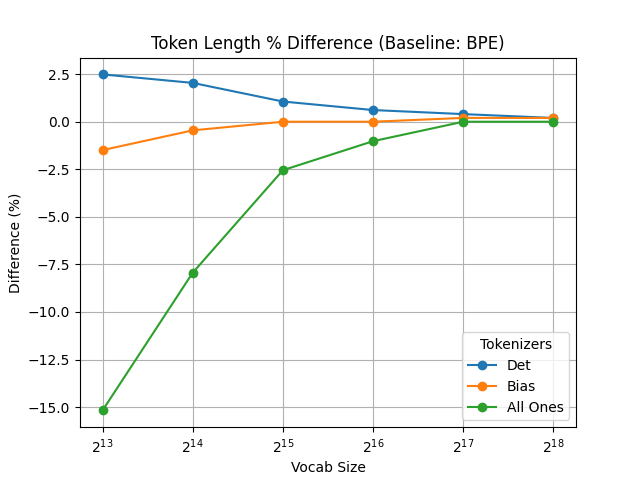}%
    \end{subfigure}
    \caption{Token length of different tokenisers. (Left) absolute and (Right) relative to \bpe.}
\end{figure}

\begin{figure}[H]
    \centering
    \begin{subfigure}[b]{0.4\textwidth}
        \centering
        \includegraphics[width=\linewidth]{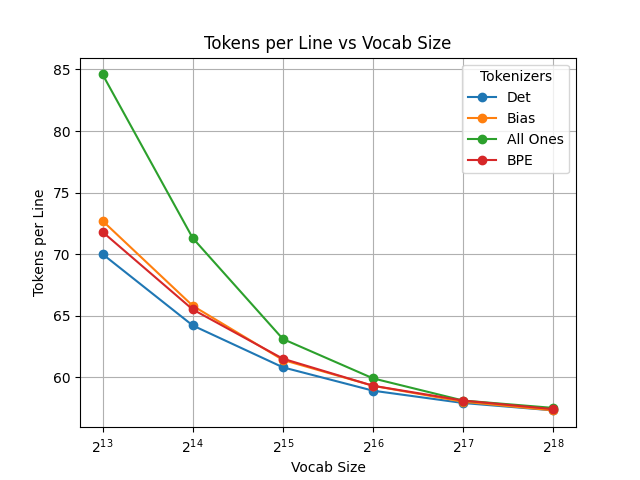}%
    \end{subfigure}\hfill
    \begin{subfigure}[b]{0.4\textwidth}
        \centering
        \includegraphics[width=\linewidth]{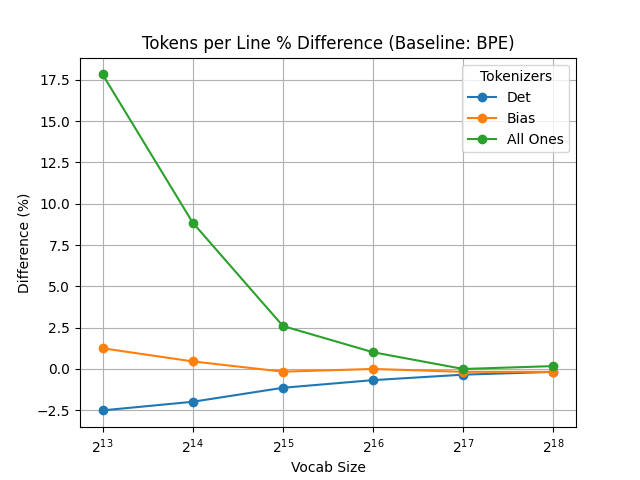}%
    \end{subfigure}
    \caption{Tokens per line of different tokenisers. (Left) absolute and (Right) relative to \bpe.}
\end{figure}

\begin{figure}[H]
    \begin{subfigure}[b]{0.4\textwidth}
        \centering
        \includegraphics[width=\linewidth]{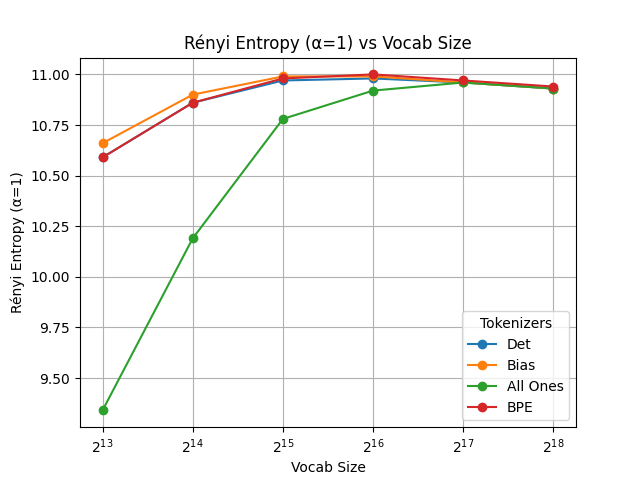}%
    \end{subfigure}\hfill
        \begin{subfigure}[b]{0.4\textwidth}
        \centering
        \includegraphics[width=\linewidth]{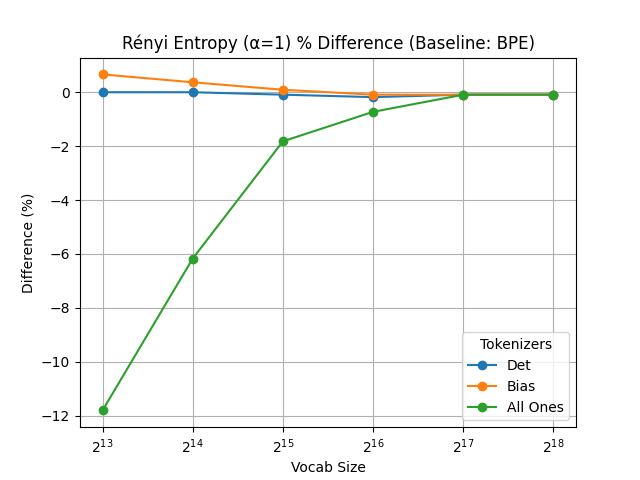}%
    \end{subfigure}
    \caption{Shannon entropy of different tokenisers. (Left) absolute and (Right) relative to \bpe.}
\end{figure}

\begin{figure}[H]
    \begin{subfigure}[b]{0.4\textwidth}
        \centering
        \includegraphics[width=\linewidth]{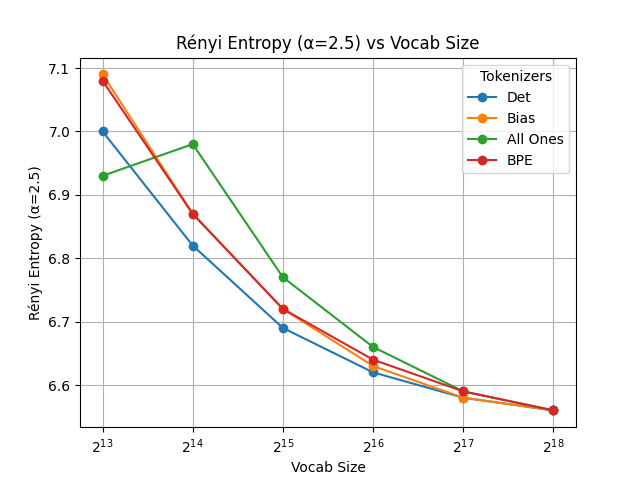}%
    \end{subfigure}\hfill
        \begin{subfigure}[b]{0.4\textwidth}
        \centering
        \includegraphics[width=\linewidth]{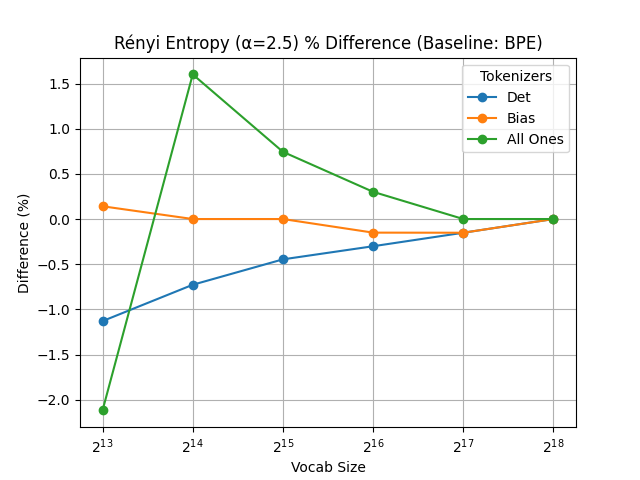}%
    \end{subfigure}
    \caption{R\'enyi entropy of different tokenisers. (Left) absolute and (Right) relative to \bpe.}
\end{figure}

\begin{figure}[H]
    \centering
    \begin{subfigure}[b]{0.4\textwidth}
        \centering
        \includegraphics[width=\linewidth]{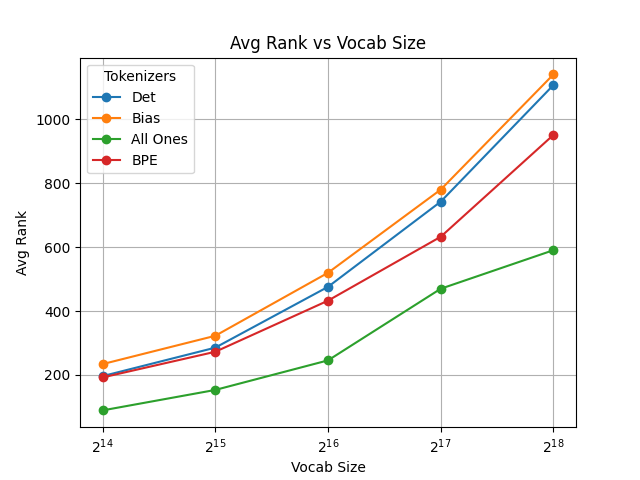}%
    \end{subfigure}\hfill
    \begin{subfigure}[b]{0.4\textwidth}
        \centering
        \includegraphics[width=\linewidth]{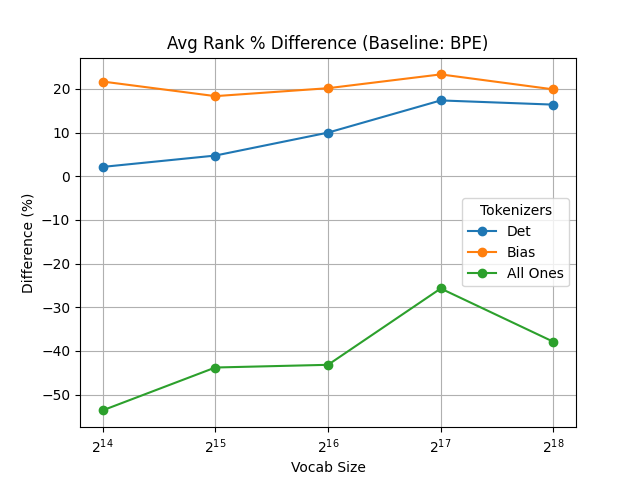}%
    \end{subfigure}
    \caption{Average rank of different tokenisers. (Left) absolute and (Right) relative to \bpe.}
\end{figure}

\section{Plots for Intrinsic Results on a Held-out Set of ClimbMix}
\label{app:nlp_plots_out_dist}

\begin{figure}[H]
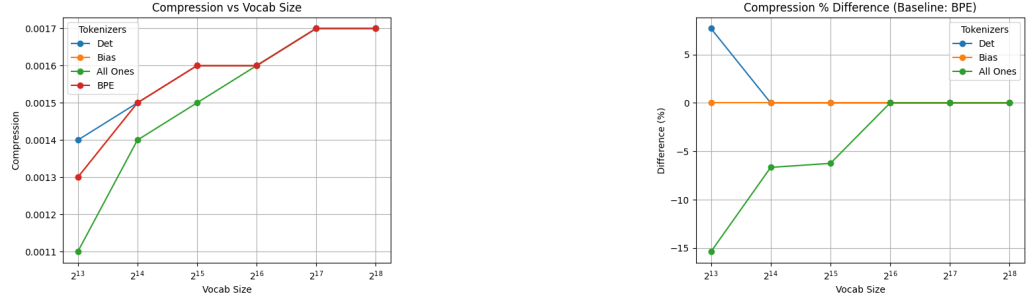

    \begin{subfigure}[b]{0.4\textwidth}
        \centering
        \includegraphics[width=\linewidth]{figs/out_dist/Compression_absolute.png}%
    \end{subfigure}\hfill
        \begin{subfigure}[b]{0.4\textwidth}
        \centering
        \includegraphics[width=\linewidth]{figs/out_dist/Compression_pct_diff_all_vs_BPE.png}%
    \end{subfigure}\hfill
      \caption{
      Compression by the different tokenisers. (Left) absolute and (Right) relative to \bpe.}
\end{figure}

\begin{figure}[H]
\centering
\begin{subfigure}[b]{0.4\textwidth}
\centering
\includegraphics[width=\linewidth]{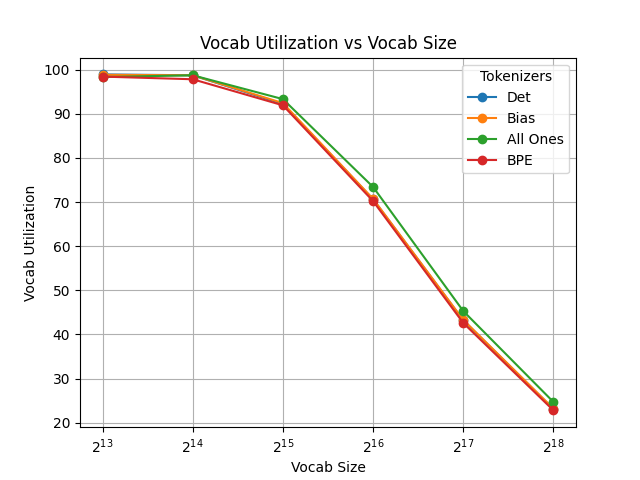}%
\end{subfigure}\hfill
    \begin{subfigure}[b]{0.4\textwidth}
        \centering
        \includegraphics[width=\linewidth]{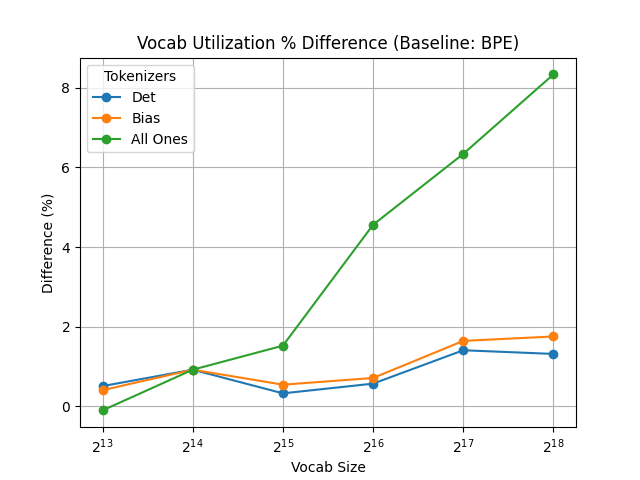}%
    \end{subfigure}
    \caption{Vocabulary utilisation by different tokenisers. (Left) absolute and (Right) relative to \bpe.}
\end{figure}

\begin{figure}[H]
    \begin{subfigure}[b]{0.4\textwidth}
        \centering
        \includegraphics[width=\linewidth]{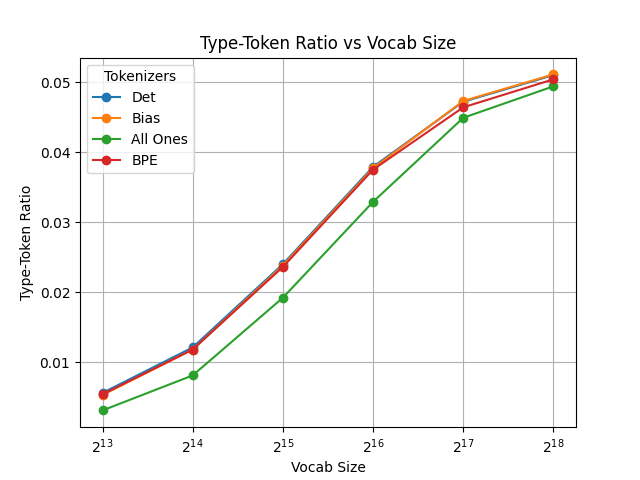}%
    \end{subfigure}\hfill
        \begin{subfigure}[b]{0.4\textwidth}
        \centering
        \includegraphics[width=\linewidth]{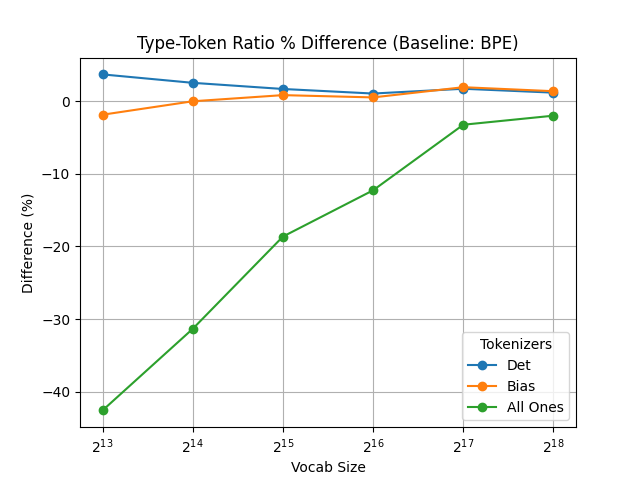}%
    \end{subfigure}
    \caption{Type-token ratio by different tokenisers. (Left) absolute and (Right) relative to \bpe.}
\end{figure}

\begin{figure}[H]
    \centering
    \begin{subfigure}[b]{0.4\textwidth}
        \centering
        \includegraphics[width=\linewidth]{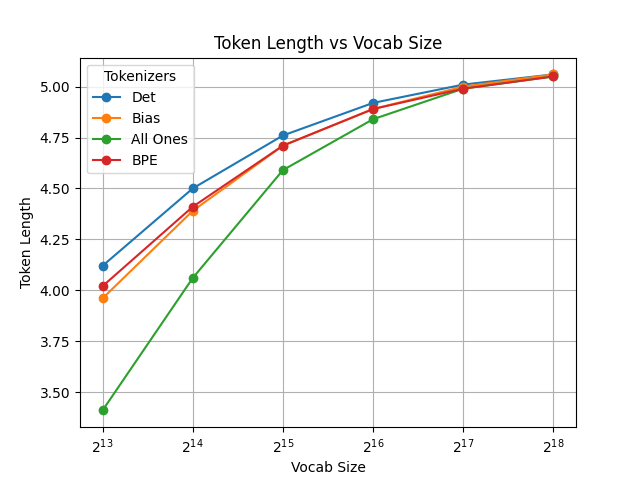}%
    \end{subfigure}\hfill
    \begin{subfigure}[b]{0.4\textwidth}
        \centering
        \includegraphics[width=\linewidth]{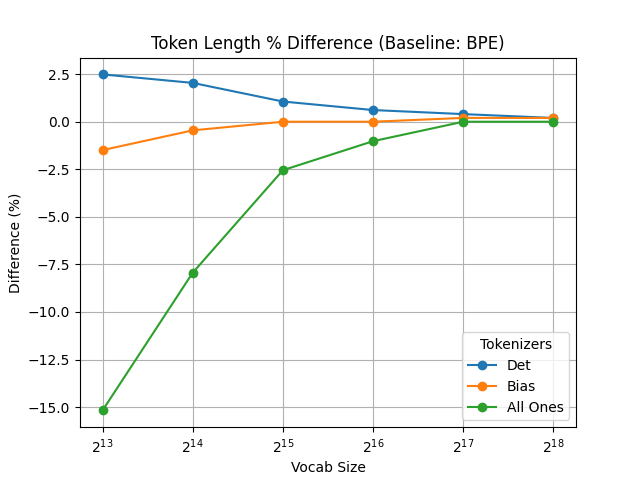}%
    \end{subfigure}
    \caption{Token length of different tokenisers. (Left) absolute and (Right) relative to \bpe.}
\end{figure}

\begin{figure}[H]
    \centering
    \begin{subfigure}[b]{0.4\textwidth}
        \centering
        \includegraphics[width=\linewidth]{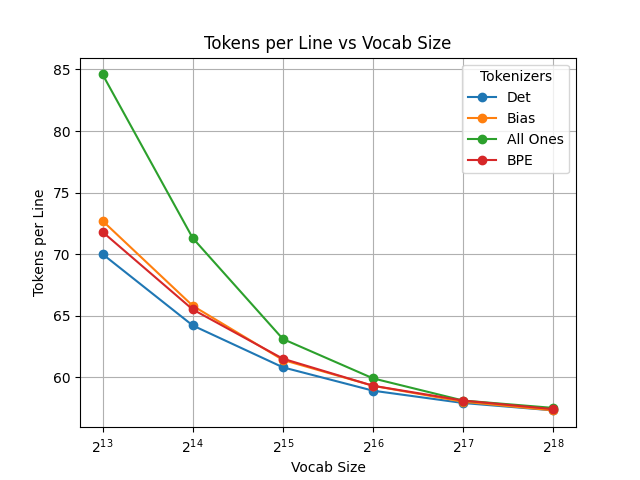}%
    \end{subfigure}\hfill
    \begin{subfigure}[b]{0.4\textwidth}
        \centering
        \includegraphics[width=\linewidth]{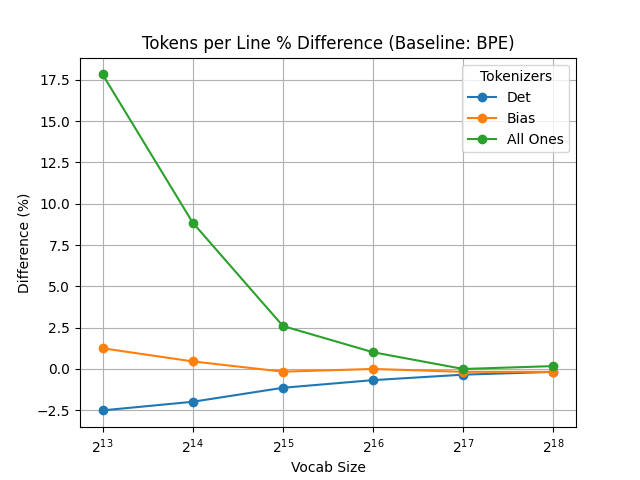}%
    \end{subfigure}
    \caption{Tokens per line of different tokenisers. (Left) absolute and (Right) relative to \bpe.}
\end{figure}

\begin{figure}[H]
    \begin{subfigure}[b]{0.4\textwidth}
        \centering
        \includegraphics[width=\linewidth]{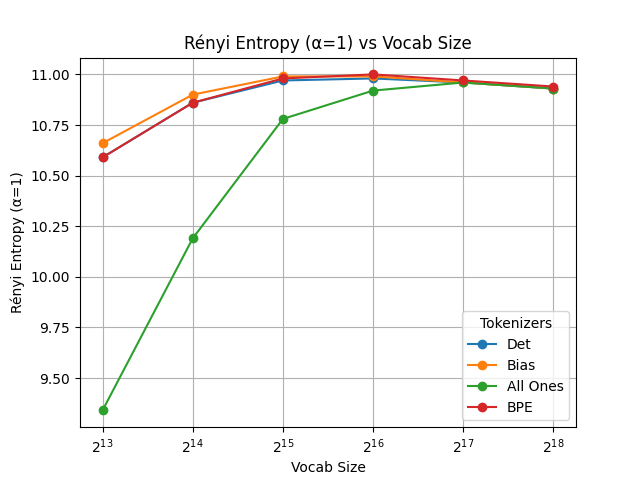}%
    \end{subfigure}\hfill
        \begin{subfigure}[b]{0.4\textwidth}
        \centering
        \includegraphics[width=\linewidth]{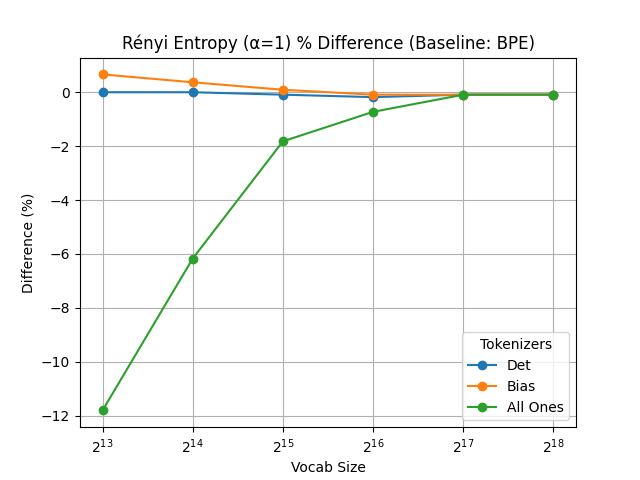}%
    \end{subfigure}
    \caption{Shannon entropy of different tokenisers. (Left) absolute and (Right) relative to \bpe.}
\end{figure}

\begin{figure}[H]
    \begin{subfigure}[b]{0.4\textwidth}
        \centering
        \includegraphics[width=\linewidth]{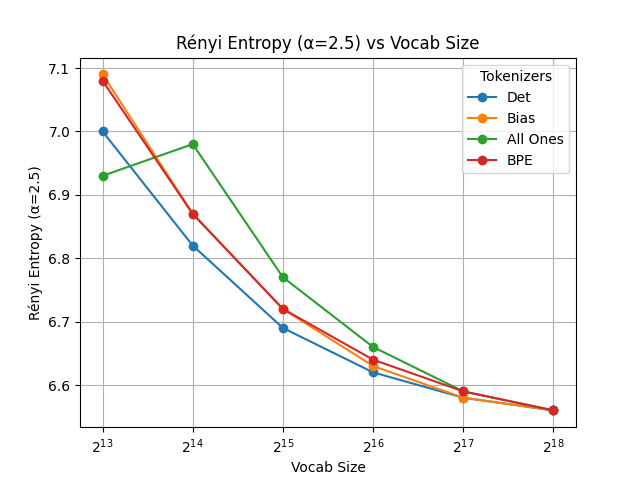}%
    \end{subfigure}\hfill
        \begin{subfigure}[b]{0.4\textwidth}
        \centering
        \includegraphics[width=\linewidth]{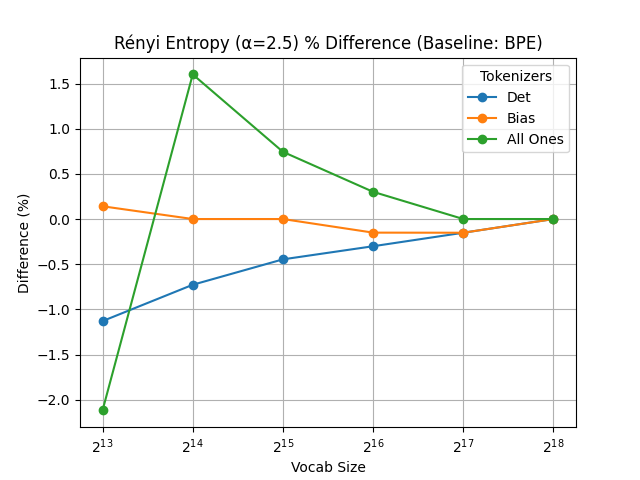}%
    \end{subfigure}
    \caption{R\'enyi entropy of different tokenisers. (Left) absolute and (Right) relative to \bpe.}
\end{figure}

\begin{figure}[H]
    \centering
    \begin{subfigure}[b]{0.4\textwidth}
        \centering
        \includegraphics[width=\linewidth]{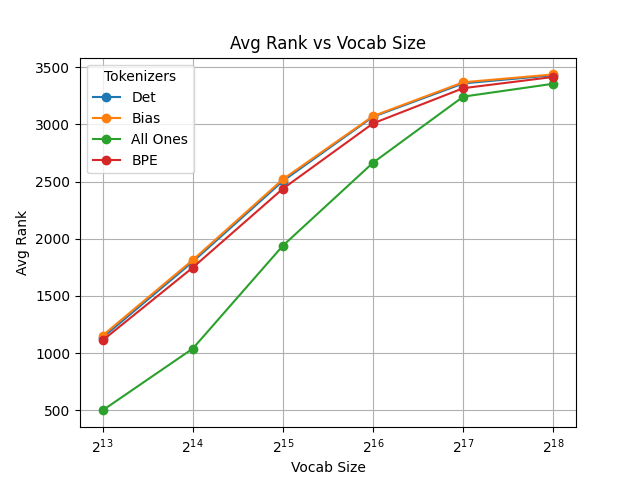}%
    \end{subfigure}\hfill
    \begin{subfigure}[b]{0.4\textwidth}
        \centering
        \includegraphics[width=\linewidth]{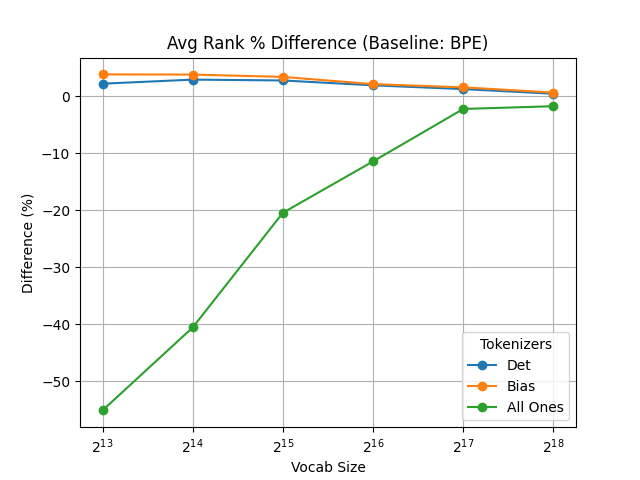}%
    \end{subfigure}
    \caption{Average rank of different tokenisers. (Left) absolute and (Right) relative to \bpe.}
\end{figure}

\section{Loss across Training for the Various Models}\label{app:plots_training}

\begin{figure}[H]
    \begin{subfigure}[b]{0.48\textwidth}
        \centering
        \includegraphics[width=\linewidth]{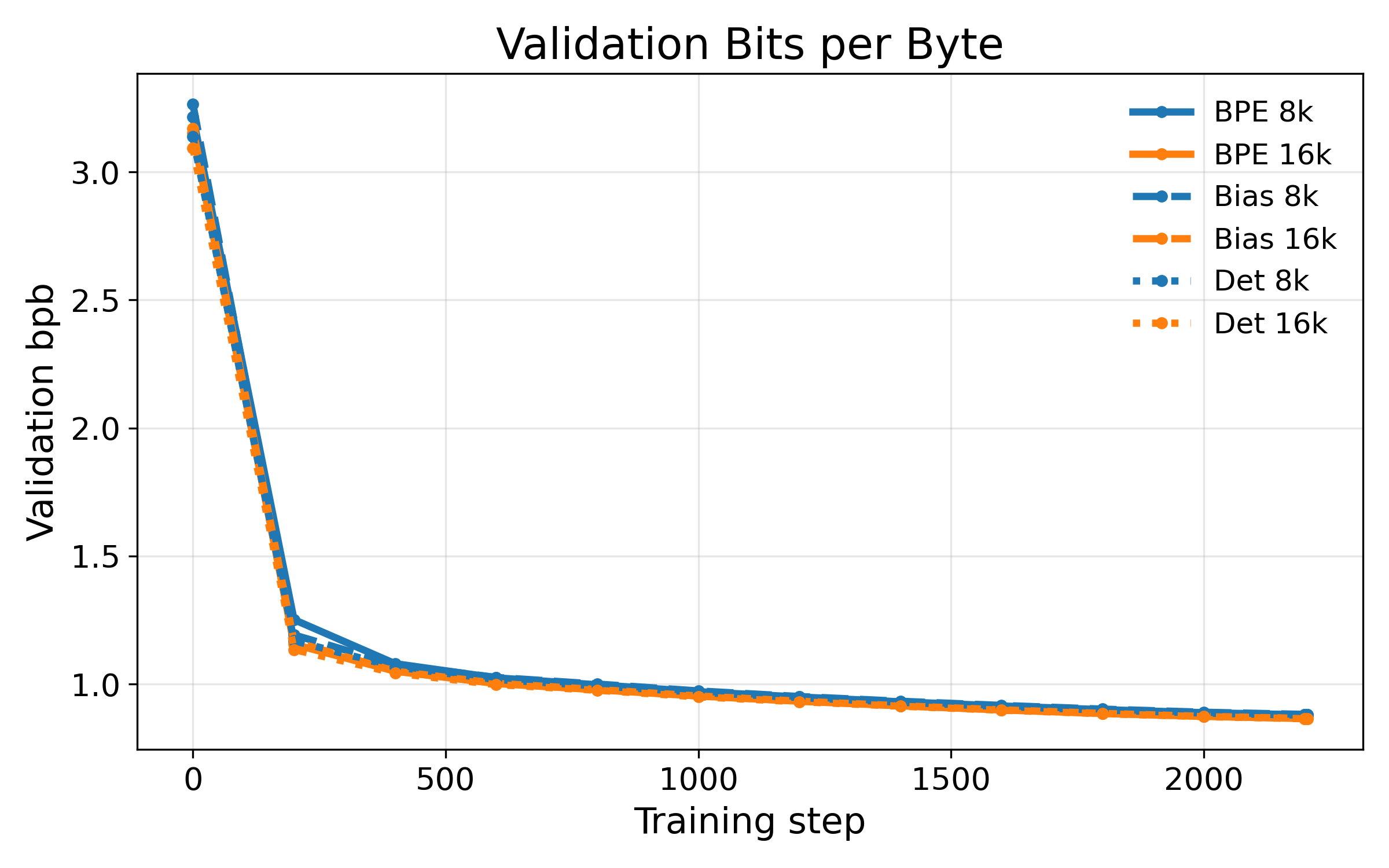}
        \caption{Val \bpb for Depth 12 $8k$ and $16k$}
    \end{subfigure}\hfill
        \begin{subfigure}[b]{0.48\textwidth}
        \centering
        \includegraphics[width=\linewidth]{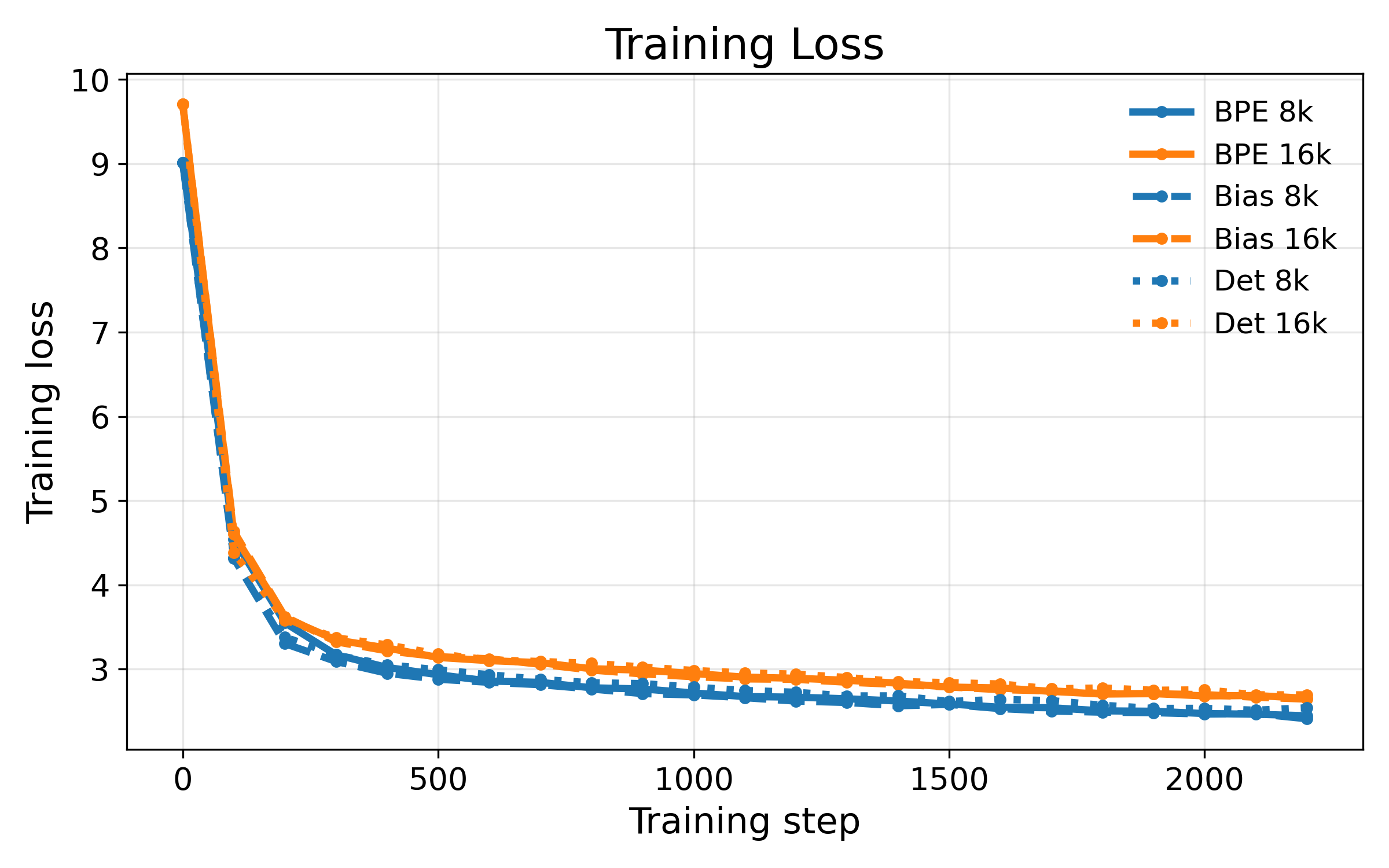}
          \caption{Train loss for Depth 12 $8k$, and $16k$}
    \end{subfigure}\hfill
\end{figure}

\begin{figure}[H]
    \begin{subfigure}[b]{0.48\textwidth}
        \centering
        \includegraphics[width=\linewidth]{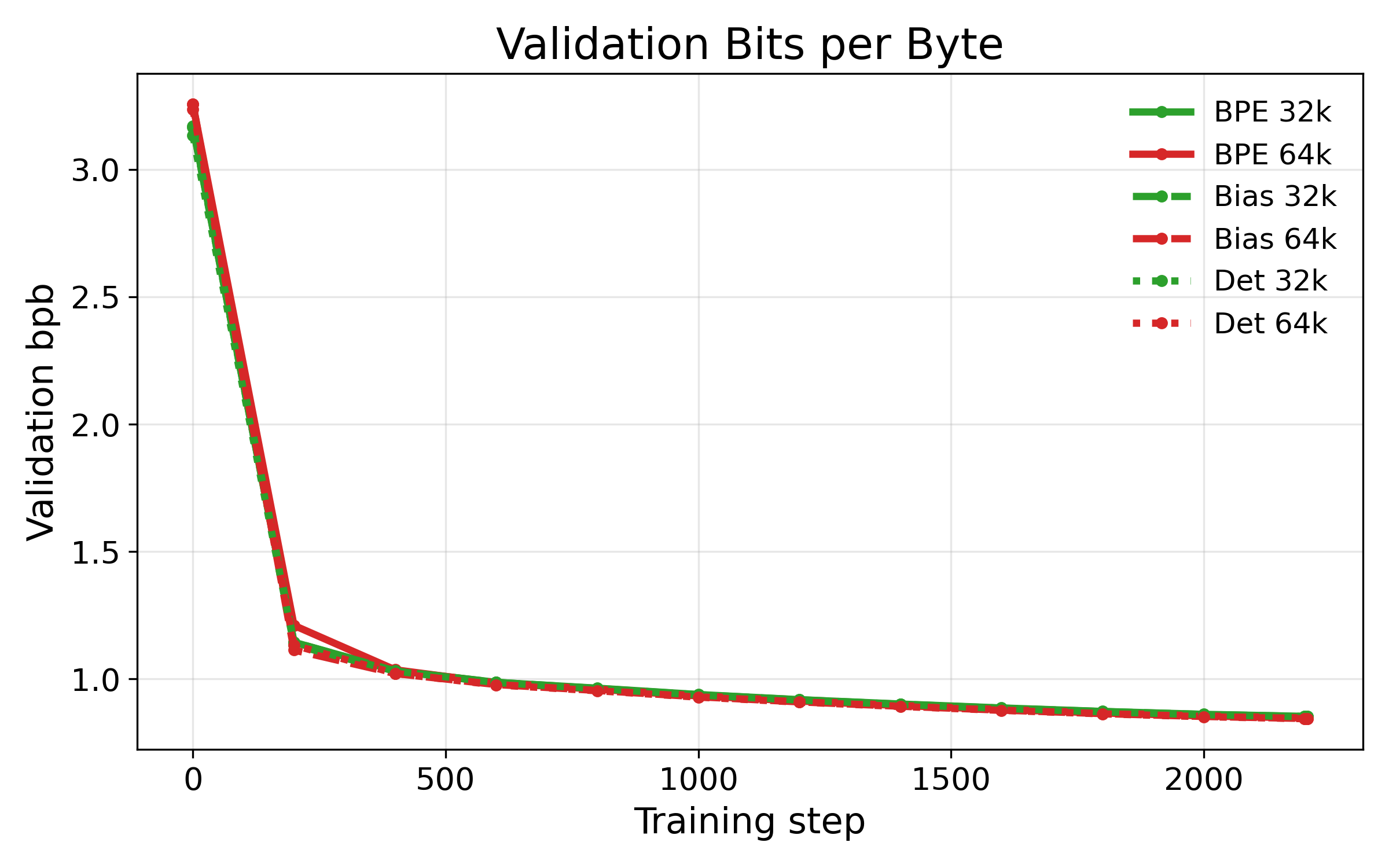}
        \caption{Val \bpb for Depth 12 $32k$, and $64k$}
    \end{subfigure}\hfill
\begin{subfigure}[b]{0.48\textwidth}
\centering
\includegraphics[width=\linewidth]{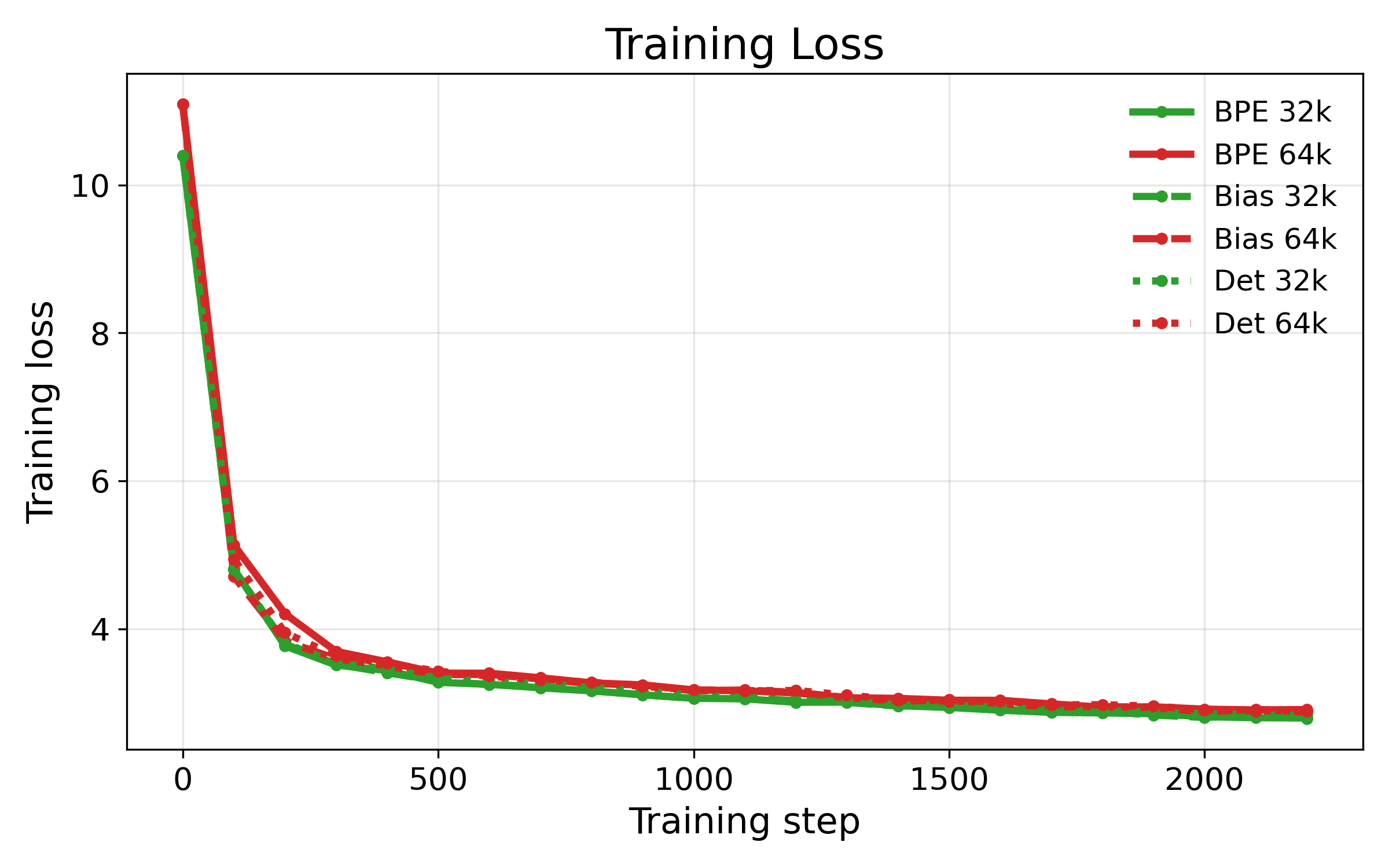}%
\caption{Train loss for Depth 12 $32k$ and $64k$}
\end{subfigure}\hfill
\end{figure}

\begin{figure}[H]
    \begin{subfigure}[b]{0.48\textwidth}
        \centering
        \includegraphics[width=\linewidth]{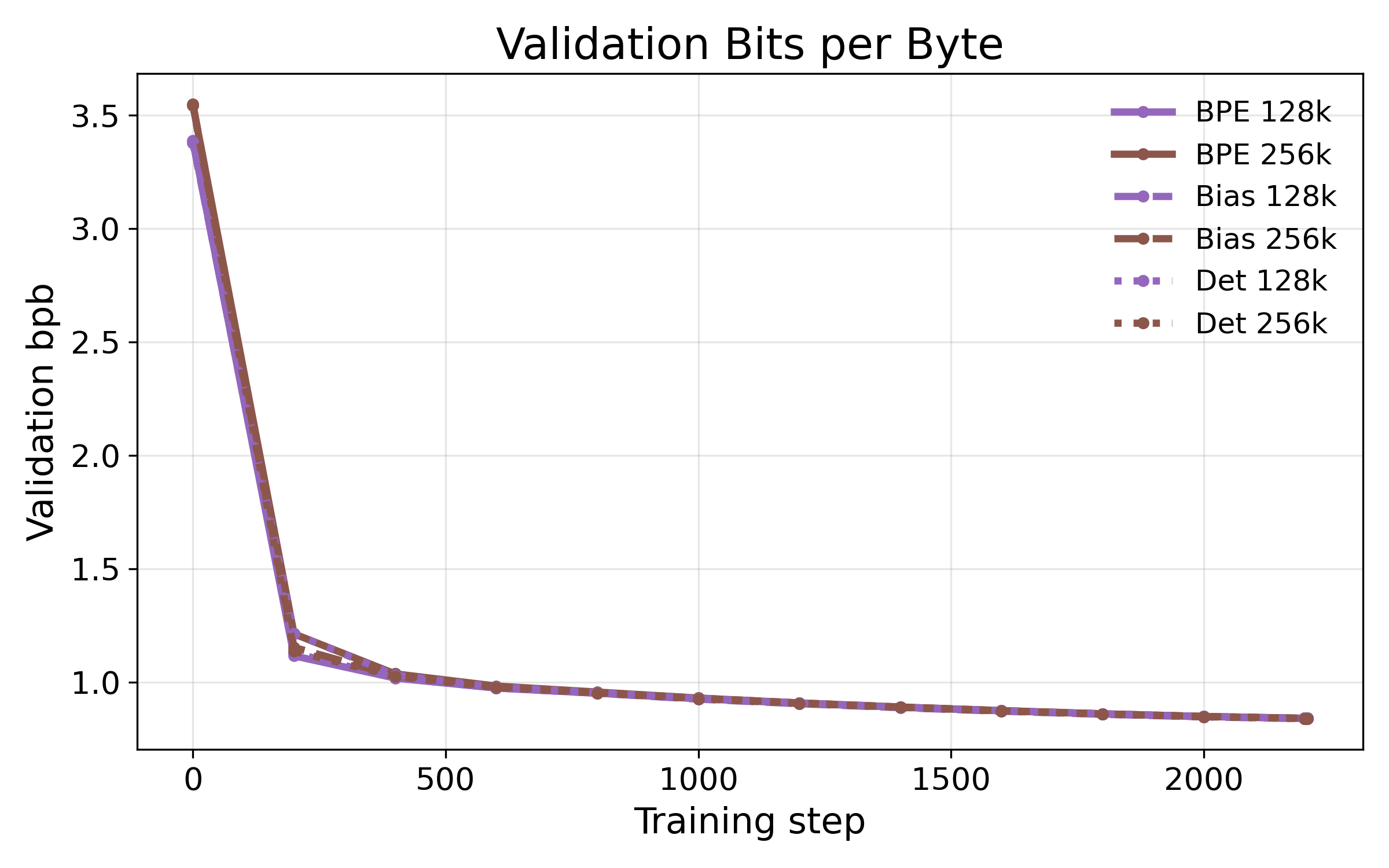}
        \caption{Val \bpb for Depth 12 $128k$, and $256k$}
    \end{subfigure}\hfill
        \begin{subfigure}[b]{0.48\textwidth}
        \centering
        \includegraphics[width=\linewidth]{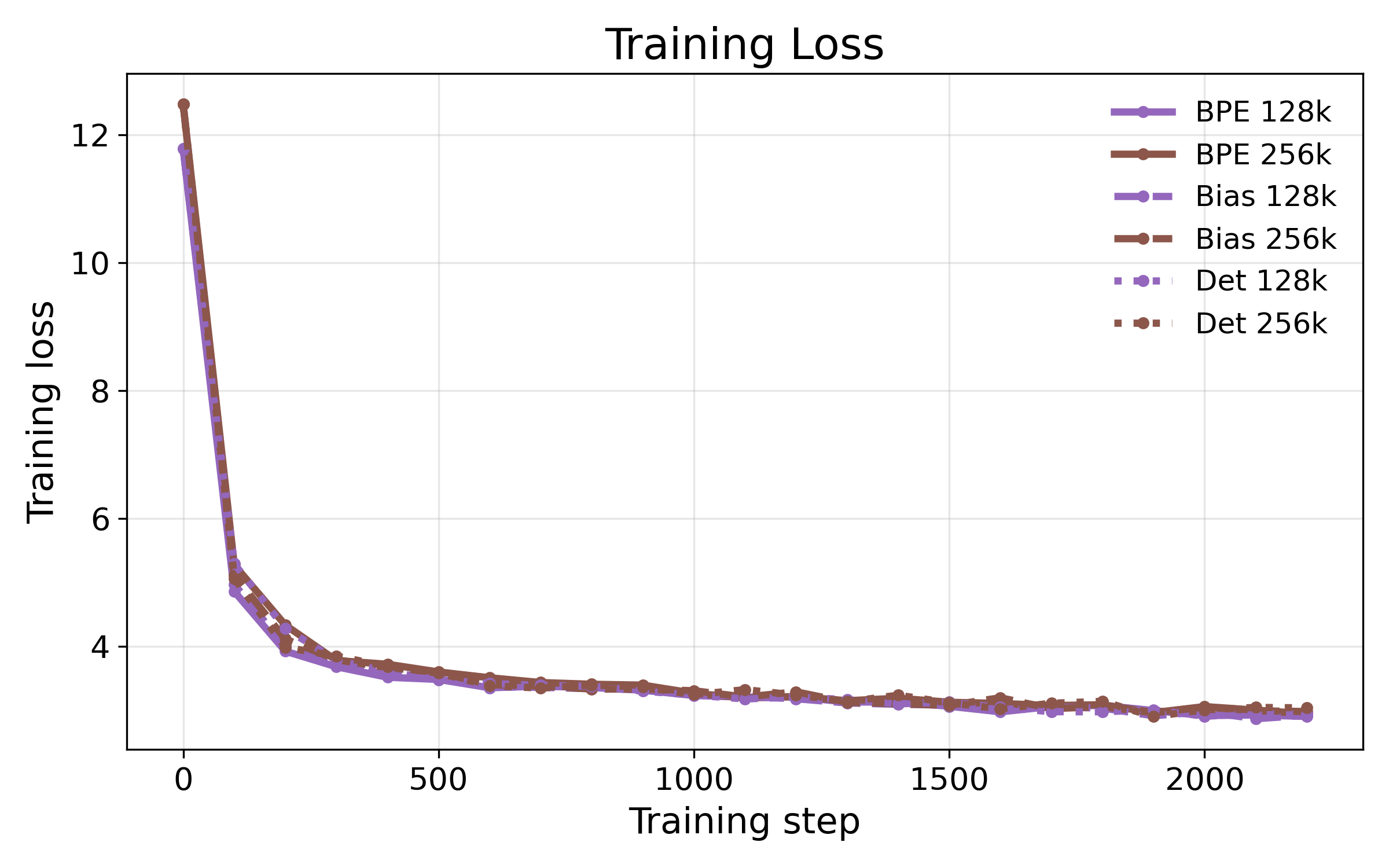}%
        \caption{Train loss for Depth 12 $128k$, and $256k$}
    \end{subfigure}\hfill
\end{figure}

\begin{figure}[H]
    \begin{subfigure}[b]{0.48\textwidth}
        \centering
        \includegraphics[width=\linewidth]{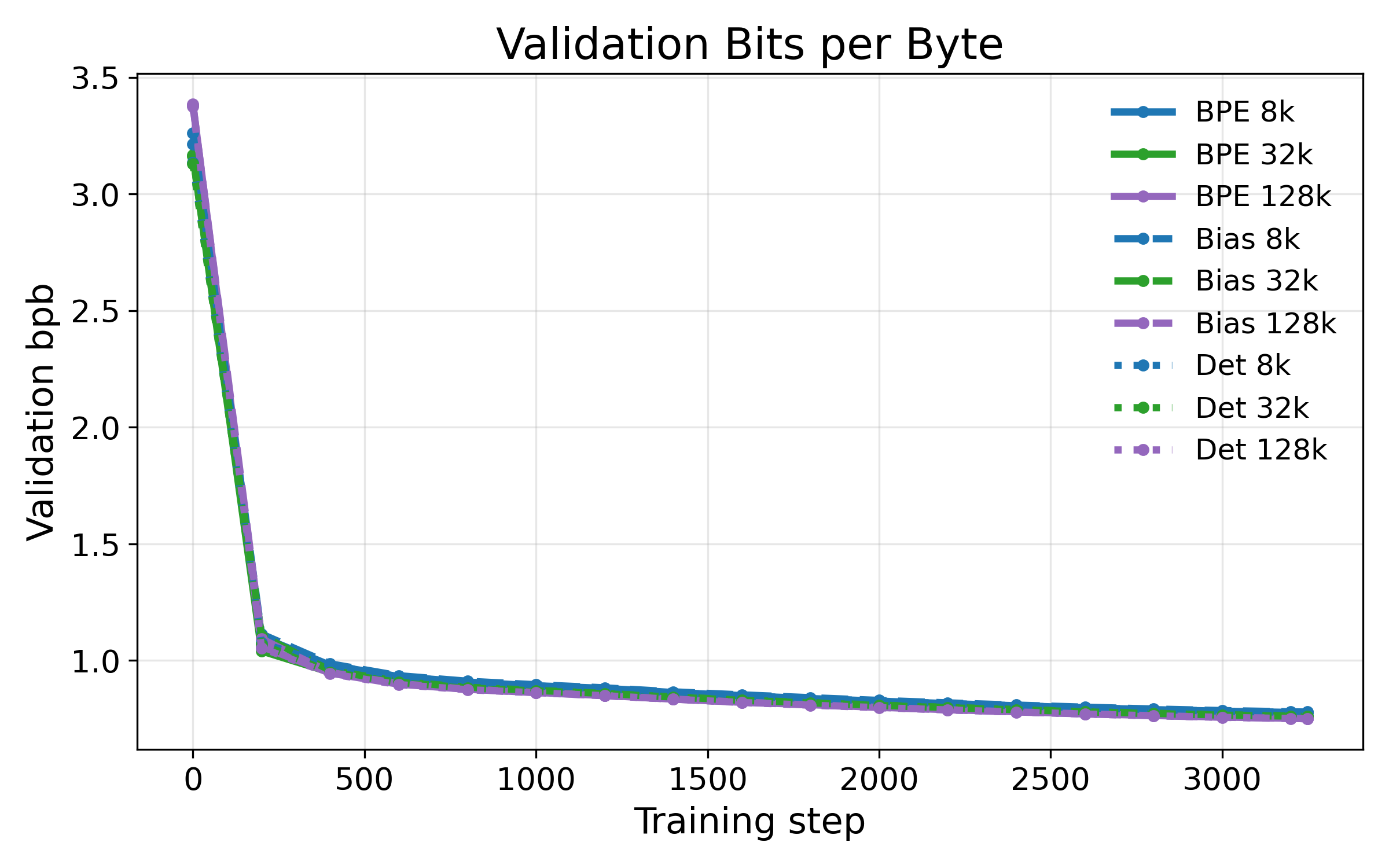}
        \caption{Val \bpb for Depth 18}
    \end{subfigure}\hfill
        \begin{subfigure}[b]{0.48\textwidth}
        \centering
        \includegraphics[width=\linewidth]{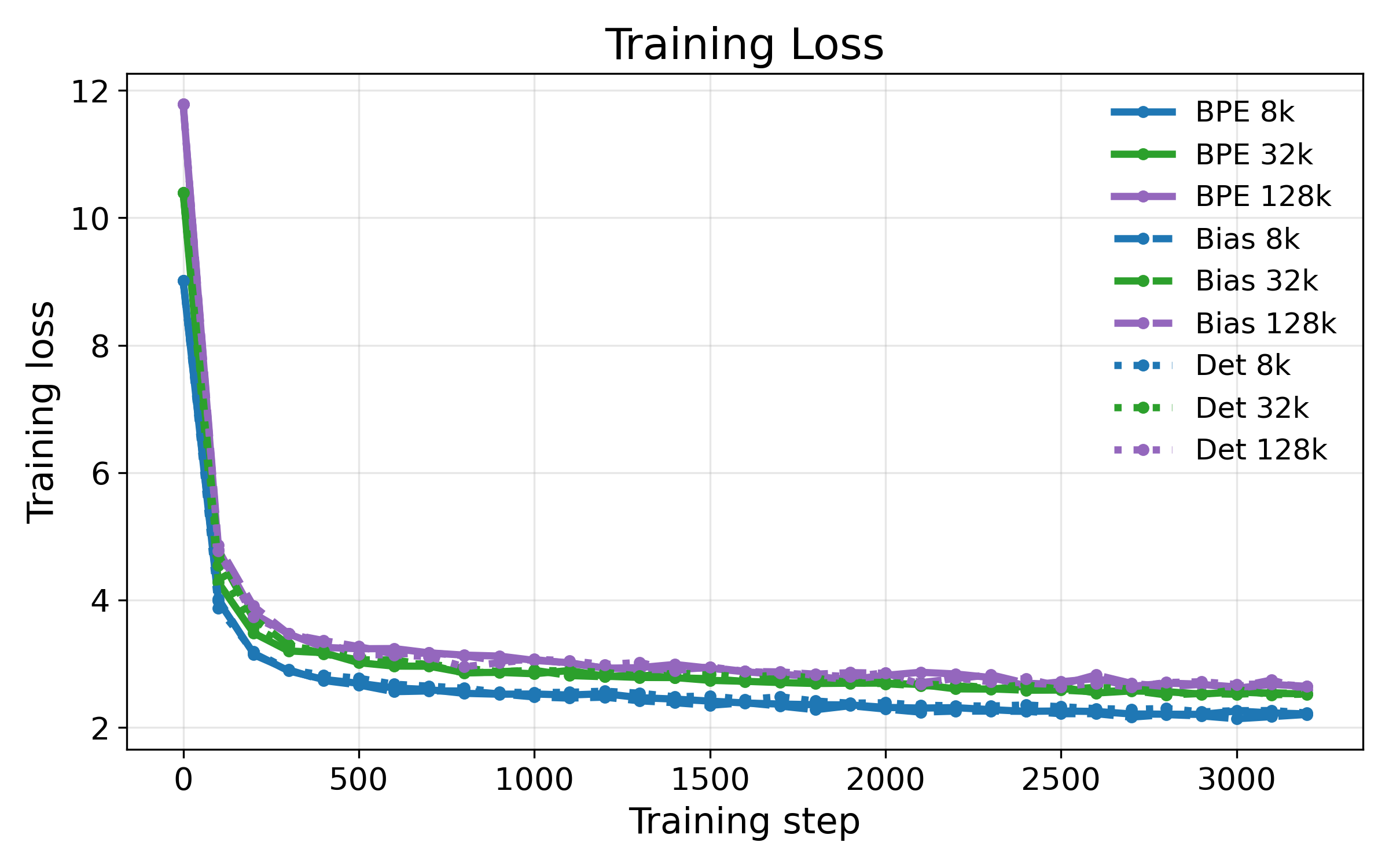}%
        \caption{Train loss for Depth 18}
    \end{subfigure}\hfill
\end{figure}

\begin{figure}[H]
    \begin{subfigure}[b]{0.48\textwidth}
        \centering
        \includegraphics[width=\linewidth]{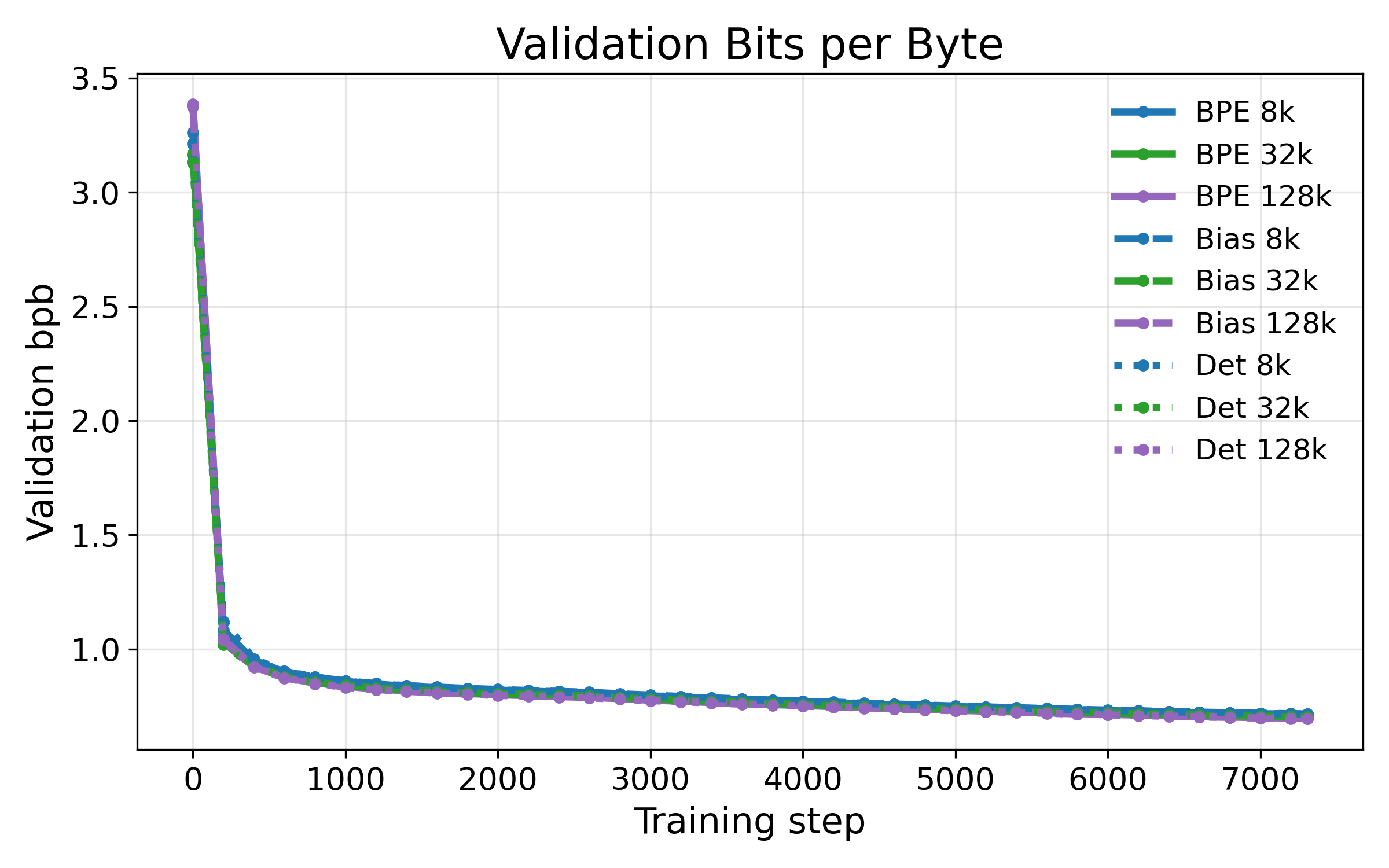}
        \caption{Val \bpb for depth 24}
    \end{subfigure}\hfill
        \begin{subfigure}[b]{0.48\textwidth}
        \centering
        \includegraphics[width=\linewidth]{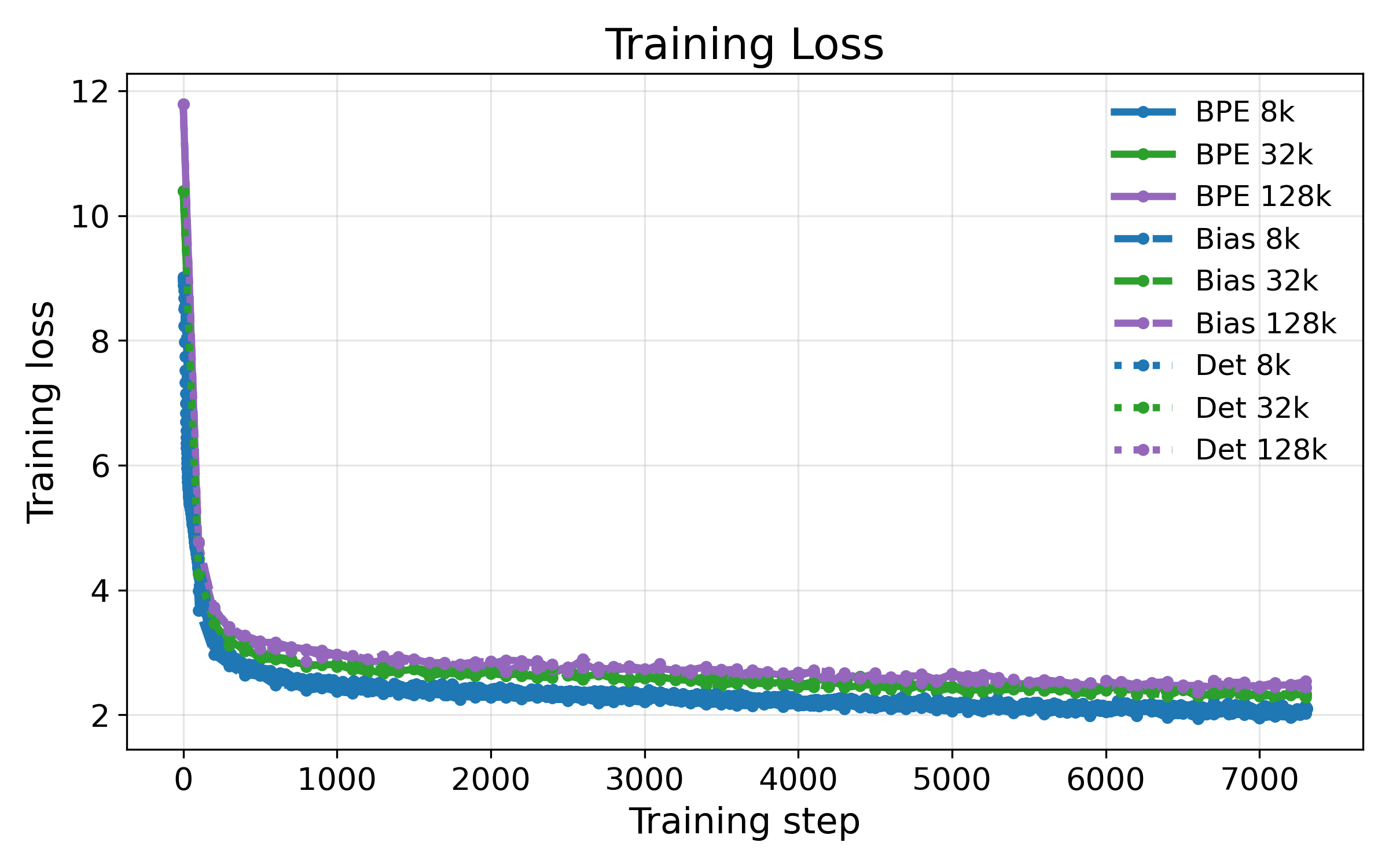}%
        \caption{Train loss for depth 24}
    \end{subfigure}\hfill
\end{figure}

\section{Detailed Results on Downstream Tasks}\label{app:plots_downstream}

\subsection{Plots for Robustness Experiments as well as for Various Depths }

\begin{figure}[H]
    \begin{subfigure}[b]{0.48\textwidth}
        \centering
         \includegraphics[width=\linewidth]{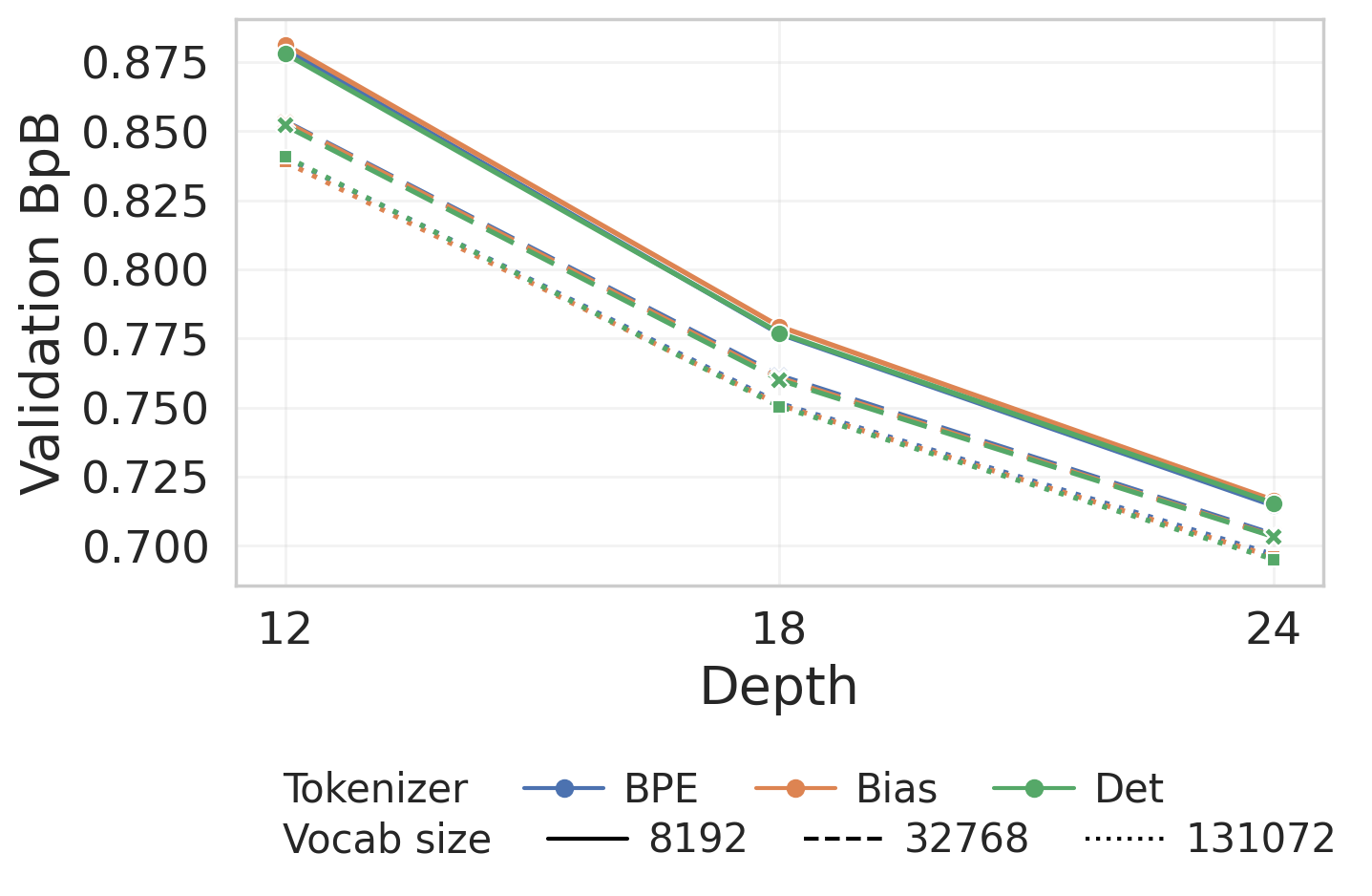}
    \end{subfigure}\hfill
        \begin{subfigure}[b]{0.48\textwidth}
        \centering
        \includegraphics[width=\linewidth]{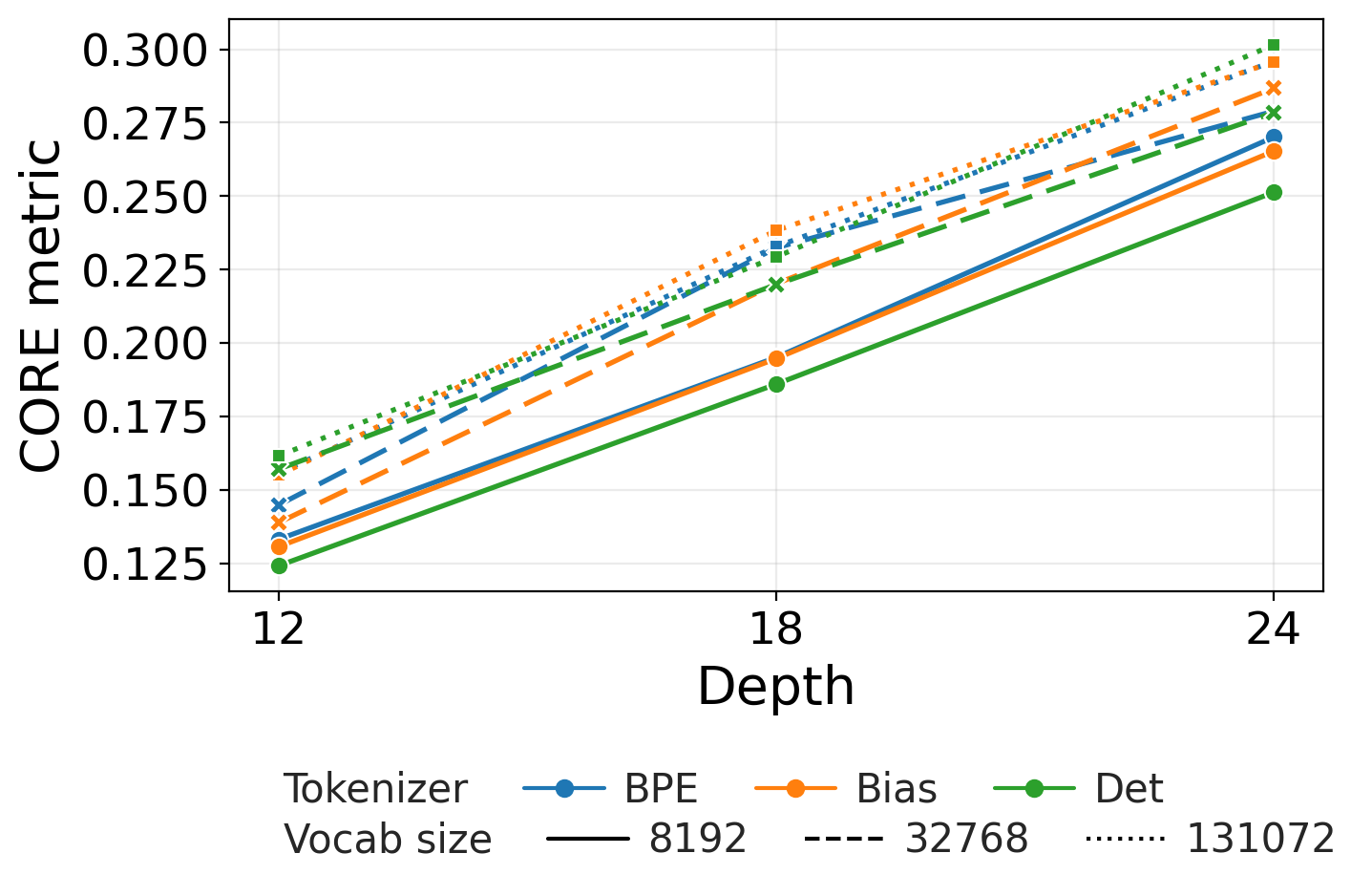}%
    \end{subfigure}
 \caption{(left) \bpb and (right) \core vs.\ vocabulary size across models with different depths.\looseness=-1}
\end{figure}

\begin{figure}
    \label{fig:core_depth_progression}
\end{figure}

\begin{figure}[H]
    \begin{subfigure}[b]{0.48\textwidth}
        \centering
        \includegraphics[width=\linewidth]{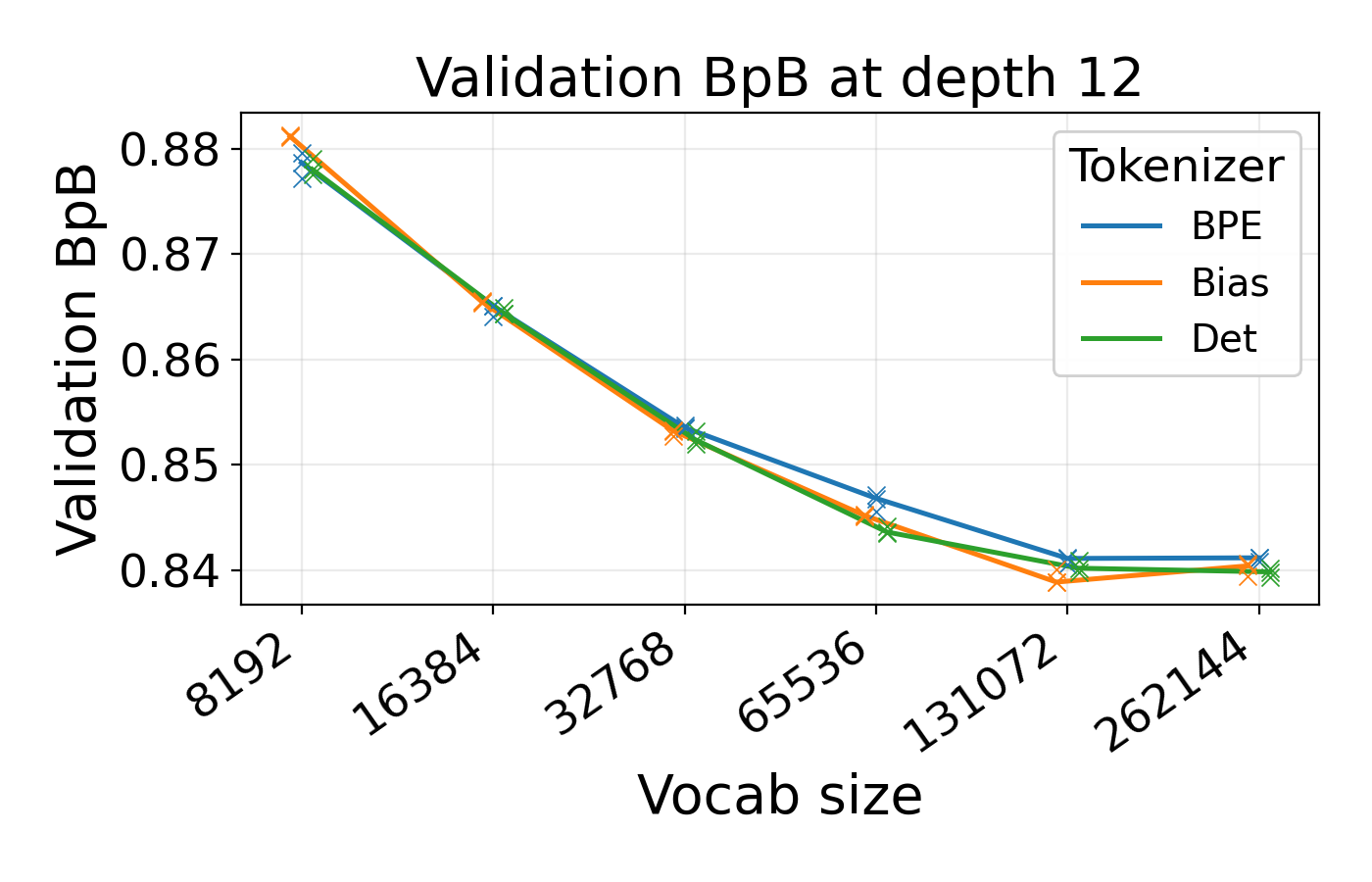}%
    \end{subfigure}\hfill
        \begin{subfigure}[b]{0.48\textwidth}
        \centering
        \includegraphics[width=\linewidth]{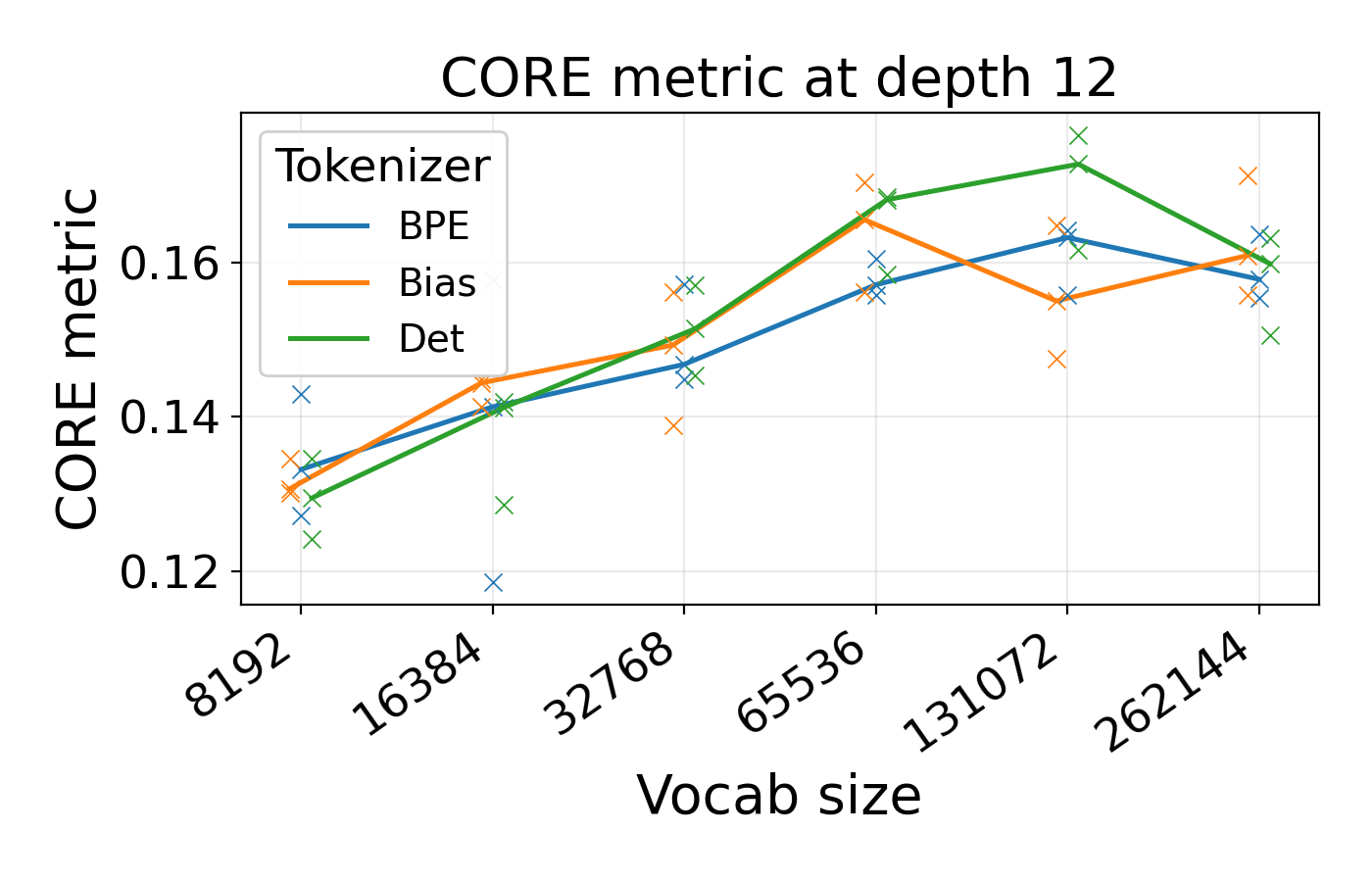}%
    \end{subfigure}
\caption{(left) \bpb and (right) \core vs.\ vocabulary size across three training seeds. All these models were trained with 12 layers.}
\end{figure}

\subsection{Depth 12}
\begin{table}[H]
\centering
\caption{Downstream Performance (1/3) --- Depth 12}
\label{tab:downstream_perf_1_d12}
\footnotesize
\resizebox{\textwidth}{!}{%
\begin{tabular}{llcccccccc}
\toprule
\makecell{Vocabulary \\ Size} & Tokeniser & \makecell{Hellaswag \\ Zeroshot} & Jeopardy & \makecell{BigBench \\ QA Wikidata} & \makecell{ARC \\ Easy} & \makecell{ARC \\ Challenge} & COPA & \makecell{Commonsense \\ QA} & PIQA \\
\midrule
\multirow{3}{*}{$8k$} & \bpetok & 0.3260 & 0.0120 & 0.1080 & 0.4540 & 0.2580 & 0.5500 & 0.2660 & 0.6760 \\
                      & \biastok & 0.3540 & 0.0080 & 0.1840 & 0.4520 & 0.2380 & 0.5700 & 0.3420 & 0.6840 \\
                      & \dettok & 0.3300 & 0.0060 & 0.1460 & 0.4660 & 0.2500 & 0.5300 & 0.2960 & 0.6700 \\
\midrule
\multirow{3}{*}{$16k$} & \bpetok & 0.3400 & 0.0080 & 0.1900 & 0.5160 & 0.2700 & 0.5500 & 0.3180 & 0.6680 \\
                      & \biastok & 0.3500 & 0.0060 & 0.2000 & 0.5240 & 0.2840 & 0.5600 & 0.3280 & 0.6780 \\
                      & \dettok & 0.3620 & 0.0080 & 0.2300 & 0.5260 & 0.2680 & 0.5400 & 0.3140 & 0.6740 \\
\midrule
\multirow{3}{*}{$32k$} & \bpetok & 0.3660 & 0.0100 & 0.3380 & 0.5720 & 0.2720 & 0.5600 & 0.2560 & 0.6900 \\
                      & \biastok & 0.3720 & 0.0120 & 0.2920 & 0.5560 & 0.2680 & 0.5600 & 0.3280 & 0.6820 \\
                      & \dettok & 0.3860 & 0.0100 & 0.3060 & 0.5880 & 0.2820 & 0.6300 & 0.3760 & 0.6780 \\
\midrule
\multirow{3}{*}{$64k$} & \bpetok & 0.3660 & 0.0180 & 0.2900 & 0.6080 & 0.2860 & 0.5900 & 0.3380 & 0.6860 \\
                      & \biastok & 0.3780 & 0.0100 & 0.3400 & 0.5840 & 0.3180 & 0.6200 & 0.3860 & 0.7000 \\
                      & \dettok & 0.3740 & 0.0280 & 0.3080 & 0.5940 & 0.2940 & 0.5800 & 0.2540 & 0.7080 \\
\midrule
\multirow{3}{*}{$128k$} & \bpetok & 0.3840 & 0.0260 & 0.3760 & 0.6080 & 0.2660 & 0.5400 & 0.3680 & 0.6920 \\
                      & \biastok & 0.3840 & 0.0140 & 0.3800 & 0.6060 & 0.3120 & 0.5700 & 0.2240 & 0.6940 \\
                      & \dettok & 0.3720 & 0.0140 & 0.4180 & 0.6100 & 0.2960 & 0.6000 & 0.2280 & 0.6940 \\
\midrule
\multirow{3}{*}{$256k$} & \bpetok & 0.3960 & 0.0200 & 0.3520 & 0.6100 & 0.2880 & 0.6000 & 0.3320 & 0.6880 \\
                      & \biastok & 0.3800 & 0.0220 & 0.3740 & 0.6080 & 0.2980 & 0.5200 & 0.3640 & 0.6920 \\
                      & \dettok & 0.3800 & 0.0300 & 0.4060 & 0.6260 & 0.2880 & 0.6000 & 0.2580 & 0.6980 \\
\bottomrule
\end{tabular}
}
\end{table}

\begin{table}
\centering
\caption{Downstream Performance (2/3) --- Depth 12}[H]
\label{tab:downstream_perf_2_d12}
\footnotesize
\resizebox{\textwidth}{!}{%
\begin{tabular}{llccccccc}
\toprule
\makecell{Vocabulary \\ Size} & Tokeniser & \makecell{OpenBook \\ QA} & \makecell{LAMBADA \\ OpenAI} & Hellaswag & Winograd & Winogrande & \makecell{BigBench \\ Dyck Lang.} & \makecell{AGIEval \\ LSAT-AR} \\
\midrule
\multirow{3}{*}{$8k$} & \bpetok & 0.3180 & 0.2860 & 0.3320 & 0.5971 & 0.5520 & 0.0860 & 0.2304 \\
                      & \biastok & 0.3220 & 0.2700 & 0.3600 & 0.5604 & 0.5200 & 0.0980 & 0.2913 \\
                      & \dettok & 0.3200 & 0.2840 & 0.3400 & 0.5604 & 0.5300 & 0.1040 & 0.2348 \\
\midrule
\multirow{3}{*}{$16k$} & \bpetok & 0.3140 & 0.2900 & 0.3240 & 0.5531 & 0.5080 & 0.0920 & 0.2435 \\
                      & \biastok & 0.3240 & 0.2680 & 0.3540 & 0.5714 & 0.4920 & 0.1160 & 0.2435 \\
                      & \dettok & 0.3180 & 0.2840 & 0.3580 & 0.5824 & 0.5020 & 0.1180 & 0.1870 \\
\midrule
\multirow{3}{*}{$32k$} & \bpetok & 0.3160 & 0.3060 & 0.3680 & 0.5971 & 0.5160 & 0.1260 & 0.2435 \\
                      & \biastok & 0.3160 & 0.3080 & 0.3540 & 0.5714 & 0.5420 & 0.0400 & 0.2174 \\
                      & \dettok & 0.2860 & 0.3140 & 0.3780 & 0.5568 & 0.5300 & 0.0480 & 0.2174 \\
\midrule
\multirow{3}{*}{$64k$} & \bpetok & 0.3000 & 0.3040 & 0.3580 & 0.5897 & 0.5360 & 0.1300 & 0.2348 \\
                      & \biastok & 0.2900 & 0.3020 & 0.3680 & 0.5714 & 0.5360 & 0.1000 & 0.2391 \\
                      & \dettok & 0.3320 & 0.3080 & 0.3700 & 0.5824 & 0.5020 & 0.1080 & 0.2130 \\
\midrule
\multirow{3}{*}{$128k$} & \bpetok & 0.2680 & 0.3140 & 0.3700 & 0.5824 & 0.5200 & 0.1640 & 0.2043 \\
                      & \biastok & 0.3000 & 0.3040 & 0.3740 & 0.5897 & 0.5120 & 0.1260 & 0.2652 \\
                      & \dettok & 0.3040 & 0.3260 & 0.3640 & 0.5788 & 0.5060 & 0.1320 & 0.2261 \\
\midrule
\multirow{3}{*}{$256k$} & \bpetok & 0.2880 & 0.3060 & 0.3760 & 0.5861 & 0.5080 & 0.1440 & 0.2261 \\
                      & \biastok & 0.2880 & 0.2980 & 0.3820 & 0.5714 & 0.5200 & 0.1140 & 0.2217 \\
                      & \dettok & 0.2780 & 0.3140 & 0.3600 & 0.5788 & 0.4720 & 0.1180 & 0.2304 \\
\bottomrule
\end{tabular}
}
\end{table}

\begin{table}
\centering
\caption{Downstream Performance (3/3) --- Depth 12}
\label{tab:downstream_perf_3_d12}
\footnotesize
\resizebox{\textwidth}{!}{%
\begin{tabular}{llccccccc}
\toprule
\makecell{Vocabulary \\ Size} & Tokeniser & \makecell{BigBench \\ CS Alg.} & \makecell{BigBench \\ Operators} & \makecell{BigBench \\ Repeat Copy} & SQuAD & CoQA & BoolQ & \makecell{BigBench \\ Lang. ID} \\
\midrule
\multirow{3}{*}{$8k$} & \bpetok & 0.4560 & 0.0905 & 0.0000 & 0.2120 & 0.1800 & 0.5620 & 0.2700 \\
                      & \biastok & 0.4340 & 0.1095 & 0.0000 & 0.1660 & 0.1600 & 0.4900 & 0.2660 \\
                      & \dettok & 0.4480 & 0.0762 & 0.0312 & 0.1660 & 0.2040 & 0.5140 & 0.2660 \\
\midrule
\multirow{3}{*}{$16k$} & \bpetok & 0.4740 & 0.1095 & 0.0000 & 0.1600 & 0.1520 & 0.4240 & 0.2760 \\
                      & \biastok & 0.4640 & 0.1095 & 0.0000 & 0.1080 & 0.1920 & 0.5540 & 0.2920 \\
                      & \dettok & 0.4680 & 0.0905 & 0.0312 & 0.1280 & 0.1960 & 0.5640 & 0.2740 \\
\midrule
\multirow{3}{*}{$32k$} & \bpetok & 0.4060 & 0.0857 & 0.0000 & 0.2000 & 0.1620 & 0.4960 & 0.2460 \\
                      & \biastok & 0.4240 & 0.0905 & 0.0000 & 0.1240 & 0.1500 & 0.5100 & 0.2640 \\
                      & \dettok & 0.4160 & 0.1333 & 0.0000 & 0.2340 & 0.2000 & 0.5000 & 0.2600 \\
\midrule
\multirow{3}{*}{$64k$} & \bpetok & 0.4440 & 0.1286 & 0.0000 & 0.1660 & 0.1880 & 0.5200 & 0.2860 \\
                      & \biastok & 0.4400 & 0.1143 & 0.0312 & 0.2760 & 0.2040 & 0.5120 & 0.2360 \\
                      & \dettok & 0.4440 & 0.1238 & 0.0000 & 0.2620 & 0.2100 & 0.5180 & 0.2560 \\
\midrule
\multirow{3}{*}{$128k$} & \bpetok & 0.3900 & 0.1190 & 0.0312 & 0.1520 & 0.2000 & 0.5200 & 0.2480 \\
                      & \biastok & 0.4060 & 0.0762 & 0.0000 & 0.2300 & 0.1700 & 0.5160 & 0.2420 \\
                      & \dettok & 0.4460 & 0.0667 & 0.0312 & 0.1780 & 0.1640 & 0.5600 & 0.2580 \\
\midrule
\multirow{3}{*}{$256k$} & \bpetok & 0.4280 & 0.1095 & 0.0000 & 0.1980 & 0.1620 & 0.5440 & 0.2500 \\
                      & \biastok & 0.4360 & 0.1048 & 0.0312 & 0.1960 & 0.1780 & 0.5040 & 0.2720 \\
                      & \dettok & 0.4440 & 0.0810 & 0.0000 & 0.1900 & 0.1600 & 0.5000 & 0.2440 \\
\bottomrule
\end{tabular}
}
\end{table}

\subsection{Depth 18}

\begin{table}[H]
\centering
\caption{Downstream Performance (1/3) --- Depth 18}
\label{tab:downstream_perf_Depth 18_1}
\footnotesize
\resizebox{\textwidth}{!}{%
\begin{tabular}{llcccccccc}
\toprule
\makecell{vocabulary \\ size} & Tokeniser & \makecell{Hellaswag \\ zeroshot} & Jeopardy & \makecell{BigBench \\ QA Wikidata} & \makecell{ARC \\ Easy} & \makecell{ARC \\ Challenge} & COPA & \makecell{Commonsense \\ QA} & PIQA \\
\midrule
\multirow{2}{*}{$8k$} & \bpetok & 0.4560 & 0.0320 & 0.3620 & 0.5860 & 0.3460 & 0.6300 & 0.2120 & 0.7040 \\
                      & \biastok & 0.4440 & 0.0120 & 0.3240 & 0.5720 & 0.3320 & 0.6000 & 0.2300 & 0.7160 \\
\midrule
\multirow{2}{*}{$32k$} & \bpetok & 0.4680 & 0.0460 & 0.4100 & 0.6660 & 0.3700 & 0.6700 & 0.3140 & 0.7420 \\
                      & \biastok & 0.4700 & 0.0400 & 0.3920 & 0.6580 & 0.3660 & 0.6600 & 0.3140 & 0.7260 \\
\midrule
\multirow{2}{*}{$128k$} & \bpetok & 0.4980 & 0.0780 & 0.4680 & 0.7080 & 0.3980 & 0.6400 & 0.4140 & 0.7480 \\
                      & \biastok & 0.4900 & 0.0700 & 0.4560 & 0.7040 & 0.3900 & 0.6300 & 0.3240 & 0.7280 \\
\bottomrule
\end{tabular}
}
\end{table}

\begin{table}[H]
\centering
\caption{Downstream Performance (2/3) --- Depth 18}
\label{tab:downstream_perf_d18_2}
\footnotesize
\resizebox{\textwidth}{!}{%
\begin{tabular}{llccccccc}
\toprule
\makecell{vocabulary \\ size} & Tokeniser & \makecell{OpenBook \\ QA} & \makecell{LAMBADA \\ OpenAI} & Hellaswag & Winograd & Winogrande & \makecell{BigBench \\ Dyck Lang.} & \makecell{AGIEval \\ LSAT-AR} \\
\midrule
\multirow{2}{*}{$8k$} & \bpetok & 0.3620 & 0.3600 & 0.4440 & 0.5971 & 0.5260 & 0.1260 & 0.2391 \\
                      & \biastok & 0.3700 & 0.4020 & 0.4440 & 0.6007 & 0.5480 & 0.1320 & 0.2130 \\
\midrule
\multirow{2}{*}{$32k$} & \bpetok & 0.3740 & 0.3940 & 0.4600 & 0.6630 & 0.5340 & 0.1320 & 0.2565 \\
                      & \biastok & 0.3780 & 0.3980 & 0.4560 & 0.6520 & 0.5440 & 0.0320 & 0.2348 \\
\midrule
\multirow{2}{*}{$128k$} & \bpetok & 0.3420 & 0.3740 & 0.4940 & 0.6044 & 0.5400 & 0.1320 & 0.2652 \\
                      & \biastok & 0.3440 & 0.3880 & 0.4940 & 0.6484 & 0.5400 & 0.1280 & 0.2304 \\
\bottomrule
\end{tabular}
}
\end{table}

\begin{table}[H]
\centering
\caption{Downstream Performance (3/3) --- Depth 18}
\label{tab:downstream_perf_d18_3}
\footnotesize
\resizebox{\textwidth}{!}{%
\begin{tabular}{llccccccc}
\toprule
\makecell{vocabulary \\ size} & Tokeniser & \makecell{BigBench \\ CS Alg.} & \makecell{BigBench \\ Operators} & \makecell{BigBench \\ Repeat Copy} & SQuAD & CoQA & BoolQ & \makecell{BigBench \\ Lang. ID} \\
\midrule
\multirow{2}{*}{$8k$} & \bpetok & 0.4420 & 0.1571 & 0.0000 & 0.3020 & 0.2260 & 0.5580 & 0.2840 \\
                      & \biastok & 0.4280 & 0.1810 & 0.0312 & 0.2900 & 0.2640 & 0.5580 & 0.2580 \\
\midrule
\multirow{2}{*}{$32k$} & \bpetok & 0.4400 & 0.1381 & 0.0000 & 0.3760 & 0.2460 & 0.5680 & 0.2600 \\
                      & \biastok & 0.4320 & 0.1476 & 0.0000 & 0.2500 & 0.2420 & 0.6020 & 0.2660 \\
\midrule
\multirow{2}{*}{$128k$} & \bpetok & 0.4520 & 0.1524 & 0.0000 & 0.3940 & 0.2420 & 0.4860 & 0.2560 \\
                      & \biastok & 0.4660 & 0.1429 & 0.0000 & 0.3520 & 0.2980 & 0.5900 & 0.2480 \\
\bottomrule
\end{tabular}
}
\end{table}

\subsection{Depth 24}
\begin{table}[H]
\centering
\caption{Downstream Performance (1/3) --- d24}
\label{tab:downstream_perf_1_d24}
\footnotesize
\resizebox{\textwidth}{!}{%
\begin{tabular}{llcccccccc}
\toprule
\makecell{vocabulary \\ size} & Tokeniser & \makecell{Hellaswag \\ zeroshot} & Jeopardy & \makecell{BigBench \\ QA Wikidata} & \makecell{ARC \\ Easy} & \makecell{ARC \\ Challenge} & COPA & \makecell{Commonsense \\ QA} & PIQA \\
\midrule
\multirow{3}{*}{$8k$} & \bpetok & 0.5640 & 0.0740 & 0.4660 & 0.6660 & 0.4220 & 0.7100 & 0.3260 & 0.7400 \\
                      & \biastok & 0.5840 & 0.0660 & 0.4760 & 0.6600 & 0.4000 & 0.6800 & 0.2260 & 0.7700 \\
\midrule
\multirow{3}{*}{$32k$} & \bpetok & 0.5940 & 0.1120 & 0.5240 & 0.7280 & 0.4380 & 0.6600 & 0.2680 & 0.7680 \\
                      & \biastok & 0.5860 & 0.1280 & 0.5160 & 0.7200 & 0.4300 & 0.6300 & 0.2600 & 0.7560 \\
\midrule
\multirow{3}{*}{$128k$} & \bpetok & 0.5820 & 0.1720 & 0.5320 & 0.7260 & 0.4280 & 0.6700 & 0.3440 & 0.7620 \\
                      & \biastok & 0.5840 & 0.1640 & 0.5280 & 0.7540 & 0.4200 & 0.7000 & 0.2820 & 0.7820 \\
\bottomrule
\end{tabular}
}
\end{table}

\begin{table}[H]
\centering
\caption{Downstream Performance (2/3) --- d24}
\label{tab:downstream_perf_2_d24}
\footnotesize
\resizebox{\textwidth}{!}{%
\begin{tabular}{llccccccc}
\toprule
\makecell{vocabulary \\ size} & Tokeniser & \makecell{OpenBook \\ QA} & \makecell{LAMBADA \\ OpenAI} & Hellaswag & Winograd & Winogrande & \makecell{BigBench \\ Dyck Lang.} & \makecell{AGIEval \\ LSAT-AR} \\
\midrule
\multirow{3}{*}{$8k$} & \bpetok & 0.3780 & 0.4460 & 0.5560 & 0.6813 & 0.5280 & 0.1160 & 0.2652 \\
                      & \biastok & 0.3880 & 0.4180 & 0.5800 & 0.6740 & 0.5900 & 0.1340 & 0.2739 \\
\midrule
\multirow{2}{*}{$32k$} & \bpetok & 0.4040 & 0.4760 & 0.5900 & 0.6703 & 0.5540 & 0.1360 & 0.2522 \\
                      & \biastok & 0.4340 & 0.4740 & 0.5900 & 0.6813 & 0.5680 & 0.0620 & 0.2565 \\

\midrule
\multirow{2}{*}{$128k$} & \bpetok & 0.3760 & 0.4620 & 0.6000 & 0.7436 & 0.5760 & 0.1300 & 0.2522 \\
                      & \biastok & 0.3700 & 0.4620 & 0.5980 & 0.6667 & 0.5700 & 0.1300 & 0.2783 \\
\bottomrule
\end{tabular}
}
\end{table}

\begin{table}[H]
\centering
\caption{Downstream Performance (3/3) --- d24}
\label{tab:downstream_perf_3_d24}
\footnotesize
\resizebox{\textwidth}{!}{%
\begin{tabular}{llccccccc}
\toprule
\makecell{vocabulary \\ size} & Tokeniser & \makecell{BigBench \\ CS Alg.} & \makecell{BigBench \\ Operators} & \makecell{BigBench \\ Repeat Copy} & SQuAD & CoQA & BoolQ & \makecell{BigBench \\ Lang. ID} \\
\midrule
\multirow{2}{*}{$8k$} & \bpetok & 0.4320 & 0.2048 & 0.0312 & 0.4220 & 0.3020 & 0.5820 & 0.2820 \\
                      & \biastok & 0.4500 & 0.1619 & 0.0312 & 0.4040 & 0.2800 & 0.5600 & 0.2760 \\

\midrule
\multirow{2}{*}{$32k$} & \bpetok & 0.4340 & 0.1619 & 0.0000 & 0.4740 & 0.2880 & 0.5840 & 0.2400 \\
                      & \biastok & 0.4520 & 0.2048 & 0.0000 & 0.5100 & 0.3300 & 0.6200 & 0.2740 \\
\midrule
\multirow{2}{*}{$128k$} & \bpetok & 0.4380 & 0.1952 & 0.0000 & 0.4580 & 0.3220 & 0.5960 & 0.2420 \\
                      & \biastok & 0.4440 & 0.1476 & 0.0312 & 0.4860 & 0.2840 & 0.6340 & 0.2600 \\
\bottomrule
\end{tabular}
}
\end{table}

\end{document}